\documentclass[lettersize,journal]{IEEEtran}

\usepackage{algorithm}
\usepackage{algorithmic}

\usepackage{amsthm}
\usepackage{amsmath,amssymb,mathtools}
\usepackage{makecell}
\usepackage{booktabs}
\usepackage{footnote}
\usepackage{cite}
\usepackage[switch]{lineno}  %

\newtheorem{theorem}{Theorem}

\newtheorem{lemma}{Lemma}
\newtheorem{corollary}{Corollary}
\newtheorem{assumption}{Assumption}

\newtheorem{remark}{Remark}

\usepackage{ifpdf}
\usepackage[flushleft]{threeparttable}
\usepackage{makecell}
\usepackage{multirow}
\usepackage{bbding}
\usepackage{graphicx}
\usepackage{dsfont}

\ifCLASSINFOpdf
\else
\fi
\begin{document}
%
\title{LightSAM: Parameter-Agnostic Sharpness-Aware Minimization}
%
%
%

\author{Yifei~Cheng, Li~Shen, Hao~Sun, Nan~Yin, Xiaochun~Cao,~\IEEEmembership{Senior Member,~IEEE}, Enhong~Chen,~\IEEEmembership{Fellow,~IEEE}

\IEEEcompsocitemizethanks{
\IEEEcompsocthanksitem Y. Cheng, L. Shen and X. Cao are with the School of Cyber Science and Technology, Sun Yat-sen University, Shenzhen Campus. E-mail: \{yfcheng.ifc, mathshenli\}@gmail.com, caoxiaochun@mail.sysu.edu.cn. \protect
\IEEEcompsocthanksitem H. Sun and E. Chen are with the School of Computer Science, University of Science and Technology of China. E-mail: ustcsh@mail.ustc.edu.cn, cheneh@ustc.edu.cn. \protect
\IEEEcompsocthanksitem N. Yin is with Hong Kong University of Science and Technology. E-mail: yinnan8911@gmail.com. \protect
\IEEEcompsocthanksitem Corresponding author: Li Shen.}
}

%
%

\markboth{Journal of \LaTeX\ Class Files,~Vol.~14, No.~8, August~2015}%
{Shell \MakeLowercase{\textit{et al.}}: Bare Advanced Demo of IEEEtran.cls for IEEE Computer Society Journals}
%



\IEEEtitleabstractindextext{%

\begin{abstract}
Sharpness-Aware Minimization (SAM) optimizer enhances the generalization ability of the machine learning model by exploring the flat minima landscape through weight perturbations. Despite its empirical success, SAM introduces an additional hyper-parameter, the perturbation radius, which causes the sensitivity of SAM to it. Moreover, it has been proved that the perturbation radius and learning rate of SAM are constrained by problem-dependent parameters to guarantee convergence. These limitations indicate the requirement of parameter-tuning in practical applications. In this paper, we propose the algorithm LightSAM which sets the perturbation radius and learning rate of SAM adaptively, thus extending the application scope of SAM. LightSAM employs three popular adaptive optimizers, including AdaGrad-Norm, AdaGrad and Adam, to replace the SGD optimizer for weight perturbation and model updating, reducing sensitivity to parameters. Theoretical results show that under weak assumptions, LightSAM could converge ideally with any choices of perturbation radius and learning rate, thus achieving parameter-agnostic. We conduct preliminary experiments on several deep learning tasks, which together with the theoretical findings validate the the effectiveness of LightSAM.
\end{abstract}

}

\maketitle

\IEEEdisplaynontitleabstractindextext

%
\IEEEpeerreviewmaketitle

\section{Introduction}
\IEEEPARstart{M}{achine} learning has achieved significant success across various application domains. As a critical component of machine learning, many optimization approaches are explored to train the model efficiently. However, most of the previous works only focus on minimizing the training loss, which would face the dilemma of over-fitting since the popular models are over-parameterized. Recently, there has been a raised attention on generalization ability since it represents the prediction ability on unseen data, thus very crucial for a model. Keskar et al. \cite{keskar2016large} and Neyshabur et al. \cite{neyshabur2017exploring} study the relationship between the flatness of loss landscape and generalization ability, which consequently suggests finding flat minima that have low curvature in the neighbourhoods.

The above idea is formalized as a novel minimax problem, named Sharpness-Aware Minimization \cite{foret2020sharpness}. The main difference from the original loss function is that Sharpness-Aware Minimization has a step that maximizes the loss function in the neighbourhood. This consideration of worst-case guarantees the low loss value in a region, thus making the loss landscape of minima flat and improving generalization ability, which results in the novel SAM optimizer: in each iteration, a weight perturbation is performed along the gradient direction with radius $\rho$, then the stochastic gradient on the perturbed weight is used in gradient descent with learning rate $\eta$ to update the model. SAM significantly improves the test performances of several deep networks \cite{foret2020sharpness}.

The convergence rates of SAM and its variants have been extensively analyzed in existing works \cite{andriushchenko2022towards,mi2022make,shin2023effects,sun2024adasam}. However, these theoretical results require restrictions on two hyper-parameters of perturbation radius $\rho$ and learning rate $\eta$, either upper bounded or unequal relationship between them. These restrictions usually involve some problem-dependent constants, such as the Lipschitz constant, whose value could not be obtained a prior and hard to be estimated. In addition, though it is proved that the normalization in the perturbation step makes SAM less sensitive on $\rho$ \cite{dai2023crucial}, the empirical studies in the above works show that the sensitivity to the learning rate still exists and the adopted values are not stable. These shortcomings make it necessary to do parameter-tuning in empirical studies, which increases cost especially when training large-scale models. Thus, we raise a question that: 

\begin{center}
    \textit{Can we make SAM parameter-agnostic}\footnote{In this paper, we follow the definition "parameter-agnostic" in \cite{wang2024convergence,hubler2024parameter} to describe an algorithm that could guarantee convergence with any parameter values. This implies that all parameters are not contingent upon any problem-dependent constants. 
    }?
\end{center}

In fact, parameter-agnostic algorithms are thoroughly studied in online learning to avoid parameter-tuning \cite{orabona2014simultaneous,cutkosky2017online,orabona2017training}. Recently, Defazio \& Mishchenko \cite{defazio2023learning} suggest to use Adagrad-like step size to achieve learning-rate-agnostic. Wang et al. \cite{wang2023convergence} and Wang et al. \cite{wang2023closing} prove the ideal convergence rate for adaptive optimizers. These motivate us to introduce adaptive learning rate into SAM to realize parameter-agnostic. Note that directly introducing adaptivity for both the perturbation radius and learning rate is technically non-trivial. This is due to that the terms that need to be bounded would involve two gradients in one iteration, and the relationship between them is hard to establish since the randomnesses in one term could not be decoupled directly in the proof for adaptive methods. 

In this paper, we study how to make the SAM optimizer parameter-agnostic. To achieve this goal, we propose an algorithm LightSAM. We provide three options for LightSAM, and in each option, we adopt one commonly used adaptive optimizer to perform weight perturbation and model update instead of SGD in vanilla SAM. As a consequence, both the weight perturbation and model update become adaptive during training. Specifically, we adopt the AdaGrad-Norm-type learning rate for LightSAM, named  LightSAM-I, which uses a scaler-type adaptive learning rate for both the perturbation ascent step and gradient descent step $(\rho,\eta)$. In addition, we also consider the AdaGrad-type and Adam-type learning rate for LightSAM, named LightSAM-II and LightSAM-III respectively, which use coordinate-wise learning rates for two hyper-parameters $(\rho,\eta)$. Theoretically, we prove the $\mathbb{E}\|\nabla f(x_t)\| \leq O(\ln T / T^{1/4})$ convergence rate for LightSAM without any restrictions on perturbation radius and learning rate, thus achieving parameter-agnostic optimizers. Additionally, we only require nearly the weakest assumptions among related studies.

Our contributions can be summarized as follows:
\begin{itemize}
    \item We propose an algorithm LightSAM for non-convex optimization. Compared to SAM, our algorithm could adopt AdaGrad-Norm, AdaGrad or Adam to implement the weight perturbation and model update steps. As a result, both the perturbation radius and learning rate become adaptively adjusted without requiring problem-dependent unknown parameters.
    \item The theoretical analysis indicates that LightSAM achieves the $\mathbb{E}\|\nabla f(x_t)\| \leq O(\ln T / T^{1/4})$ convergence rate without the gradient bounded assumption which is commonly used in adaptive optimizer analysis. Our result holds under any choices of hyper-parameters $(\rho,\eta)$, indicating that LightSAM is a parameter-agnostic optimizer, thereby saving the cost of parameter-tuning. 
    \item The technicality of our proof is mainly reflected in two aspects: firstly, we deal with the misalignment between the norms of gradients of two model parameters $x_t$ and $w_t$ by applying the $L$-smoothness inequality on both of them to establish the relationship (the first step in the "Proof Sketch"); secondly, we propose two lemmas (Lemmas \ref{bound} and \ref{bound2} below) to solve the complex inequalities encountered in the proof.
    \item We conduct several experiments to show the effectiveness of LightSAM, whose performance is stable under different parameter settings and coincides with our theoretical findings.
    \end{itemize}

\section{Related Work}

\subsection{Sharpness-Aware Minimization.}
SAM optimizer \cite{foret2020sharpness} enhances the model generalization ability by minimizing the sharpness of loss landscape through an extra step of parameter perturbation. Wen et al. \cite{wen2023sharpness} reveal the mechanism of SAM by analyzing the relationship between sharpness-aware loss and the Hessian of the original loss function. However, SAM still has some shortcomings in practical use, e.g., double gradient calculation and double learning rate hyper-parameter tuning. To address the issue where SAM exhibits insensitivity to parameter scaling, Kwon et al. \cite{kwon2021asam} propose ASAM which incorporates a normalization operator into the perturbation step to ensure adaptive sharpness. R-SAM \cite{liu2022random} suggests adding noise into the perturbation step to further maximize the loss function in the neighborhood. 
Recognizing the increased computational cost due to SAM's double forward and backward steps, SSAM \cite{mi2022make} generates a mask to sparsify the perturbation while SAF \cite{du2022sharpness} replaces SAM's sharpness measure loss with a trajectory loss to achieve almost zero additional computation cost. GSAM \cite{zhuang2022surrogate} introduces an ascent step in the orthogonal direction to minimize the surrogate gap. SAMAR \cite{zou2024sharpness} views the sharpness reduction as a regularization and tunes the regularization parameter adaptively by measuring the sharpness change. SAMPa \cite{xie2024sampa} parallelizes two gradient calculations to reduce the computational time to half of SAM. SALA \cite{tan2024sharpness} performs the weight perturbation step once the distance between the slow and fast weights is shorter than the threshold. Un-normalized SAM (USAM) \cite{andriushchenko2022towards} removes the normalization term in SAM and analyzes the convergence. However, in order to guarantee the $O(1/\sqrt{T})$ convergence rate, the values of perturbation radius $\rho$ and learning rate $\eta$ are required to be dependent on the smoothness constant. Furthermore, Sun et al. \cite{sun2024adasam} propose the adaptive SAM by utilizing AMSGrad-type \cite{reddi2019convergence} learning rate in SAM. However, the perturbation radius still requires heavy tuning. 

\begin{table*}[t]
\renewcommand{\arraystretch}{1.6}
    \caption{Comparison between SAM-related works.}
    \centering
    \begin{threeparttable}
    \begin{tabular}{c|c|c|c|c}
    \hline 
    Algorithm & \makecell[c]{Adaptive pert-\\urbation radius} & \makecell[c]{Adaptive\\learning rate} & Convergence rate\tnote{a} & Additional requirements \\
    \hline 
    SAM  &  \XSolidBrush & \XSolidBrush & $\mathbb{E}\|\nabla f(x_t)\|^2 \leq O(\ln T / \sqrt{T})$\tnote{b} & Gradient bounded; Dependent on gradient bound \\
    \hline 
    USAM & \XSolidBrush & \XSolidBrush & $\mathbb{E}\|\nabla f(x_t)\|^2 \leq O(1/\sqrt{T})$ & Dependent on Lipschitz constant \\
    \hline 
    ASAM & \Checkmark & \XSolidBrush & - & - \\
    \hline 
    AdaSAM & \XSolidBrush & \Checkmark & $\mathbb{E}\|\nabla f(x_t)\|^2 \leq O(1/\sqrt{T})$ & 
    Dependent on Lipschitz constant; Gradient bounded \\
    \hline 
    \makecell[c]{LightSAM\\(this work)} & \Checkmark & \Checkmark & $\mathbb{E}\|\nabla f(x_t)\| \leq O(\ln T / T^{1/4})$ & None \\
    \hline 
    \end{tabular}
    \begin{tablenotes}
    \item[a]``-'' represents the convergence rate is not given in the work. 
    \item[b] This result is obtained in \cite{mi2022make} and could be improved to $O(1/\sqrt{T})$ by adjusting values of hyper-parameters. We maintain the result in the original work here.
    \end{tablenotes}
    \end{threeparttable}
    \label{comp}
\end{table*}

\subsection{Adaptive Optimizer.}
Adaptive optimizers make the learning rate adjust adaptively during the training process. Duchi et al. \cite{duchi2011adaptive} propose Adagrad, which accumulates the gradient second raw moment, i.e. the square of historical gradients, and makes the learning rate of each element inversely proportional to the square root of this sum. RMSProp \cite{tieleman2012lecture} suggests adopting an exponential moving average for the stochastic gradients to make adaptive optimizer work well in deep learning. Adam \cite{kingma2014adam} further introduces the exponential moving average to the gradient second raw moment and becomes the most commonly used adaptive method. AMSGrad \cite{reddi2019convergence} improves the performance of Adam by making the second-order momentum non-decreasing.

It is showed that Adagrad could converge in both convex and non-convex settings \cite{li2019convergence}. Adam-type algorithms achieve the $O(\ln T / \sqrt{T})$ convergence rate for non-convex optimization problems \cite{chen2018convergence}. The convergence rate $O(\sqrt{d/T})$ for AMSGrad, and $O(d/\sqrt{T})$ for Adagrad and RMSProp are theoretically proved \cite{zhou2018convergence}. Additionally, D{\'e}fossez et al. \cite{defossez2020simple} and Shen et al. \cite{shen2023unified} analyze Adagrad and Adam under a framework with momentum and recover the $O(\ln T / \sqrt{T})$ convergence rate. However, most of these theoretical results rely on a strong assumption, i.e. the stochastic gradient is upper bounded. The analysis for RMSProp removes this assumption and concludes the convergence to a bounded region \cite{shi2021rmsprop}. With the hyper-parameters commonly used in practice, Adam also converges to a region near critical points \cite{zhang2022adam}. Recently, Wang et al. \cite{wang2023convergence} and Wang et al. \cite{wang2023closing} make breakthroughs that recover the $O(\ln T / \sqrt{T})$ convergence rate without gradient bounded assumption.

\subsection{Parameter-Agnostic Optimization.}
Parameter-agnostic (also known as parameter-free) algorithms are studied to achieve the optimal regret bound for the online optimization problem at first \cite{orabona2013dimension,mcmahan2014unconstrained,orabona2016coin}. 
Kernel-based SGD \cite{orabona2014simultaneous} performs model selection and optimization without prior knowledge of problem and parameter-tuning.
Orabona \& Tommasi \cite{orabona2017training} remove the learning rate from the gradient descent step to optimize the objective function. Carmon \& Hinder \cite{carmon2022making} focus on stochastic optimization and select the learning rate by a computable certificate. As a result, a nearly optimal convergence rate and parameter-agnostic are both achieved. D-Adaptation \cite{defazio2023learning} adopts Adagrad-like learning rate to iteratively lower bound the distance between the initial and optimal point. Ivgi et al. \cite{ivgi2023dog}, Khaled et al. \cite{khaled2023dowg} and Tao et al. \cite{tao2024towards} design the learning rate as the ratio of maximal model distance to the root of the sum of historical gradients' squares, consequently making the base optimizer parameter-free. Normalized SGDM \cite{hubler2024parameter} converges with a nearly optimal rate in the $(L_0,L_1)$-smoothness setting.

The above mentioned SAM-related works adopt SGD optimizer in weight perturbation or model update or both, which makes the parameters lack of adaptivity, and adaptive optimizer-related works seldom consider enhancing the generalization ability. Our work improves this by making both the perturbation radius and learning rate adaptive, and further parameter-agnostic. The most related work to this paper is \cite{sun2024adasam}. However, it only employs the adaptive learning rate in the gradient descent step. Furthermore, their analysis requires the gradient bound assumption, which is too strong to be satisfied for practical applications \cite{nguyen2018sgd}. We also notice SA-SAM \cite{naganumasmoothness} which sets the learning rate by adaptively estimating the local smoothness constant, but it lacks of convergence guarantee. We list the comparison between these works and our work in Table \ref{comp}.

\section{Methodology}

In this section, we propose a class of parameter-agnostic variants of SAM optimizer, named LightSAM. LightSAM could adopt the Adagrad-Norm-type learning rate \cite{levy2017online,ward2020adagrad}, AdaGrad-type learning rate \cite{duchi2011adaptive} and Adam-type learning rate \cite{kingma2014adam} for estimating the double learning rate hyperparameters in SAM optimizer, denoted as LightSAM-I (AdaGrad-Norm), LightSAM-II (AdaGrad) and LightSAM-III (Adam) respectively. Below, we first introduce the problem setup for SAM and LightSAM.

\subsection{Problem Setup}
In this paper, we focus on the following stochastic non-convex optimization problem:
\begin{equation}
    \underset{x\in R^d}{\min} \; f(x) := \frac{1}{n} \sum_{i=1}^n f(x,\xi_i), \nonumber
\end{equation}
where $f(x,\xi_i)$ denotes the loss function about $d$-dimensional model weights $x$ and data $\xi_i$, $n$ represents the number of training data. We further assume that this optimization problem is well-defined.

\textbf{Notations.} We use the following notations in this paper: $\|\cdot\|_1$ and $\|\cdot\|$ denote the $l_1$ and $l_2$ norm of a vector. $\mathds{1}_d$ represents a $d$-dimensional vector with all elements equal to 1. $\nabla f(x)$ represents the gradient of function $f(x)$, $\nabla f(x)_l$ represents the $l$-th element of $\nabla f(x)$. $\odot$ represents element-wise multiplication. For the vector sequences $\{a_t\}$, $a_{t,l}$ denotes the $l$-th element of $a_t$.

\textbf{SAM Optimizer.} Sharpness-Aware Minimization problem \cite{foret2020sharpness}  focuses on minimax saddle point optimization to seek a flat minimum by introducing the weight perturbation step
\begin{equation}
    \underset{x}{\min} \underset{\|\epsilon\| \leq \rho}{\max} f_{\mathcal{S}}(x + \epsilon). \nonumber
\end{equation}
By alternatively performing a dual ascent step for the perturbation and a gradient descent step for the primal weight, SAM takes the following two-time scale update rule: 
\begin{gather}
    w_t = x_t + \rho \nabla f(x_t,\xi_t)/\|\nabla f(x_t,\xi_t)\|, \label{perturbation-step} \nonumber \\
    x_{t+1} = x_t - \eta \nabla f(w_t,\xi_t) \label{descent-step} \nonumber. 
\end{gather}
According to this update rule, SAM faces the challenge that there
exist two learning rate hyperparameters $(\rho, \eta)$ that need to be carefully tuned.  \cite{dai2023crucial} show that the learning rate $\rho$ for the perturbation step is crucial for the final performance of SAM. Classic trial-and-error learning tuning techniques for $\rho$ suffer from high tuning costs due to double gradient calculation in SAM. It is urgent to design cheap,  lightweight, and automatic learning rate tuning techniques for SAM.

\subsection{LightSAM-I (AdaGrad-Norm)}

\begin{algorithm}[t]
  \caption{LightSAM-I (AdaGrad-Norm)}\label{alg1}
  \begin{algorithmic}[1]
  \REQUIRE Initial values $x_0$, $u_0=v_0=\epsilon^2$, perturbation radius $\rho$, learning rate $\eta$.
  \FOR{$t = 1, ..., T$}
    \STATE Sample a minibatch $\xi_t$ from the dataset;
    \STATE Compute stochastic gradient $s_t = \nabla f(x_t, \xi_t)$;
    \STATE $u_t= u_{t-1} + \|s_t\|^2$;
    \STATE $w_t = x_t + \rho \frac{s_t}{\sqrt{u_t}}$;
    \STATE Compute stochastic gradient $g_t = \nabla f(w_t, \xi_t)$;
    \STATE $v_t = v_{t-1} + \|g_t\|^2$;
    \STATE Update weights $x_{t+1} = x_t - \eta \frac{g_t}{\sqrt{v_t}}$;
  \ENDFOR
  \end{algorithmic}
\end{algorithm}

In this section, we propose our first algorithm LightSAM-I as described in Algorithm \ref{alg1}. Adagrad-Norm \cite{levy2017online,ward2020adagrad} only updates the scalar learning rate by historical gradients rather than the element-wise learning rate in AdaGrad. In the weight perturbation steps (lines 3-5) of our algorithm, we use the Adagrad-Norm to generate the perturbed weights $w_t$ instead of SGD optimizer in SAM. Meanwhile, we adopt the same strategy in the gradient descent steps (lines 6-8) to update model weights.

Before giving the theoretical analysis for Algorithm \ref{alg1}, we list some necessary assumptions. We denote $\mathcal{F}_t=\sigma\{s_1,g_1,...,s_t,g_t\}$ as the sigma algebra generated by the observations of LightSAM after observing the stochastic gradients in the first $t$ iterations. $\mathbb{E}^{|\mathcal{F}_t}[\cdot]$ represents $\mathbb{E}[\cdot|\mathcal{F}_t]$, and $\mathbb{E}$ denotes taking expectation over all randomnesses.
\begin{assumption}[$L$-smoothness]\label{assu1}
    $f(x,\xi)$ is differentiable and satisfies the following inequality:
    \begin{equation}
        \| \nabla f(x,\xi) - \nabla f(y,\xi) \| \leq L \|x - y \|, \forall x, y \in R^d.
    \nonumber
    \end{equation}
\end{assumption}

\begin{assumption}[Affine noise variance]\label{assu2}
    There exist positive constants $(D_0,D_1)$ such that the following inequality holds:
    \begin{equation}
        \mathbb{E}^{|\mathcal{F}_t} \|\nabla f(x, \xi)\|^2 \leq D_0 + D_1 \|\nabla f(x)\|^2, \forall x\in R^d.
    \nonumber
    \end{equation}
\end{assumption}

Straightforwardly, we could obtain the $L$-smoothness of $f(x)$ based on Assumption \ref{assu1}. These two assumptions are nearly the weakest requirements in stochastic optimization works, except that Assumption \ref{assu1} assumes the $L$-smoothness of $f(x,\xi)$ instead of $f(x)$ as Assumption 1 in \cite{wang2023convergence}. This change is necessary for SAM-type works \cite{andriushchenko2022towards} since we need to establish the relationship between two stochastic gradients ($\nabla f(x_t,\xi_t)$ and $\nabla f(w_t,\xi_t)$) in one iteration.

\textbf{Technical Challenge.} 
In order to prove the convergence, we need to bound the term $\mathbb{E} \|\nabla f(x_t)\|^2$. However, LightSAM involves two stochastic gradients in one iteration. Thus when we want to bound the terms concerning $\mathbb{E} \|\nabla f(x_t)\|^2$, the upper bound would contain the terms concerning $\mathbb{E} \|\nabla f(w_t)\|^2$. On the other hand, the numerator and denominator of one term in adaptive optimization often share the same randomness which is hard to decouple. Thus, it is hard to establish the inequality relationship in the analysis for LightSAM.

Based on the above assumptions, we have the following theorem.
\begin{theorem}\label{theo1}
If $f(x)$ in Algorithm \ref{alg1} satisfies Assumptions \ref{assu1} and \ref{assu2}, for any perturbation radius $\rho$ and learning rate $\eta > 0$, we have that 
\begin{eqnarray}
    \makebox[7.8cm][l]{\ensuremath{\displaystyle \sum_{t=1}^T \mathbb{E} \|\nabla f(x_t)\| \leq  T^{\frac{1}{2}}[2A_6(A_3+2A_5\ln A_6)}} \nonumber \\
    \makebox[7.8cm][l]{\ensuremath{\displaystyle + 2A_7\ln(A_7\!+\!e) + 4096D_1^2 A_4^2 (2A_4\!+\!\frac{16D_1 A_4 A_5}{A_6})^2\!+\!1]^{\frac{1}{2}}}} \nonumber
\end{eqnarray}
Here, we denote constants $D_2, A_1$ to $A_7$ as following
\begin{equation}
    \makebox[8.5cm][l]{\ensuremath{\displaystyle D_2 = \max\{1,4D_1,32(1+\sqrt{D_1})D_1 \rho\sqrt{\rho L +\epsilon}/ (\eta \sqrt{\epsilon})\},}} \nonumber 
\end{equation}
\begin{equation}
    \makebox[8.5cm][l]{\ensuremath{A_1 = \frac{\|\nabla f(\hat{w}_1)\|^2}{\epsilon} + \frac{4(1+2 D_2)L^2}{\epsilon} (\eta^2 - 4 \rho^2 \ln \epsilon),}}\nonumber 
\end{equation}
\begin{eqnarray}
    \makebox[7.8cm][l]{\ensuremath{\displaystyle A_2 = 2\rho^2L\!+\!\frac{\rho D_0}{2\epsilon \sqrt{D_1}}\!+\!\rho \|\nabla f(x_1)\|\!+\!\frac{(D_0+4\rho^2 L^2)\eta}{\epsilon}}} \nonumber \\
    \makebox[7.8cm][l]{\ensuremath{\displaystyle +f(w_1)\!-\! (\frac{12\rho^2 \eta L^2}{\epsilon}\!+\!\eta\!+\!\rho\!+\!(1\!+\!\rho L)(2\eta^2 L\!+\!8\rho^2 L)) \ln \epsilon,}} \nonumber
\end{eqnarray}
\begin{eqnarray}
    \makebox[7.8cm][l]{\ensuremath{\displaystyle A_3\!=\!\sqrt{\frac{\rho L}{\epsilon}\!+\!1}[\frac{4f(x_1)}{\eta}\!+\!\frac{8(D_0\!+\!4\rho^2 L^2)}{\epsilon}\!+\!16 D_1 A_1\!+\!\frac{8A_2}{\eta}}} \nonumber \\
    \makebox[7.8cm][l]{\ensuremath{\displaystyle -(\frac{80 \rho^2 L^2}{\epsilon} + 4\eta L) \ln \epsilon + (4\eta L (3+\rho L)+8) \ln (1\!+\!\frac{\rho L}{\epsilon})],}} \nonumber
\end{eqnarray}
\begin{equation}
    \makebox[8.5cm][l]{\ensuremath{\displaystyle A_4 = \sqrt{\rho L+\epsilon}(16(8(1+2D_2)D_1+3)\rho^2 L^2/\epsilon^{3/2},}} \nonumber 
\end{equation}
\begin{eqnarray}
    \makebox[7.8cm][l]{\ensuremath{\displaystyle A_5 = \sqrt{\frac{\rho L}{\epsilon}+1}[\frac{40\rho^2 L^2}{\epsilon} +\frac{4\rho}{\eta}(1+8\rho L(1+\rho L))}}\nonumber \\
    \makebox[7.8cm][l]{\ensuremath{\displaystyle+4\eta L (3+2\rho L)+8],}} \nonumber
\end{eqnarray}
\begin{equation}
    \makebox[8.5cm][l]{\ensuremath{\displaystyle A_6 = 2\sqrt{2D_0 T + \epsilon^2} + 4D_1 A_3 + 8D_1 A_5 \ln (4D_1 A_5\!+\!e),}} \nonumber
\end{equation}
\begin{equation}
    \makebox[8.5cm][l]{\ensuremath{\displaystyle A_7 = 2A_4 A_6+16D_1A_4A_5+8D_1A_4(A_3+2A_5\ln A_6).}} \nonumber
\end{equation}
\end{theorem}
\begin{corollary}\label{coro1}
From Theorem \ref{theo1}, we notice that $A_6=O(\sqrt{T})$ and $A_7=O(\sqrt{T}+\ln T)$, thus we can obtain the following convergence rate for Algorithm \ref{alg1}
\begin{equation}
    \frac{1}{T} \sum_{t=1}^T \mathbb{E} \|\nabla f(x_t)\| \leq O\bigg(\frac{\ln T}{T^{1/4}}\bigg). \nonumber
\end{equation}
\end{corollary}
\begin{remark}\label{remark1}
This convergence rate of LightSAM recovers the result in previous works about adaptive optimizers \cite{zou2019sufficient,defossez2020simple,ward2020adagrad,shi2021rmsprop,shen2023unified,wang2023convergence}. When $T$ is sufficiently large, it converges with the same rate as USAM \cite{andriushchenko2022towards}.
\end{remark}
\begin{remark}\label{remark2}
LightSAM not only requires nearly the lowest requirements on the assumptions but also has no restrictions on hyper-parameters, thus achieving parameter-agnostic.
\end{remark}

Due to limited space, we list the proof sketch here. The details could be referred to the Supplementary Material.
\begin{IEEEproof}[Proof Sketch] Firstly, we would aim to bound the objective $\sum_{t=1}^T \mathbb{E} \|\nabla f(x_t)\|^2 / \sqrt{v_{t-1}}$. Applying the $L$-smoothness of $f(\cdot)$ on $\{x_t\}$, we have
\begin{eqnarray}\label{theo1_1}
\hspace*{-0.8cm}&&\mathbb{E}[f(x_{T+1})] \leq f(x_1) + \underbrace{\eta \sum_{t=1}^T \mathbb{E} \langle \nabla f(x_t), \frac{-g_t}{\sqrt{v_{t-1}}} \rangle}_{T_1} \nonumber \\
\hspace*{-0.8cm}&&+\underbrace{\eta \sum_{t=1}^T \mathbb{E}\langle \nabla f(x_t), \frac{g_t}{\sqrt{v_{t-1}}}-\frac{g_t}{\sqrt{v_t}} \rangle}_{T_2}+ \underbrace{\frac{\eta^2 L}{2} \sum_{t=1}^T \mathbb{E}\|\frac{g_t}{\sqrt{v_t}}\|^2}_{T_3} 
\end{eqnarray}
Since $T_1$ and $T_3$ is easy to bound, we turn to focus on $T_2$. We define a virtual sequence $\{\hat{w}_t\}$ as $\hat{u}_0 = u_0, \ \hat{u}_t = \hat{u}_{t-1} + \|\nabla f(x_t)\|^2, \ \hat{w}_t = x_t + \rho \frac{\nabla f(x_t)}{\sqrt{\hat{u}_t}}$
to remove the randomness in the perturbation parameter $w_t$. Further, with appropriate derivation and Assumption \ref{assu2}, we obtain that
\begin{eqnarray}\label{theo1_2}
\hspace*{-0.2cm}T_2&\!\!\leq\!& \frac{\eta}{4} \sum_{t=1}^T\mathbb{E}\frac{\|\nabla f(x_t)\|^2}{\sqrt{v_{t-1}}}+ \frac{(2D_0+8\rho^2 L^2)\eta}{\epsilon} \nonumber \\
\hspace*{-0.2cm}&&+ 2D_1 \eta \sum_{t=1}^T\mathbb{E}\|\nabla f(\hat{w}_t)\|^2 (\frac{1}{\sqrt{v_{t-1}}}-\frac{1}{\sqrt{v_t}})
\end{eqnarray}
The last term above has a similar form to the term $\sum_{t=1}^T \|\nabla f(x_t)\|^2 \mathbb{E} (\frac{1}{\sqrt{v_{t-1}}}-\frac{1}{\sqrt{v_t}})$ in the proof of \cite{wang2023convergence}, which could be straightforwardly bounded by the targeted term $\sum_{t=1}^T \mathbb{E} \|\nabla f(x_t)\|^2 / \sqrt{v_{t-1}}$ in that work. However, this derivation does not hold in our proof because of the misalignment between $\|\nabla f(x_t)\|^2$ and $\|\nabla f(\hat{w}_t)\|^2$ which comes from that SAM-type algorithms involve different weights $x_t$ and $w_t$. Thus, it is non-trivial to bound the last term in (\ref{theo1_2}). We give the following two lemmas to fill this gap, the second of which is obtained by applying $L$-smoothness on $\{w_t\}$.

\begin{lemma}\label{lemma1}
If $f(x)$ in Algorithm \ref{alg1} satisfies Assumptions \ref{assu1} and \ref{assu2}, we have that
\begin{eqnarray}
\hspace*{-0.7cm}&&\sum_{t=1}^T \mathbb{E}\|\nabla f(\hat{w}_t)\|^2 (\frac{1}{\sqrt{v_{t-1}}}-\frac{1}{\sqrt{v_t}}) \leq A_1 - \mathbb{E}\frac{\|\nabla f(\hat{w}_T)\|^2}{\sqrt{v_T}} \nonumber \\
\hspace*{-0.7cm}&&+ \frac{1}{2 D_2}\sum_{t=1}^T \mathbb{E} \frac{\|\nabla f(\hat{w}_t)\|^2}{\sqrt{v_{t-1}}} + \frac{8(1+2 D_2) \rho^2 L^2}{\epsilon} \ln \hat{u}_T. \nonumber
\end{eqnarray}
\end{lemma}

\begin{lemma}\label{lemma2}
If $f(x)$ in Algorithm \ref{alg1} satisfies Assumptions \ref{assu1} and \ref{assu2}, we have that
\begin{eqnarray}
\hspace*{-0.7cm}&&\!\eta\!\sum_{t=1}^{T-1} \frac{\|\nabla f(\hat{w}_t)\|^2}{\sqrt{v_{t-1}}} \leq 4(1\!+\!\sqrt{D_1})\rho \mathbb{E} \frac{\|\nabla f(x_T)\|^2}{\sqrt{u_{T-1}}}  \nonumber \\
\hspace*{-0.7cm}&&+ (\frac{12\rho^2 \eta L^2}{\epsilon}+16\rho^2 L+16\rho^3 L^2 + 2\rho) \mathbb{E} \ln u_T \nonumber \\
\hspace*{-0.7cm}&& + \frac{12\rho^2 \eta L^2}{\epsilon} \mathbb{E}\ln \hat{u}_T+(2\eta\!+\!4\eta^2 L \!+\!4\rho \eta^2 L^2)\mathbb{E} \ln v_T \nonumber \\
\hspace*{-0.7cm}&& 4D_1 \eta \sum_{t=1}^{T-1} \mathbb{E}\|\nabla f(\hat{w}_t)\|^2 (\frac{1}{\sqrt{v_{t-1}}}-\frac{1}{\sqrt{v_t}}) + 4A_2. \nonumber
\end{eqnarray}
\end{lemma}
The first lemma bounds $\sum_t \mathbb{E}\|\nabla f(\hat{w}_t)\|^2 (\frac{1}{\sqrt{v_{t-1}}}-\frac{1}{\sqrt{v_t}})$ with $\sum_t \frac{\|\nabla f(\hat{w}_t)\|^2}{\sqrt{v_{t-1}}}$ and some other terms, while the second lemma reversely bounds $\sum_t \frac{\|\nabla f(\hat{w}_t)\|^2}{\sqrt{v_{t-1}}}$ with $\sum_t \mathbb{E}\|\nabla f(\hat{w}_t)\|^2 (\frac{1}{\sqrt{v_{t-1}}}-\frac{1}{\sqrt{v_t}})$. Combining these two lemmas helps us bound $T_2$. Substituting the result together with $T_1$ and $T_3$ into (\ref{theo1_1}), we obtain the bound of $\sum_{t=1}^T \mathbb{E} \|\nabla f(x_t)\|^2 / \sqrt{v_{t-1}}$ successfully. Then we establish the relationship between $v_t$ and $u_t$ as the following:
\begin{lemma}\label{lemma3}
If $f(x)$ in Algorithm \ref{alg1} satisfies Assumption \ref{assu1}, 
    \begin{equation}
    \|\nabla f(w_t,\xi_t)\|^2\!\leq\!(\frac{\rho L}{\epsilon} + 1)\|\nabla f(x_t,\xi_t)\|^2,
    v_t\!\leq\!(\frac{\rho L}{\epsilon}+1) u_t \nonumber
\end{equation}
\end{lemma}
Substituting Lemma \ref{lemma3} into the bound of $\sum_{t=1}^T \mathbb{E} \frac{\|\nabla f(x_t)\|^2}{\sqrt{v_{t-1}}}$ yields the upper bound of $\sum_{t=1}^T \mathbb{E} \frac{\|\nabla f(x_t)\|^2}{\sqrt{u_{t-1}}}$ as follow:
\begin{equation}\label{theo1_3}
    \sum_{t=1}^T \mathbb{E} \frac{\|\nabla f(x_t)\|^2}{\sqrt{u_{t-1}}} \leq A_3 + 2A_4 \ln \mathbb{E} \sqrt{\hat{u}_T} + 2A_5 \ln \mathbb{E} \sqrt{u_T}.
\end{equation}
Secondly, we inherit the intermediate result in \cite{wang2023convergence} and combine it with (\ref{theo1_3}) to obtain that 
\begin{eqnarray}\label{theo1_4}
    \mathbb{E} \sqrt{u_T} &\leq& \sqrt{2D_0 T + \epsilon^2} + 2D_1 \sum_{t=1}^T \mathbb{E} \frac{\|\nabla f(x_t)\|^2}{\sqrt{u_{t-1}}} \nonumber \\
    &\leq& \sqrt{2D_0 T + \epsilon^2} + 2D_1 A_3 + 4D_1 A_4 \ln \mathbb{E} \sqrt{\hat{u}_T} \nonumber \\
    &&+ 4D_1 A_5 \ln \mathbb{E} \sqrt{u_T}.
\end{eqnarray}
Further, we propose the following lemma:
\begin{lemma}\label{main_bound}
    For any $A, B, x>0$, if it satisfies that $x \leq A + B \ln x$, then $x$ is upper bounded by
    \begin{equation}
        x \leq 2A+2B \ln (B+e). \nonumber
    \end{equation}
\end{lemma}
Applying this Lemma on (\ref{theo1_4}) yields that 
\begin{eqnarray}\label{theo1_5}
    \mathbb{E} \sqrt{u_T} &\leq& 2\sqrt{2D_0 T + \epsilon^2} + 4D_1 A_3 + 8D_1 A_4 \ln \mathbb{E} \sqrt{\hat{u}_T} \nonumber \\
    &&+ 8D_1 A_5 \ln (4D_1 A_5+e).
\end{eqnarray}
According to the Cauchy's Inequality, we could obtain that
\begin{equation}
    \frac{\bigg(\mathbb{E}\sqrt{\sum_{t=1}^T \|\nabla f(x_t)\|^2}\bigg)^2}{\mathbb{E} \sqrt{u_T}} \leq \sum_{t=1}^T \mathbb{E} \frac{\|\nabla f(x_t)\|^2}{\sqrt{u_{t-1}}}. \nonumber
\end{equation}
Substituting (\ref{theo1_3}) and (\ref{theo1_5}) into the above inequality yields that
\begin{align}
    (\mathbb{E}\sqrt{\hat{u}_T})^2 &\leq A_6(A_3+2A_5\ln A_6) + (2A_4 A_6+16D_1A_4A_5 \nonumber \\
    &+8D_1A_4(A_3+2A_5\ln A_6)) \ln \mathbb{E}  \sqrt{\hat{u}_T} \nonumber \\
    &+ 8D_1 A_4 (2A_4+\frac{16D_1 A_4 A_5}{A_6})(\ln \mathbb{E}\sqrt{\hat{u}_T})^2 \nonumber
\end{align}
To solve this inequality, we establish another lemma:
\begin{lemma}\label{main_bound2}
    For any $A, B, C, x>0$, if it satisfies that $x^2 \leq A + B \ln x + C(\ln x)^2$, then $x$ is upper bounded by
    \begin{equation}
        x \leq \sqrt{2A+2B\ln(B+e)+64C^2+1}. \nonumber
    \end{equation}
\end{lemma}
We need to emphasize that the order of the coefficients in the bound above is crucial for the final convergence rate. By this lemma, we obtain the upper bound of $\mathbb{E}\sqrt{\hat{u}_T}$, and then $\sum_{t=1}^T \mathbb{E}\|\nabla f(x_t)\|$, thus complete the proof.
\end{IEEEproof}

\textbf{Discussion.} 
ASAM \cite{kwon2021asam} is proposed to alleviate the insensitivity of SAM to weight scaling. Though the element-wise operator is performed on the gradients to achieve sharpness adaptivity, the perturbation radius does not consider historical gradients like common adaptive optimizers (Adagrad-Norm, Adagrad and Adam). AdaSAM \cite{sun2024adasam} does not introduce adaptivity to the perturbation radius like LightSAM. Additionally, its theoretical analysis relies on a strong assumption, i.e. the stochastic gradient is upper bounded.

\subsection{LightSAM-II (AdaGrad)}

In LightSAM-II (see Algorithm \ref{alg2}), we adopt the AdaGrad-type learning rate to perturb and update model weights. LightSAM-II adopts the coordinate-wise learning rates to scale the perturbation step and gradient descent step, which can better utilize the historical gradients and achieve a stable convergence. Thus, compared to Algorithm \ref{alg1}, the initialized $u_0$ and $v_0$ become vectors with each element equal to $\epsilon^2$, and the multiplication and division become element-wise between vectors.

\begin{algorithm}[t]
  \caption{LightSAM-II (AdaGrad)}\label{alg2}
  \begin{algorithmic}[1]
  \REQUIRE Initial values $x_0$, $u_0=v_0=\mathbf{\epsilon^2}$, perturbation radius $\rho$, learning rate $\eta$.
  \FOR{$t = 1, ..., T$}
    \STATE Sample a minibatch $\xi_t$ from the dataset;
    \STATE Compute stochastic gradient $s_t = \nabla f(x_t, \xi_t)$;
    \STATE $u_t = u_{t-1} + s_t \odot s_t$;
    \STATE $w_t = x_t + \rho \frac{1}{\sqrt{u_t}} \odot s_t$;
    \STATE Compute stochastic gradient $g_t = \nabla f(w_t, \xi_t)$;
    \STATE $v_t = v_{t-1} + g_t \odot g_t$;
    \STATE Update weights $x_{t+1} = x_t -\eta \frac{1}{\sqrt{v_t}} \odot g_t$;
  \ENDFOR
  \end{algorithmic}
\end{algorithm}

To prove the convergence of LightSAM-II with coordinate-wise learning rates, we require the following coordinate-wise smoothness and affine noise variance assumptions. 

\begin{assumption}[Coordinate-wise $L$-smoothness]\label{assu3}
For $ \forall l \in [d]$, $f(x)$ is differentiable and satisfies:
    \begin{equation}
        | \nabla f(x,\xi)_l - \nabla f(y,\xi)_l | \leq L |x_l - y_l|, \forall x, y \in R^d.
    \nonumber
    \end{equation}
\end{assumption}

\begin{assumption}[Coordinate-wise affine noise variance]\label{assu4}
    There exist positive constants $D_0$ and $D_1$:
    \begin{equation}
        \mathbb{E}^{|\mathcal{F}_t}\nabla f(x, \xi)_l^2 \leq D_0 + D_1 \nabla f(x)_l^2, \forall x\in R^d, \forall l \in [d].
    \nonumber
    \end{equation}
\end{assumption}

Assumption \ref{assu3} is adopted in \cite{richtarik2014iteration,das2024towards} and necessary here since the inequality relationship between $\nabla f(x_t,\xi_t)$ and $\nabla f(w_t,\xi_t)$ is established coordinate-wisely. Assumption \ref{assu4} is commonly used in adaptive optimization works which do not need to assume the bounded gradient \cite{crawshaw2022robustness,wang2023convergence,wang2023closing}.  

\begin{theorem}\label{theo2}
If $f(x)$ in Algorithm \ref{alg2} satisfies Assumptions \ref{assu3} and \ref{assu4}, for any perturbation radius $\rho$ and learning rate $\eta > 0$, we have that 
\begin{eqnarray}
   \makebox[7.8cm][l]{\ensuremath{\displaystyle \sum_{t=1}^T \mathbb{E} \|\nabla f(x_t)\|_1 \leq  T^{\frac{1}{2}}d^{\frac{1}{2}}[2B_6(B_3+2A_5\ln B_6)}} \nonumber \\
   \makebox[7.8cm][l]{\ensuremath{\displaystyle+ 2B_7\ln(B_7+e) + 4096D_1^2 A_4^2 (2A_4+\frac{16D_1 A_4 A_5}{B_6})^2+1]^{\frac{1}{2}}}} \nonumber
\end{eqnarray}
Here, we denote constants $\Bar{w}_1, B_1, B_2, B_3, B_5$ as following
\begin{equation}
    \makebox[8.5cm][l]{\ensuremath{\displaystyle D_2 = \max\{1,4D_1,32(1+\sqrt{D_1})D_1 \rho\sqrt{\rho L +\epsilon}/ (\eta \sqrt{\epsilon})\},}} \nonumber 
\end{equation}
\begin{equation}
    \makebox[8.5cm][l]{\ensuremath{\displaystyle \Bar{w}_1 = x_1 + \rho \frac{1}{\sqrt{\mathbf{\epsilon^2}+\nabla f(x_1)^{\odot 2} }}\odot \nabla f(x_1),}} \nonumber
\end{equation}
\begin{equation}
    \makebox[8.5cm][l]{\ensuremath{\displaystyle B_1 = \frac{\|\nabla f(\Bar{w}_1)\|^2}{\epsilon} + \frac{4(1+2 D_2)dL^2}{\epsilon} (\eta^2 - 4 \rho^2 \ln \epsilon),}} \nonumber
\end{equation}
\begin{eqnarray}
    \makebox[7.8cm][l]{\ensuremath{\displaystyle B_2 = d(2\rho^2L + \frac{\rho D_0}{2\epsilon \sqrt{D_1}} + \rho \|\nabla f(x_1)\|_1 +\frac{(D_0+4\rho^2 L^2)\eta}{\epsilon}}} \nonumber \\
    \makebox[7.8cm][l]{\ensuremath{\displaystyle - (\frac{12\rho^2 \eta L^2}{\epsilon}+\!\eta\!+\!\rho\!+ (1\!+\!\rho L)(2\eta^2 L\!+\!8\rho^2 L)) \ln \epsilon)\!+\!f(w_1),}} \nonumber
\end{eqnarray}
\begin{eqnarray}
    \makebox[7.8cm][l]{\ensuremath{\displaystyle B_3 =\!\sqrt{\frac{\rho L}{\epsilon}\!+\!1}[\frac{4f(x_1)}{\eta}\!+\!\frac{8d(D_0\!+\!4\rho^2 L^2)}{\epsilon}\!+\!16 D_1 B_1\!+\!\frac{8B_2}{\eta}}} \nonumber \\
    \makebox[7.8cm][l]{\ensuremath{\displaystyle - (\frac{80 \rho^2 L^2}{\epsilon} + 4\eta L)d \ln \epsilon + (4\eta L (3\!+\!\rho L)+8)d \ln (1+\frac{\rho L}{\epsilon})],}} \nonumber
\end{eqnarray}
\begin{equation}
    \makebox[8.5cm][l]{\ensuremath{\displaystyle B_6 = 2\sqrt{2D_0 T + \epsilon^2} + 4D_1 B_3 + 8D_1 A_5 \ln (4D_1 A_5+e),}} \nonumber
\end{equation}
\begin{equation}
\makebox[8.5cm][l]{\ensuremath{\displaystyle B_7 = 2A_4 B_6+16D_1A_4A_5+8D_1A_4(B_3+2A_5\ln B_6),}} \nonumber
\end{equation}
\ $D_2$, $A_4$ and $A_5$ are the same as Theorem \ref{theo1}.
\end{theorem}

\begin{corollary}\label{coro2}
From Theorem \ref{theo2}, we can obtain the following convergence rate for Algorithm \ref{alg2}
\begin{equation}
    \frac{1}{T} \sum_{t=1}^T \mathbb{E} \|\nabla f(x_t)\|_1 \leq O\bigg(\frac{\ln T}{T^{1/4}}\bigg). \nonumber
\end{equation}
\end{corollary}

\begin{algorithm}[t]
  \caption{LightSAM-III (Adam)}\label{alg3}
  \begin{algorithmic}[1]
  \REQUIRE Initial values $x_0$, $u_0=v_0=\mathbf{\epsilon^2}$, perturbation radius $\rho$, learning rate $\eta$, coefficients $\beta_1, \beta_2$.
  \FOR{$t = 1, ..., T$}
    \STATE Sample a minibatch $\xi_t$ from the dataset;
    \STATE Compute stochastic gradient $s_t = \nabla f(x_t, \xi_t)$;
    \STATE $r_t = \beta_1 r_{t-1} + (1-\beta_1) s_t$;
    \STATE $u_t = \beta_2 u_{t-1} + (1-\beta_2) s_t \odot s_t$;
    \STATE $w_t = x_t + \rho \frac{1}{\sqrt{u_t}} \odot r_t$;
    \STATE Compute stochastic gradient $g_t = \nabla f(w_t, \xi_t)$;
    \STATE $m_t = \beta_1 m_{t-1} + (1-\beta_1) g_t$;
    \STATE $v_t = \beta_2 v_{t-1} + (1-\beta_2) g_t \odot g_t$;
    \STATE Update weights $x_{t+1} = x_t - \eta \frac{1}{\sqrt{v_t}} \odot m_t$;
  \ENDFOR
  \end{algorithmic}
\end{algorithm}

\subsection{LightSAM-III (Adam)}
Adam \cite{kingma2014adam} is another popular optimizer for deep learning, especially in Transformer-based models, which replaces the gradient aggregation step for estimating adaptive learning rate in AdaGrad with an exponential moving average step by introducing two additional momentum parameters $(\beta_1,\beta_2)$ and achieves a stable and fast convergence. 
In this section, we integrate the Adam-type learning rate to update the parameters $(\rho,\eta)$ in SAM, which yields LightSAM-III (Adam), as shown in Algorithm \ref{alg3}. We also perform a theoretical analysis for LightSAM-III and obtain the following result:
\begin{theorem}\label{theo3}
    If $f(x)$ in Algorithm \ref{alg3} satisfies Assumptions \ref{assu3} and \ref{assu4}, and $0\leq \beta_1 < \sqrt{\beta_2} < 1$, $\beta_2 \geq \frac{\sqrt{D_3^2+4D_3}-D_3}{2}$. Then, for any $\beta_2$, perturbation radius $\rho$ and learning rate $\eta$ satisfy that $1-\beta_2 = O(T^{-1})$, $\eta=O(T^{-\frac{1}{2}})$, $\rho=O(T^{-\frac{1}{2}})$, we have the convergence rate
\begin{equation}
    \frac{1}{T} \sum_{t=1}^T \mathbb{E} \|\nabla f(x_t)\|_1 \leq O\bigg(\frac{\ln T}{T^{1/4}}\bigg), \nonumber
\end{equation}
where the constant $D_3$ satisfies that \\[10pt]
$D_3 = \max\{4\sqrt{\beta_2}, \frac{256\sqrt{\beta_2}D_1}{\beta_2-\beta_1^2}, \frac{2048\sqrt{C_1} D_1 \rho}{(1-\beta_1)(1-\frac{\beta_1^2}{\beta_2})\sqrt{\beta_2}\eta}(1+\frac{2D_1}{\beta_2-\beta_1^2})\}.$
\end{theorem}

\begin{remark}
    In our result, the dependence on $T$ of $\beta_2$ and learning rates are consistent with those in Adam \cite{wang2023closing}. The only difference between the two results is that the constraint on $\beta_2$ is a little more complex and stricter. This is unsurprising since LightSAM has double Adam-type steps, which makes the conditions to ensure convergence more complex. 
\end{remark}
We also present a proof sketch, and the complete derivations are referred to the Supplementary Material.

\begin{IEEEproof}
As a preliminary, we define the sequences
\begin{eqnarray}
    \check{r}_t &=& \beta_1 \check{r}_{t-1} + (1-\beta_1)\nabla f(x_t), \nonumber \\
    \check{u}_t &=& \beta_2 \check{u}_{t-1} + (1-\beta_2)\nabla f(x_t) \odot \nabla f(x_t), \nonumber \\
    \check{w}_t &=& x_t + \rho \frac{1}{\sqrt{\check{u}_t+\mathbf{\epsilon^2}}} \odot \check{r}_t \nonumber
\end{eqnarray}    
to remove the randomness of the stochastic gradient in the perturbation step. We in addition define the following sequences
\begin{equation}
    \makebox[8.5cm][l]{\ensuremath{\displaystyle p_t = \frac{w_t-\frac{\beta_1}{\sqrt{\beta_2}}w_{t-1}}{1-\frac{\beta_1}{\sqrt{\beta_2}}},  \quad\check{p}_t = \frac{\check{w}_t-\frac{\beta_1}{\sqrt{\beta_2}}\check{w}_{t-1}}{1-\frac{\beta_1}{\sqrt{\beta_2}}},}} \nonumber 
\end{equation}
\begin{equation}
    \makebox[8.5cm][l]{\ensuremath{\displaystyle q_t = \frac{x_t-\frac{\beta_1}{\sqrt{\beta_2}}x_{t-1}}{1-\frac{\beta_1}{\sqrt{\beta_2}}},}} \nonumber 
\end{equation}
\begin{equation}
    \makebox[8.5cm][l]{\ensuremath{\displaystyle \Tilde{u}_{t} = \beta_2 u_{t-1} + (1-\beta_2)D_0 \mathds{1}_d, \Tilde{v}_{t} = \beta_2 v_{t-1} + (1-\beta_2)D_0 \mathds{1}_d}} \nonumber 
\end{equation}
for the Adam-type algorithm. Firstly, we aim to bound $\sum_{t=1}^T \sum_{l=1}^d \mathbb{E} \frac{\nabla f(x_t)_l^2}{\sqrt{\Tilde{v}_{t,l}}}$. Applying the $L$-smoothness of $f(\cdot)$ on $\{q_t\}$, we have that
\begin{align}\label{theo3_1}
    &\mathbb{E}[f(q_{T+1})] - f(q_1) \leq \underbrace{- \frac{\eta(1-\beta_1)}{1-\frac{\beta_1} {\sqrt{\beta_2}}}\sum_{t=1}^T \sum_{l=1}^d \mathbb{E}\frac{\nabla f(x_t)_l g_{t,l}}{\sqrt{\Tilde{v}_{t,l}}}}_{T_1} \nonumber \\
    &\underbrace{- \frac{\eta}{1-\frac{\beta_1}{\sqrt{\beta_2}}} \sum_{t=1}^T \sum_{l=1}^d \mathbb{E}\nabla f(x_t)_l m_{t,l}(\frac{1}{\sqrt{v_{t,l}}}-\frac{1}{\sqrt{\Tilde{v}_{t,l}}})}_{T_2} \nonumber \\    
    &+ \underbrace{\frac{\eta \beta_1}{1-\frac{\beta_1}{\sqrt{\beta_2}}} \sum_{t=1}^T \sum_{l=1}^d \mathbb{E}\nabla f(x_t)_l m_{t-1,l}(\frac{1}{\sqrt{\beta_2 v_{t-1,l}}}-\frac{1}{\sqrt{\Tilde{v}_{t,l}}})}_{T_3}  \nonumber \\
    &+ \underbrace{\sum_{t=1}^T \mathbb{E}\langle \nabla f(q_t)\!-\!\nabla f(x_t),q_{t+1}\!-\!q_t\rangle\!+\!\frac{L}{2}\sum_{t=1}^T \mathbb{E}\|q_{t+1}\!-\!q_t\|^2}_{T_4}.
\end{align}
In the above inequality, $T_1$ pluses $T_3$ and $T_4$ could be bounded by the linear combination of the following four targeted or manageable terms $\sum_{t=1}^T \sum_{l=1}^d \mathbb{E} \{-\frac{\nabla f(x_t)_l^2}{\sqrt{\Tilde{v}_{t,l}}}, \frac{r_{t,l}^2}{u_{t,l}}, \frac{\check{r}_{t,l}^2}{\check{u}_{t,l}}, \frac{m_{t,l}^2}{v_{t,l}}\}$. Hence, we focus on $T_2$. We obtain that $\nabla f(x_t)_l m_{t,l}(\frac{1}{\sqrt{\Tilde{v}_{t,l}}}-\frac{1}{\sqrt{v_{t,l}}}) 
\leq |\nabla f(x_t)_l| |m_{t,l}| \frac{(1-\beta_2)(g_{t,l}^2+D_0)}{\sqrt{v_{t,l}} \sqrt{\Tilde{v}_{t,l}}(\sqrt{v_{t,l}} +\sqrt{\Tilde{v}_{t,l}})}$. The first term with respect to $g_{t,l}^2$ of two terms in the RHS is the key point to deal with. By separating the variance from $g_{t,l}^2$, we have that $\sum_{t=1}^T \sum_{l=1}^d \mathbb{E} |\nabla f(x_t)_l| |m_{t,l}| \frac{(1-\beta_2)g_{t,l}^2}{\sqrt{v_{t,l}} \sqrt{\Tilde{v}_{t,l}}(\sqrt{v_{t,l}} +\sqrt{\Tilde{v}_{t,l}})}$ could be bounded by the linear combination of $\sum_{t=1}^T \sum_{l=1}^d \mathbb{E} \{\frac{\nabla f(x_t)_l^2}{\sqrt{\Tilde{v}_{t,l}}}, \frac{g_{t,l}^2}{v_{t,l}}, (\frac{1}{\sqrt{\beta_2 \Tilde{v}_{t,l}}} - \frac{1}{\sqrt{\Tilde{v}_{t+1,l}}})\nabla f(\check{w}_t)_l^2\}$, where the last term is not easy to bound. In \cite{wang2023closing}, the similar term $\sum_{t=1}^T (\frac{1}{\sqrt{\beta_2 \Tilde{v}_{t,l}}} - \frac{1}{\sqrt{\Tilde{v}_{t+1,l}}}) G_{t,l}^2$ could be bounded by the targeted term $\sum_{t=1}^T \frac{G_{t,l}^2}{\Tilde{v}_{t,l}}$. Similar to the proof of Theorem 1, we could not follow this process since the misalignment between $x_t$ and $\check{w}_t$.

Instead, firstly, since $\|x\|^2 - \|y\|^2 \leq 2\|x-y\|\|x\| + \|x-y\|^2$, we bound $\sum_{t=1}^T \sum_{l=1}^d \mathbb{E} (\frac{1}{\sqrt{\beta_2 \Tilde{v}_{t,l}}} - \frac{1}{\sqrt{\Tilde{v}_{t+1,l}}})\nabla f(\check{w}_t)_l^2$ with the linear combination of $\sum_{t=1}^T \sum_{l=1}^d \mathbb{E} \frac{\nabla f(\check{w}_t)_l^2}{\sqrt{\Tilde{v}_{t,l}}}$ and some other manageable terms. In reverse, applying the $L$-smoothness of $f(\cdot)$ on the sequence $\{p_t\}$ bounds $\sum_{t=1}^T \sum_{l=1}^d \mathbb{E} \frac{\nabla f(\check{w}_t)_l^2}{\sqrt{\Tilde{v}_{t,l}}}$ with $\sum_{t=1}^T \sum_{l=1}^d \mathbb{E} (\frac{1}{\sqrt{\beta_2 \Tilde{v}_{t,l}}} - \frac{1}{\sqrt{\Tilde{v}_{t+1,l}}})\nabla f(\check{w}_t)_l^2$ and some terms. By carefully setting the coefficients, we combine these two results and obtain the bound of $\sum_{t=1}^T \sum_{l=1}^d \mathbb{E} (\frac{1}{\sqrt{\beta_2 \Tilde{v}_{t,l}}} - \frac{1}{\sqrt{\Tilde{v}_{t+1,l}}})\nabla f(\check{w}_t)_l^2$. Substituting this result and other manageable terms into (\ref{theo3_1}), and rearranging the inequality yields that
\begin{equation}\label{theo3_2}
    \sum_{t=1}^T \sum_{l=1}^d \mathbb{E} \frac{\nabla f(x_t)_l^2}{\sqrt{\Tilde{u}_{t,l}}} \leq  C_4 +C_5 \sum_{l=1}^d \mathbb{E} \ln \check{u}_{T,l} + C_6 \sum_{l=1}^d \mathbb{E} \ln u_{T,l}.
\end{equation}
Then, following the intermediate result in \cite{wang2023closing}, we have
\begin{align}\label{theo3_3}
    &\sum_{t=1}^{T+1} \sum_{l=1}^d \mathbb{E}\sqrt{\Tilde{u}_{t,l}} 
    \leq  \frac{3(1+\sqrt{\beta_2})D_1}{\sqrt{\beta_2}}(C_4 +2dC_5 \ln \mathbb{E} \sum_{l=1}^d  \sqrt{\check{u}_{T,l}} \nonumber \\
    &- 2d C_5 \ln d) + (T+1)d\sqrt{D_0+\epsilon^2} + \frac{6(1+\sqrt{\beta_2})d D_1 C_6}{\beta_2} \nonumber \\
    &\times(\ln \sum_{t=1}^{T+1} \sum_{l=1}^d \mathbb{E}\sqrt{\Tilde{u}_{t,l}} - \ln d).
\end{align}
Combining (\ref{theo3_2}), (\ref{theo3_3}) and Lemmas \ref{main_bound}, \ref{main_bound2}, we finally obtain that
\begin{eqnarray}
    &&\sum_{t=1}^T \mathbb{E} \|\nabla f(x_t)\|_1 \nonumber \\
    &&\leq \sqrt{2C_8 + 2C_9 \ln (C_9+e) + 64 (1-\beta_2)C_{10}^2 + 1}, \nonumber
\end{eqnarray}
where $C_4$-$C_6$ and $C_8$-$C_{10}$ are all constants. Substituting $1-\beta_2 = O(T^{-1})$, $\eta=O(T^{-\frac{1}{2}})$, $\rho=O(T^{-\frac{1}{2}})$ into the formula of these constants yields $C_8 = O(T^{\frac{3}{2}} ln T)$, $C_9 = O(T^{\frac{3}{2}})$, $C_{10} = O(T)$. As a consequence, we obtain
\begin{equation}
    \frac{1}{T} \sum_{t=1}^T \mathbb{E} \|\nabla f(x_t)\|_1 \leq O\bigg(\frac{\ln T}{T^{1/4}}\bigg), \nonumber
\end{equation}
\end{IEEEproof}

\begin{table*}[t]
\renewcommand{\arraystretch}{1.3}
    \caption{Best test accuracies (\%) on MNIST dataset.}
    \label{mnist}
    \centering
    \begin{tabular}{>{\centering\arraybackslash}p{1.2cm}>{\centering\arraybackslash}p{1.2cm}>{\centering\arraybackslash}p{1.2cm}>{\centering\arraybackslash}p{1.2cm}>{\centering\arraybackslash}p{1.2cm}>{\centering\arraybackslash}p{1.2cm}>{\centering\arraybackslash}p{1.2cm}>{\centering\arraybackslash}p{1.2cm}>{\centering\arraybackslash}p{1.4cm}}
    \hline
    Method & SGD & SAM & ASAM & AdaSAM & AdaGrad & L-SAM-II & Adam & L-SAM-III \\
    \hline
    3-layer &  98.21 & 98.29 & 98.24 & 98.57 & 98.26 & 98.33 & 98.57 & \textbf{98.59} \\
    \hline 
    LeNet & 99.29  & 99.37 & 99.48 & 99.48 & 99.25 & 99.31 & 99.41 & \textbf{99.49} \\
    \hline 
    \end{tabular}
\end{table*}

\begin{table}[t]
\renewcommand{\arraystretch}{1.3}
    \caption{Average best test accuracies (\%) of LightSAM on MNIST dataset under different hyper-parameters.}
    \label{mnist_avg}
    \centering
    \begin{tabular}{>{\centering\arraybackslash}p{1.6cm}>{\centering\arraybackslash}p{1.6cm}>{\centering\arraybackslash}p{1.6cm}>{\centering\arraybackslash}p{1.6cm}}
    \hline
    \multicolumn{2}{c}{3-layer NN} & \multicolumn{2}{c}{LeNet}  \\
    \cmidrule(r){1-2} \cmidrule(r){3-4}
    LightSAM-II&LightSAM-III&LightSAM-II&LightSAM-III \\
    \hline
    98.29$\pm$0.03&98.56$\pm$0.03&99.25$\pm$0.07&99.41$\pm$0.07 \\
    \hline 
    \end{tabular}
\end{table}

\section{Experiments}
In this section, we conduct experiments to show the effectiveness of our proposed algorithm. Experiments include the CV task conducted on MNIST and Imagenet datasets and the NLP task conducted on the GLUE benchmark. The main goal of this paper is to validate that parameter-agnostic SAM optimizers without parameter tuning can achieve comparable performance with the carefully handcrafted learning rate schedule. All the experiments are conducted on a machine with NVIDIA 3090 GPUs.

\subsection{MNIST dataset}

\textbf{Implementation detail.} 
We first conduct the image classification task on the MNIST dataset. A simple 3-layer neural network and LeNet \cite{lecun1998gradient} are adopted as the training models. We select SGD, AdaGrad, Adam, SAM, ASAM, AdaSAM, LightSAM-II and LightSAM-III as the baselines. The initial learning rate $\eta$ is set to 0.1 for SGD, SAM, and ASAM, 0.01 for AdaGrad and LightSAM-II, 0.001 for AdaSAM and LightSAM-III. The perturbation radius $\rho$ is set to 0.05 and 0.5 for SAM and ASAM respectively as suggested in \cite{foret2020sharpness,kwon2021asam}, 0.1 for AdaSAM, 0.001 for LightSAM-II and III. We run all methods for 30 epochs. The learning rate is decayed two times by a factor of 0.2. 

\textbf{Results on MNIST.} We summarize the best test accuracies of all baselines in the two experimental settings in Table \ref{mnist}. For each model, LightSAM-II achieves higher accuracy than AdaGrad, meanwhile, LightSAM-III achieves higher accuracy than Adam. This result indicates that parameter perturbation could improve the test accuracies of adaptive optimizers, the same as the phenomenon in the comparison between SAM and SGD. Additionally, LightSAM-II performs better than SAM in 3-layer neural network and LightSAM-III performs better than SAM in two cases, which is consistent with the advantage of Adam over SGD.  

In the theoretical analysis, we prove that LightSAM could converge without tuning any hyper-parameters. Thus, in each experimental case, we scale the adopted $\rho$ and $\eta$ respectively, as a result obtaining four hyper-parameter settings $(\rho, 2\rho)*(\eta, 2\eta)$. We run LightSAM under these four settings and list the average result in Table \ref{mnist_avg}. We can find that the average best accuracies are still higher than some baselines. The low standard deviations show the insensitivities of LightSAM to hyper-parameters.

\subsection{Fine-tuning on Imagenet dataset}
\textbf{Implementation detail.} We conduct the fine-tuning task on transformer models. Specifically, we fine-tune the ViT-Tiny and ViT-Small \cite{touvron2021training} on the Imagenet-1k dataset for 10 epochs from the checkpoints pre-trained on the Imagenet-21k dataset. The utilized checkpoints are open-sourced on \texttt{Huggingface}. We select SGD, Adam, SAM, ASAM, AdaSAM and LightSAM-III as the baselines. Following \cite{foret2020sharpness,kwon2021asam} and common choices, we set the learning rate as 0.1 for SGD, SAM and ASAM, 1e-4 for Adam, AdaSAM and LightSAM. And the perturbation radius is set as 0.05 for SAM, 0.5 for ASAM, 0.01 for AdaSAM and 1e-4 for LightSAM. Weight decay is not utilized for all optimizers. Momentum is set as 0.9 for all SGD optimizers.

\begin{table}[t]
\renewcommand{\arraystretch}{1.3}
    \caption{Best test accuracies (\%) on Imagenet dataset after fine-tuning the ViT models.}
    \label{vit}
    \centering
    \begin{tabular}{lcc}
    \hline
    Algorithms & ViT-Tiny & ViT-Small \\
    \hline
    SGD & 45.59 & 63.78 \\
    \hline
    Adam & 60.82 & 77.10 \\
    \hline
    SAM & 60.10 & 74.27 \\
    \hline
    ASAM & 59.95 & 74.12 \\
    \hline
    AdaSAM & 64.43 & 78.02 \\
    \hline
    LightSAM & \textbf{64.58} & \textbf{78.09} \\
    \hline
    \end{tabular}
\end{table}

\begin{figure*}[t]
\centering
\includegraphics[scale=0.21]{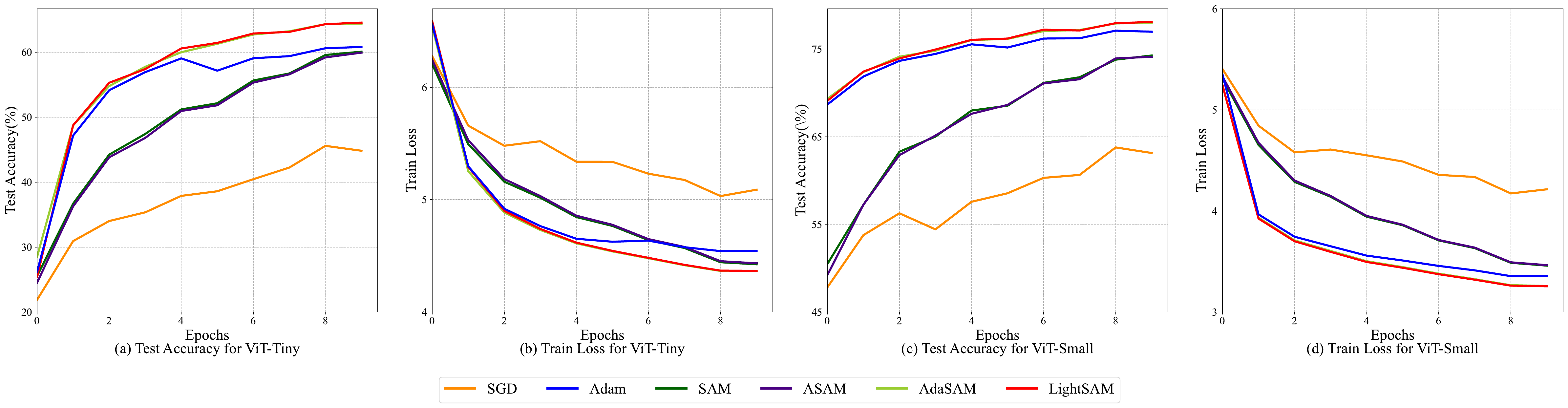}
\caption{Experimental results of fine-tuning ViT models on Imagenet. (a): Test accuracy w.r.t. epochs for ViT-Tiny; (b): Train loss w.r.t. epochs for ViT-Tiny; (c): Test accuracy w.r.t. epochs for ViT-Small; (d) Train loss w.r.t. epochs for ViT-Small.}
\label{fig}
\end{figure*}

 \begin{table*}[t]
  \renewcommand{\arraystretch}{1.3}
     \caption{Best test accuracies (\%) of SAM-type algorithms on ViT-Small model under different parameter settings.}
     \label{rob}
     \centering
     \begin{threeparttable}
     \begin{tabular}{>{\centering\arraybackslash}p{1cm}>{\centering\arraybackslash}p{1cm}>{\centering\arraybackslash}p{1cm}>{\centering\arraybackslash}p{1cm}>{\centering\arraybackslash}p{1cm}>{\centering\arraybackslash}p{1cm}>{\centering\arraybackslash}p{1cm}>{\centering\arraybackslash}p{1cm}>{\centering\arraybackslash}p{1cm}>{\centering\arraybackslash}p{2cm}}
     \hline
     \multicolumn{4}{l}{SAM $(\eta,\rho)$=(0.1,0.05)} & & & & & & Avg.\\
     \hline
     75.68 & 75.81 & 76.02 & 73.89 & 74.27 & 74.11 & 71.58 & 71.56 & 71.86 & $73.86 \pm 1.72$ \\
     \hline
     \multicolumn{4}{l}{ASAM $(\eta,\rho)$=(0.1,0.5)} & & & & & & Avg.\\
     \hline
     75.72 & 75.71 & 75.78 & 73.88 & 74.12 & 74.22 & 71.45 & -\tnote{a} & - & $74.41 \pm 1.44 $ \\
     \hline 
     \multicolumn{4}{l}{AdaSAM $(\eta,\rho)$=(1e-4,0.01)} & & & & & & Avg.\\
     \hline
     78.00 & 77.98 & 78.02 & 78.00 & 78.02 & 77.99 & 77.16 & 77.10 & 77.04 & $77.70 \pm 0.43$ \\
     \hline
     \multicolumn{4}{l}{LightSAM $(\eta,\rho)$=(1e-4,1e-4)} & & & & & & Avg.\\
     \hline
     77.97 & 78.00 & 78.04 & 77.99 & 78.09 & 78.06 & 77.29 & 77.10 & 77.27 & \textbf{77.76 $\pm$ 0.38} \\
     \hline
     \end{tabular}
     \begin{tablenotes}
     \small
     \item[a] ``-'' represents the divergence of the algorithm.
     \end{tablenotes}
     \end{threeparttable}
 \end{table*}

 \begin{table*}[t]
\renewcommand{\arraystretch}{1.25}
    \caption{Experimental performances on GLUE benchmark after fine-tuning.}
    \label{glue}
    \centering
    \begin{tabular}{>{\centering\arraybackslash}p{2cm}|>{\centering\arraybackslash}p{1cm}>{\centering\arraybackslash}p{1cm}>{\centering\arraybackslash}p{1cm}>{\centering\arraybackslash}p{1cm}>{\centering\arraybackslash}p{1cm}>{\centering\arraybackslash}p{1cm}>{\centering\arraybackslash}p{1cm}>{\centering\arraybackslash}p{1cm}|>{\centering\arraybackslash}p{1cm}}
    \hline
    Algorithms & CoLA & STS-B & MRPC & RTE & SST2 & MNLI & QNLI & QQP &Avg.\\
    \hline
    SGD & 59.39 & 87.85 & 91.65 & 76.53 & 93.69 & 86.33 & 89.27 & 91.49 & 84.53\\
    Adam & 62.08 & 90.77 & 92.50 & 78.70 & 94.84 & 87.42 & 92.82 & 91.90 & 86.38\\
    SAM & 61.71 & 89.25 & 92.01 & 79.42 & 94.27 & 86.42 & 89.53 & 91.38 & 85.50\\
    ASAM & 63.51 & 89.14 & 92.48 & 78.70 & 93.81 & 86.44 & 90.17 & 91.57 & 85.73\\
    AdaSAM & 62.11 & 90.55 & 93.12 & 80.14 & 95.30 & 87.57 & \textbf{93.10} & 92.01 & 86.74\\
    LightSAM & \textbf{63.77} & \textbf{90.77} & \textbf{93.33} & \textbf{81.95} & \textbf{95.41} & \textbf{87.63} & 92.92 & \textbf{92.04} & \textbf{87.23}\\
    \hline 
    \end{tabular}
\end{table*}

\textbf{Results on Imagenet.} In Table \ref{vit}, we list the best test accuracies of all baselines. Firstly, we could observe that the optimizers which adopt adaptive learning rate in the model update step (Adam, AdaSAM and LightSAM) perform better than those adopt constant learning rate (SGD, SAM and ASAM). This is in line with the advantage of adaptive optimizers over SGD on transformer based models \cite{zhang2020adaptive}. Secondly, the optimizers utilize the weight perturbation step achieve higher test accuracies than the corresponding base optimizers (SAM and ASAM over SGD, AdaSAM and LightSAM over Adam), which presents the positive effect of weight perturbation in improving test performance. Finally, AdaSAM and LightSAM achieve comparable accuracies while LightSAM is still ahead of AdaSAM, thus the adaptive perturbation radius in LightSAM is comparable with the carefully handcrafted constant radius. We also plot the curves of training loss and test accuracy of fine-tuning ViT models in Figure \ref{fig}. From the figure, we could observe that regardless of the test accuracy and training loss, AdaSAM and our proposed algorithm LightSAM are ahead of other baselines obviously throughout the whole process, and LightSAM has a little advantage over AdaSAM. Though this performance is partly due to the power of Adam in Transformer-based model, it still illustrates the capability of adopting adaptive hyper-parameters in the SAM optimizer.

\textbf{Sensitivity to hyper-parameters.} For several SAM-type algorithms, we enrich the experiment on a wide range of parameter values. For one baseline, denote the selected hyper-parameters in the above subsection as $\eta$ and $\rho$, we take nine combinations of parameters $(0.5\eta,\eta,2\eta)*(0.5\rho,\rho,2\rho)$ to show its sensitivity to these parameters. 
The results are shown in Table \ref{rob}. The first nine columns record the best accuracy of one set of parameter values and the last column represents the mean and standard deviation (also for Table \ref{glue_rob} below). 

We could observe that SAM which does not have any adaptive modules has the highest deviation. ASAM does not converge in two settings with a large learning rate and performs worse than AdaSAM which adopts the commonly used adaptive learning rate. Under various parameter selections, our proposed algorithm achieves the highest mean accuracy and lowest deviation, which is in line with the "parameter-agnostic" property of LightSAM and indicates its insensitivity to hyper-parameters including both the learning rate and perturbation radius.

\subsection{Fine-tuning on GLUE task}
\textbf{Implementation detail.} We also consider training the language models. We fine-tune the RoBERTa model \cite{liu2019roberta} for 8 downstream tasks in the GLUE benchmark. The learning rate is set to 1e-2 for SGD, SAM and ASAM, 1e-5 for Adam, AdaSAM and LightSAM-III. The perturbation radius is set to 5e-3 for SAM and 1e-5 for LightSAM-III to maintain its ratio to learning rate same as the ViT experiment, 1e-2 for AdaSAM as adopted in \cite{sun2024adasam}, 1e-2 for ASAM after careful tuning. The batch size is set to 32 for all tasks except 16 for QNLI. We run all algorithms for 20 epochs. 

 \begin{table*}[t]
 \renewcommand{\arraystretch}{1.3}
     \caption{Performances of SAM-type algorithms under different parameter settings for STS-B.}
     \label{glue_rob}
     \centering
     \begin{tabular}{>{\centering\arraybackslash}p{1cm}>{\centering\arraybackslash}p{1cm}>{\centering\arraybackslash}p{1cm}>{\centering\arraybackslash}p{1cm}>{\centering\arraybackslash}p{1cm}>{\centering\arraybackslash}p{1cm}>{\centering\arraybackslash}p{1cm}>{\centering\arraybackslash}p{1cm}>{\centering\arraybackslash}p{1cm}>{\centering\arraybackslash}p{2cm}}
     \hline
     \multicolumn{4}{l}{SAM $(\eta,\rho)$=(0.01,5e-3)} & & & & & & Avg.\\
     \hline
     - & 89.53 & 87.87 & 89.31 & 89.25 & 89.19 & - & - & - & $88.97 \pm 0.79$ \\
     \hline
     \multicolumn{4}{l}{ASAM $(\eta,\rho)$=(0.01,0.01)} & & & & & & Avg.\\
     \hline
     85.74 & 83.26 & - & 88.99 & 89.14 & 88.58 & - & - & - & $87.14 \pm 2.57 $ \\
     \hline 
     \multicolumn{4}{l}{AdaSAM $(\eta,\rho)$=(1e-5,0.01)} & & & & & & Avg.\\
     \hline
     90.20 & 90.29 & 90.27 & 90.54 & 90.55 & 90.48 & 90.86 & 91.01 & 90.92 & $90.57 \pm 0.30$ \\
     \hline
     \multicolumn{4}{l}{LightSAM $(\eta,\rho)$=(1e-5,1e-5)} & & & & & & Avg.\\
     \hline
     90.42 & 90.31 & 90.39 & 90.79 & 90.77 & 90.69 & 90.97 & 91.09 & 91.05 & \textbf{90.72 $\pm$ 0.29} \\
     \hline
     \end{tabular}
 \end{table*}

\textbf{Results and parameter sensitivity on GLUE.} We list the experimental results in Table \ref{glue}. We report the Matthew's correlation for CoLA, Pearson correlation for STS-B, F1 score for MRPC, averaged accuracy for MNLI, and accuracy for other tasks. Similar to the experiment on Imagenet, the algorithms that use the adaptive learning rate in the gradient descent step achieve the highest three scores, and each algorithm that adopts the extra perturbation step is ahead of its version that does not. LightSAM performs best in seven tasks except the QNLI dataset, which again verifies its excellence in practice.

Samely, we conduct the experiments under nine sets of parameters $(0.5\eta,\eta,2\eta)*(0.5\rho,\rho,2\rho)$ on the STS-B task to test the sensitivity to the hyper-parameters for SAM-type optimizers, where $\eta$ and $\rho$ are the parameters set above. The results in Table \ref{glue_rob} show the strong sensitivity of SAM and ASAM in this task as they fail to converge under four hyper-parameter settings. AdaSAM and LightSAM could converge to great solutions, which demonstrates the efficacy of the adaptive learning rate in the high stability. Between them, our proposed method has an advantage over AdaSAM, again indicating its insensitivity to the perturbation radius.

\section{Conclusion}

In this paper, we propose an algorithm LightSAM for non-convex optimization. LightSAM sets the perturbation radius and learning rate adaptively through adopting Adagrad-Norm, Adagrad, and Adam, respectively. We make a solid theoretical analysis for our proposed algorithm and observe that it converges with the $\mathbb{E}\|\nabla f(x_t)\| \leq O(\ln T / T^{1/4})$ rate without requiring the gradient bounded assumption. Particularly, our result does not require perturbation radius and learning rate satisfying any conditions, realizing parameter-agnostic optimizers. Finally, we conduct experiments in several computer vision tasks. The superiority of LightSAM to other baselines and the insensitivity to hyper-parameters are verified. Thus, we prove the potential of our work in reducing the necessity of parameter tuning from both theory and experiments.


%

\appendices


\ifCLASSOPTIONcaptionsoff
  \newpage
\fi



\bibliographystyle{IEEEtran}
\bibliography{references}
%

%



\newpage
\onecolumn
\section{Useful Inequalities}
We first show some inequalities which are useful for our analysis.
\setcounter{lemma}{5}
\setcounter{theorem}{3}
\begin{lemma}\label{ine}
    (Lemma 10 in \cite{wang2023convergence}) Consider sequence $\{a_t\}_{t=0}^T$ with $a_0 > 0, a_i \geq 0$ for $i >0$, then we have
    \begin{equation}
    \sum_{t=1}^T \frac{a_t}{\sum_{\tau=0}^t a_{\tau}} \leq \ln \sum_{t=0}^T a_t - \ln a_0, \quad \sum_{t=1}^T \frac{a_t}{(\sum_{\tau=0}^t a_{\tau})^{3/2}} \leq \frac{2}{\sqrt{a_0}},\nonumber
    \end{equation}
    \begin{equation}
    \sum_{t=1}^T \frac{a_t}{(\sum_{\tau=0}^{t-1} a_{\tau})^{1/2}((\sum_{\tau=0}^{t-1} a_{\tau})^{1/2}+(\sum_{\tau=0}^t a_{\tau})^{1/2})^2} \leq \frac{1}{\sqrt{a_0}}. \nonumber
    \end{equation}
\end{lemma}

\begin{lemma}\label{ine2}
    (Lemmas 4 and 5 in \cite{wang2023closing}) Assume the constants $0<\beta_1^2<\beta_2<1$. Consider sequences $\{a_t\}_{t=1}^T$, $b_n = \beta_2 b_{n-1} + (1-\beta_2) a_n^2$ with $b_0>0$, $c_n = \beta_2 c_{n-1} + (1-\beta_2) a_n$ with $c_n=0$, then we have
    \begin{equation}
    \sum_{t=1}^T \frac{a_n^2}{b_n} \leq \frac{1}{1-\beta_2}(\ln \frac{b_T}{b_0}-T \ln \beta_2), \quad \sum_{t=1}^T \frac{c_n^2}{b_n} \leq \frac{(1-\beta_1)^2}{(1-\frac{\beta_1}{\sqrt{\beta_2}})^2 (1-\beta_2)}(\ln \frac{b_T}{b_0}-T \ln \beta_2). \nonumber
    \end{equation}
\end{lemma}

\begin{lemma}\label{bound}
    (Restatement of Lemma 4) For any $A, B, x>0$, if it satisfies that $x \leq A + B \ln x$, then $x$ is upper bounded by
    \begin{equation}
        x \leq 2A+2B \ln (B+e). \nonumber
    \end{equation}
\end{lemma}
\begin{IEEEproof}
We turn to prove the contrapositive: if $x> 2A+2B \ln (B+e)$, then $x > A + B \ln x$. Define $g(x,A,B) = x - A-B \ln x$, first we have
\begin{equation}
    \frac{\partial g(x,A,B)}{\partial x} = 1-\frac{B}{x} >1-\frac{B}{2A+2B} >0. \nonumber
\end{equation}
Thus, $g(x,A,B) > g(2A+2B \ln (B+e),A,B)=A+2B \ln (B+e)-B\ln(2A+2B \ln (B+e))$.
Then, we have
\begin{equation}
    \frac{\partial [A+2B \ln (B+e)-B\ln(2A+2B \ln (B+e))]}{\partial A} = 1-\frac{2B}{2A+2B\ln(B+e)} > 0. \nonumber
\end{equation}
Thus, we have $g(x,A,B)>g(2A+2B \ln (B+e),0,B)=B(\ln(B+e)^2-\ln (2B\ln(B+e))$.
Consider $h(B) = (B+e)^2-2B \ln(B+e)$, since $h'(B)=2(B+e)-2\ln(B+e)-\frac{2B}{B+e}>2((B+e)-\ln(B+e)-1)>0$, $h(B)\geq h(0)>0$. Therefore, $(B+e)^2 > 2B\ln(B+e)$, and finally, $g(x,A,B)>0$.
\end{IEEEproof}

\begin{lemma}\label{bound2}
    (Restatement of Lemma 5) For any $A, B, C, x>0$, if it satisfies that $x^2 \leq A + B \ln x + C(\ln x)^2$, then $x$ is upper bounded by
    \begin{equation}
        x \leq \sqrt{2A+2B\ln(B+e)+64C^2+1}. \nonumber
    \end{equation}
\end{lemma}
\begin{IEEEproof}
Similarly, we turn to prove that if $x>\sqrt{2A+2B\ln(B+e)+64C^2+1}$, then $x^2>A + B \ln x + C(\ln x)^2$. Define $g(x,A,B,C)=x^2-A-B\ln x-C(\ln x)^2$, first we have
\begin{equation}
    \frac{\partial g(x,A,B,C)}{\partial x} = 2x-\frac{B}{x}-\frac{2C \ln x}{x} > 2x-\frac{B}{x}-2C. \nonumber
\end{equation}
Since $x>\sqrt{B+C^2} >\frac{2C+\sqrt{4C^2+8B}}{4}$, we have $2x^2-2Cx-B>0$. Thus,  
\begin{eqnarray}
g(x,A,B,C) &>& A +2B\ln(B+e)+64C^2 +1- \frac{B}{2} \ln(2A+2B\ln(B+e)+64C^2+1) \nonumber \\
&&- \frac{C}{4}(\ln (2A+2B\ln(B+e)+64C^2+1))^2 \nonumber
\end{eqnarray}
Denoting the right hand of the inequality as $h(A,B,C)$. Then, we have
\begin{eqnarray}
    \frac{\partial h(A,B,C)}{\partial A} &=& 1-\frac{B}{2A+2B\ln(B+e)+64C^2+1} - \frac{C \ln (2A+2B\ln(B+e)+64C^2+1)}{2A+2B\ln(B+e)+64C^2+1} \nonumber \\
    &>& 1- \frac{1}{2} - \frac{2C \sqrt{2A+2B\ln(B+e)+64C^2+1}}{2A+2B\ln(B+e)+64C^2+1} > 0. \nonumber
\end{eqnarray}
Thus, we have 
\begin{eqnarray}
h(A,B,C)&>&h(0,B,C) = 2B\ln(B+e)+64C^2+1 - \frac{B}{2} \ln(2B\ln(B+e)+64C^2+1) \nonumber \\
&&- \frac{C}{4}(\ln (2B\ln(B+e)+64C^2+1))^2 \nonumber \\
&>& 2B\ln(B+e)+64C^2+1 - \frac{B}{2} \ln(2B\ln(B+e)+64C^2+1) - 4C \sqrt{2B\ln(B+e)+64C^2+1}, \nonumber
\end{eqnarray}
where the last inequality holds because $(\ln a)^2 = (4\ln a^{1/4})^2 < (4a^{1/4})^2$ for $a>1$. Since $\ln(a+b)\leq \ln a+\frac{b}{a}$, we have
\begin{eqnarray}
    h(A,B,C)&\geq& 2B\ln(B\!+\!e)\!+\!64C^2\!+\!1 - \frac{B}{2}(\ln (2B\ln(B+e))+\frac{64C^2+1}{2B\ln(B\!+\!e)}) - 4C \sqrt{2B\ln(B+e)+64C^2+1} \nonumber \\
    &>& B\ln(B+e)+48C^2+\frac{3}{4}- 4C\sqrt{2B\ln(B\!+\!e)\!+\!64C^2\!+\!1} >0 \nonumber
\end{eqnarray}
where the second inequality comes from that $B\ln(B+e) > \frac{B}{2}\ln(2B\ln(B+e))$ in the proof of the last lemma. Thus, we complete the proof.
\end{IEEEproof}

\section{Proof of Theorem 1 and 2}
We first define the virtual sequence $\{\hat{w}_t\}$ as 
\begin{equation}
    \hat{u}_0 = u_0, \quad \hat{u}_t = \hat{u}_{t-1} + \|\nabla f(x_t)\|^2, \quad\hat{w}_t = x_t + \rho \frac{\nabla f(x_t)}{\sqrt{\hat{u}_t}}
\end{equation}
\begin{lemma}\label{var}
    If $f(x)$ in Algorithm 1 satisfies Assumptions 1 and 2, we have that
    \begin{equation}
        \mathbb{E}^{\mathcal{F}_t} \|\nabla f(w_t,\xi_t)\|^2 \leq 2D_0+8\rho^2 L^2 + 2D_1\|\nabla f(\hat{w}_t)\|^2. \nonumber
    \end{equation}
\end{lemma}
\begin{IEEEproof}
\begin{eqnarray}
\mathbb{E}^{\mathcal{F}_t} \|\nabla f(w_t,\xi_t)\|^2 &=&  \mathbb{E}^{\mathcal{F}_t} \|\nabla f(w_t,\xi_t) - \nabla f(\hat{w}_t,\xi_t) + \nabla f(\hat{w}_t,\xi_t)\|^2 \nonumber \\
&\stackrel{(a)}{\leq}& 2L^2 \mathbb{E}^{\mathcal{F}_t} \|w_t-\hat{w}_t\|^2 + 2 \mathbb{E}^{\mathcal{F}_t}\|\nabla f(\hat{w}_t,\xi_t)\|^2 \nonumber \\
&\stackrel{(b)}{\leq}& 2\rho^2 L^2 \mathbb{E}^{\mathcal{F}_t} \|\frac{s_t}{\sqrt{u_{t-1}+\|s_t\|^2}}-\frac{\nabla f(x_t)}{\sqrt{u_{t-1}+\|\nabla f(x_t)\|^2}}\|^2 + 2(D_0+D_1\|\nabla f(\hat{w}_t)\|^2) \nonumber \\
&\leq& (2D_0+8\rho^2 L^2) + 2D_1\|\nabla f(\hat{w}_t)\|^2, \nonumber
\end{eqnarray}
where (a) and (b) come from Assumptions 1 and 2 respectively.
\end{IEEEproof}

\begin{lemma}\label{relemma1}
(Restatement of Lemma 1) If $f(x)$ in Algorithm 1 satisfies Assumptions 1 and 2, we have that
\begin{eqnarray}
\sum_{t=1}^T \mathbb{E}\|\nabla f(\hat{w}_t)\|^2 (\frac{1}{\sqrt{v_{t-1}}}-\frac{1}{\sqrt{v_t}}) &\leq& A_1 - \mathbb{E}\frac{\|\nabla f(\hat{w}_T)\|^2}{\sqrt{v_T}} + \frac{1}{2 D_2}\sum_{t=1}^T \mathbb{E} \frac{\|\nabla f(\hat{w}_t)\|^2}{\sqrt{v_{t-1}}} + \frac{8(1+2 D_2) \rho^2 L^2}{\epsilon} \ln \hat{u}_T \nonumber
\end{eqnarray}
where $D_2 = \max\{1,4D_1,\sqrt{\frac{\rho L}{\epsilon}+1}\frac{32(1+\sqrt{D_1})D_1 \rho}{\eta}\}, \ \hat{w}_1 = x_1+\rho\frac{\nabla f(x_1)}{\sqrt{\epsilon^2+\|\nabla f(x_1)\|^2}}, \ A_1 = \frac{\|\nabla f(\hat{w}_1)\|^2}{\epsilon} + \frac{4(1+2 D_2)L^2}{\epsilon} (\eta^2 - 4 \rho^2 \ln \epsilon)$. \nonumber
\end{lemma}
\begin{IEEEproof}
For two vectors $x$ and $y$, consider that $\langle x-y,y \rangle \leq \langle x-y,x\rangle$, we could further infer that $\langle x-y,y \rangle \leq \|x-y\|\|x\|$. And further $2\langle x,y \rangle - 2\|y\|^2 \leq 2 \|x-y\|\|x\|$. Finally, we obtain
\begin{equation}
    \|x\|^2 - \|y\|^2 \leq 2\|x-y\|\|x\| + \|x\|^2 + \|y\|^2 - 2\langle x,y \rangle = 2\|x-y\|\|x\| + \|x-y\|^2 \nonumber
\end{equation}
Based on this and Assumption 1, we have that
\begin{eqnarray}\label{lemma1_1}
\|\nabla f(\hat{w}_t)\|^2 (\frac{1}{\sqrt{v_{t-1}}}-\frac{1}{\sqrt{v_t}}) \leq [\frac{\|\nabla f(\hat{w}_{t-1})\|^2}{\sqrt{v_{t-1}}} - \frac{\|\nabla f(\hat{w}_t)\|^2}{\sqrt{v_t}}] + \frac{2L\|\hat{w}_t-\hat{w}_{t-1}\|\|\nabla f(\hat{w}_t)\|+L^2 \|\hat{w}_t-\hat{w}_{t-1}\|^2}{\sqrt{v_{t-1}}} 
\end{eqnarray}
Consider
\begin{equation}\label{lemma1_2}
    \|\hat{w}_t-\hat{w}_{t-1}\| \leq \eta \frac{\|\nabla f(w_{t-1}, \xi_{t-1})\|}{\sqrt{v_{t-1}}} + \rho \|\frac{\nabla f(x_t)}{\sqrt{\hat{u}_t}} - \frac{\nabla f(x_{t-1})}{\sqrt{\hat{u}_{t-1}}}\|
\end{equation}
\begin{equation}\label{lemma1_3}
    \|\hat{w}_t-\hat{w}_{t-1}\|^2 \leq 2\eta^2 \frac{\|\nabla f(w_{t-1}, \xi_{t-1})\|^2}{v_{t-1}} + 2\rho^2 \|\frac{\nabla f(x_t)}{\sqrt{\hat{u}_t}} - \frac{\nabla f(x_{t-1})}{\sqrt{\hat{u}_{t-1}}}\|^2
\end{equation}
Substituting (\ref{lemma1_2}) and (\ref{lemma1_3}) into (\ref{lemma1_1}) and summing the result over $t \in \{2,...,T\}$ yields that
\begin{eqnarray}\label{lemma1_4}
&&\sum_{t=2}^T \mathbb{E}\|\nabla f(\hat{w}_t)\|^2 (\frac{1}{\sqrt{v_{t-1}}}-\frac{1}{\sqrt{v_t}}) \nonumber \\
&\leq& 2\eta L \sum_{t=2}^{T} \mathbb{E} \frac{\|\nabla f(w_{t-1}, \xi_{t-1})\| \|\nabla f(\hat{w}_t)\|}{v_{t-1}} +2\rho L \sum_{t=2}^{T} \mathbb{E} \frac{\|\frac{\nabla f(x_t)}{\sqrt{\hat{u}_t}} - \frac{\nabla f(x_{t-1})}{\sqrt{\hat{u}_{t-1}}}\| \|\nabla f(\hat{w}_t)\|}{\sqrt{v_{t-1}}}\nonumber \\
&& \mathbb{E}[\frac{\|\nabla f(\hat{w}_{1})\|^2}{\sqrt{v_1}} - \frac{\|\nabla f(\hat{w}_T)\|^2}{\sqrt{v_T}}] + 2\eta^2 L^2 \sum_{t=2}^{T} \mathbb{E} \frac{\|\nabla f(w_{t-1}, \xi_{t-1})\|^2}{v_{t-1}^{3/2}} + 2\rho^2 L^2 \sum_{t=2}^{T} \mathbb{E} \frac{\|\frac{\nabla f(x_t)}{\sqrt{\hat{u}_t}} - \frac{\nabla f(x_{t-1})}{\sqrt{\hat{u}_{t-1}}}\|^2}{\sqrt{v_{t-1}}} 
\end{eqnarray}
In the RHS of (\ref{lemma1_4})
\begin{eqnarray}
    2\eta L \sum_{t=2}^{T} \mathbb{E} \frac{\|\nabla f(w_{t-1}, \xi_{t-1})\| \|\nabla f(\hat{w}_t)\|}{v_{t-1}} \leq 4 D_2 \eta^2 L^2 \sum_{t=2}^{T} \mathbb{E} \frac{\|\nabla f(w_{t-1}, \xi_{t-1})\|^2}{v_{t-1}^{3/2}} + \frac{1}{4 D_2}\sum_{t=2}^T \mathbb{E} \frac{\|\nabla f(\hat{w}_t)\|^2}{\sqrt{v_{t-1}}} \nonumber
\end{eqnarray}
\begin{eqnarray}
    2 \rho L \sum_{t=2}^{T} \mathbb{E}\frac{\|\frac{\nabla f(x_t)}{\sqrt{\hat{u}_t}} - \frac{\nabla f(x_{t-1})}{\sqrt{\hat{u}_{t-1}}}\| \|\nabla f(\hat{w}_t)\|}{\sqrt{v_{t-1}}} 
    &\leq& 4 D_2 \rho^2 L^2 \sum_{t=2}^{T} \mathbb{E} \frac{\|\frac{\nabla f(x_t)}{\sqrt{\hat{u}_t}} - \frac{\nabla f(x_{t-1})}{\sqrt{\hat{u}_{t-1}}}\|^2}{\sqrt{v_{t-1}}} + \frac{1}{4 D_2}\sum_{t=2}^T \mathbb{E} \frac{\|\nabla f(\hat{w}_t)\|^2}{\sqrt{v_{t-1}}} \nonumber
\end{eqnarray}
Thus, we have
\begin{eqnarray}
&&\sum_{t=2}^T \mathbb{E}\|\nabla f(\hat{w}_t)\|^2 (\frac{1}{\sqrt{v_{t-1}}}-\frac{1}{\sqrt{v_t}}) \nonumber \\
&\leq& \mathbb{E}[\frac{\|\nabla f(\hat{w}_{1})\|^2}{\sqrt{v_1}} - \frac{\|\nabla f(\hat{w}_T)\|^2}{\sqrt{v_T}}] + 2(1+2 D_2)\eta^2 L^2 \sum_{t=2}^{T} \mathbb{E}\frac{\|\nabla f(w_{t-1}, \xi_{t-1})\|^2}{v_{t-1}^{3/2}} \nonumber \\
&& + \frac{1}{2 D_2}\sum_{t=2}^T \mathbb{E}\frac{\|\nabla f(\hat{w}_t)\|^2}{\sqrt{v_{t-1}}} + 2(1+2 D_2)\rho^2 L^2 \sum_{t=2}^{T} \mathbb{E}\frac{\|\frac{\nabla f(x_t)}{\sqrt{\hat{u}_t}} - \frac{\nabla f(x_{t-1})}{\sqrt{\hat{u}_{t-1}}}\|^2}{\sqrt{v_{t-1}}} \nonumber \\
&\stackrel{(a)}{\leq}& \mathbb{E}[\frac{\|\nabla f(\hat{w}_{1})\|^2}{\sqrt{v_1}} - \frac{\|\nabla f(\hat{w}_T)\|^2}{\sqrt{v_T}}] + \frac{1}{2 D_2}\sum_{t=2}^T \mathbb{E}\frac{\|\nabla f(\hat{w}_t)\|^2}{\sqrt{v_{t-1}}} + 4(1+2 D_2)\eta^2L^2 \frac{1}{\epsilon}+ \frac{8(1+2 D_2) \rho^2 L^2}{\epsilon} \sum_{t=1}^T \mathbb{E} \frac{\|\nabla f(x_t)\|^2}{\hat{u}_t} \nonumber \\
&\stackrel{(b)}{\leq}& \mathbb{E}[\frac{\|\nabla f(\hat{w}_{1})\|^2}{\sqrt{v_1}} - \frac{\|\nabla f(\hat{w}_T)\|^2}{\sqrt{v_T}}] + \frac{1}{2 D_2}\sum_{t=1}^T \mathbb{E}\frac{\|\nabla f(\hat{w}_t)\|^2}{\sqrt{v_{t-1}}} + 4(1+2 D_2)\eta^2L^2 \frac{1}{\epsilon}+ \frac{8(1+2 D_2) \rho^2 L^2}{\epsilon} (\ln \hat{u}_T - \ln u_0) \nonumber
\end{eqnarray}
where (a) and (b) come from Lemma \ref{ine}. Finally, we have
\begin{eqnarray}
&&\sum_{t=1}^T \mathbb{E}\|\nabla f(\hat{w}_t)\|^2 (\frac{1}{\sqrt{v_{t-1}}}\!-\!\frac{1}{\sqrt{v_t}}) \nonumber \\
&\leq& \mathbb{E}[\frac{\|\nabla f(\hat{w}_{1})\|^2}{\sqrt{v_0}}\!-\!\frac{\|\nabla f(\hat{w}_T)\|^2}{\sqrt{v_T}}] + \frac{1}{2 D_2}\sum_{t=1}^T \mathbb{E}\frac{\|\nabla f(\hat{w}_t)\|^2}{\sqrt{v_{t-1}}} + 4(1+2 D_2)\eta^2L^2 \frac{1}{\epsilon}+\frac{8(1+2 D_2) \rho^2 L^2}{\epsilon} (\ln \hat{u}_T - \ln u_0) \nonumber \\
&\leq& \frac{\|\nabla f(\hat{w}_{1})\|^2}{\epsilon} - \mathbb{E}\frac{\|\nabla f(\hat{w}_T)\|^2}{\sqrt{v_T}} + \frac{4(1+2 D_2)L^2}{\epsilon} (\eta^2 - 4 \rho^2 \ln \epsilon) + \frac{1}{2 D_2}\sum_{t=1}^T \mathbb{E} \frac{\|\nabla f(\hat{w}_t)\|^2}{\sqrt{v_{t-1}}} + \frac{8(1+2 D_2) \rho^2 L^2}{\epsilon} \ln \hat{u}_T \nonumber
\end{eqnarray}
\end{IEEEproof}

\begin{lemma}\label{relemma2}
(Restatement of Lemma 2) If $f(x)$ in Algorithm 1 satisfies Assumptions 1 and 2, we have that
\begin{eqnarray}
&&\eta \sum_{t=1}^{T-1} \frac{\|\nabla f(\hat{w}_t)\|^2}{\sqrt{v_{t-1}}} \leq  4D_1 \eta \sum_{t=1}^{T-1} \mathbb{E}\|\nabla f(\hat{w}_t)\|^2 (\frac{1}{\sqrt{v_{t-1}}}-\frac{1}{\sqrt{v_t}}) + 4A_2 + (2\eta+4\eta^2 L +4\rho \eta^2 L^2)\mathbb{E} \ln v_T \nonumber \\
&& +4(1+\sqrt{D_1})\rho \mathbb{E} \frac{\|\nabla f(x_T)\|^2}{\sqrt{u_{T-1}}}+ (\frac{12\rho^2 \eta L^2}{\epsilon}+16\rho^2 L+16\rho^3 L^2 + 2\rho) \mathbb{E} \ln u_T + \frac{12\rho^2 \eta L^2}{\epsilon} \mathbb{E}\ln \hat{u}_T\nonumber
\end{eqnarray}
where \\
$A_2 = f(w_1)+2\rho^2L + \frac{\rho D_0}{2\epsilon \sqrt{D_1}} + \rho \|\nabla f(x_1)\| +\frac{(D_0+4\rho^2 L^2)\eta}{\epsilon}- (\frac{12\rho^2 \eta L^2}{\epsilon}+\eta + \rho + (1+\rho L)(2\eta^2 L+8\rho^2 L)) \ln \epsilon $.
\end{lemma}
\begin{IEEEproof}
According to the $L$-smoothness of $f(x)$, we have
\begin{eqnarray}\label{lemma2_1}
\mathbb{E}^{|\mathcal{F}_t}[f(w_{t+1})] &\leq& \mathbb{E}^{|\mathcal{F}_t}[f(w_t)] + \mathbb{E}^{|\mathcal{F}_t} \langle \nabla f(w_t), w_{t+1}-w_t \rangle + \frac{L}{2}\mathbb{E}^{|\mathcal{F}_t}\|w_{t+1}-w_t\|^2 \nonumber \\
&=& \mathbb{E}^{|\mathcal{F}_t}[f(w_t)] + \eta \mathbb{E}^{|\mathcal{F}_t} \langle \nabla f(w_t), -\frac{\nabla f(w_t, \xi_t)}{\sqrt{v_t}}\rangle \nonumber \\
&&+ \mathbb{E}^{|\mathcal{F}_t}\langle \nabla f(w_t), \rho(\frac{\nabla f(x_{t+1}, \xi_{t+1})}{\sqrt{u_{t+1}}} - \frac{\nabla f(x_t, \xi_t)}{\sqrt{u_t}}) \rangle + \frac{L}{2}\mathbb{E}^{|\mathcal{F}_t}\|w_{t+1}-w_t\|^2
\end{eqnarray}
Since
\begin{eqnarray}\label{lemma2_2}
&&\mathbb{E}^{|\mathcal{F}_t} \langle \nabla f(w_t), -\frac{\nabla f(w_t, \xi_t)}{\sqrt{v_t}}\rangle \nonumber \\
&=& \mathbb{E}^{|\mathcal{F}_t} \langle \nabla f(\hat{w}_t), -\frac{\nabla f(w_t, \xi_t)}{\sqrt{v_t}}\rangle + \mathbb{E}^{|\mathcal{F}_t} \langle \nabla f(w_t)-\nabla f(\hat{w}_t), -\frac{\nabla f(w_t, \xi_t)}{\sqrt{v_t}}\rangle\nonumber \\
&=& -\mathbb{E}^{|\mathcal{F}_t} \langle \nabla f(\hat{w}_t), \frac{\nabla f(\hat{w}_t, \xi_t)}{\sqrt{v_{t-1}}} \rangle+ \mathbb{E}^{|\mathcal{F}_t} \langle \nabla f(\hat{w}_t), \frac{\nabla f(\hat{w}_t,\xi_t)-\nabla f(w_t,\xi_t)}{\sqrt{v_{t-1}}}\rangle+\mathbb{E}^{|\mathcal{F}_t} \langle \nabla f(\hat{w}_t), \nabla f(w_t, \xi_t)(\frac{1}{\sqrt{v_{t-1}}}-\frac{1}{\sqrt{v_t}}) \rangle \nonumber \\
&&+\mathbb{E}^{|\mathcal{F}_t} \langle \nabla f(w_t)-\nabla f(\hat{w}_t), -\frac{\nabla f(w_t, \xi_t)}{\sqrt{v_t}} \rangle\nonumber \\
&\leq& -\mathbb{E}^{|\mathcal{F}_t} \frac{\|\nabla f(\hat{w}_t)\|^2}{\sqrt{v_{t-1}}} + \frac{1}{4}\mathbb{E}^{|\mathcal{F}_t} \frac{\|\nabla f(\hat{w}_t)\|^2}{\sqrt{v_{t-1}}} + L^2 \mathbb{E}^{|\mathcal{F}_t}\frac{\|\hat{w}_t-w_t\|^2}{\sqrt{v_{t-1}}} + \mathbb{E}^{|\mathcal{F}_t}\langle \nabla f(\hat{w}_t), \nabla f(w_t, \xi_t)(\frac{1}{\sqrt{v_{t-1}}}-\frac{1}{\sqrt{v_t}}) \rangle \nonumber \\
&&+\frac{L^2}{2} \mathbb{E}^{|\mathcal{F}_t} \|w_t-\hat{w}_t\|^2 + \mathbb{E}^{|\mathcal{F}_t}\frac{\|g_t\|^2}{2v_t} \nonumber \\
&\leq& -\frac{3}{4}\mathbb{E}^{|\mathcal{F}_t} \frac{\|\nabla f(\hat{w}_t)\|^2}{\sqrt{v_{t-1}}} + \frac{(2+\epsilon)\rho^2 L^2}{\epsilon}\mathbb{E}^{|\mathcal{F}_t}(\frac{\|s_t\|^2}{u_t}+\frac{\|\nabla f(w_t)\|^2}{\hat{u}_t}) + \frac{1}{2}\mathbb{E}^{|\mathcal{F}_t} \frac{\|g_t\|^2}{v_t}\nonumber \\
&&+ \mathbb{E}^{|\mathcal{F}_t}\langle \nabla f(\hat{w}_t), \nabla f(w_t, \xi_t)(\frac{1}{\sqrt{v_{t-1}}}-\frac{1}{\sqrt{v_t}})\rangle
\end{eqnarray}
Substituting (\ref{lemma2_2}) into (\ref{lemma2_1}) yields that
\begin{eqnarray}\label{lemma2_3}
\frac{3\eta}{4} \mathbb{E}^{|\mathcal{F}_t} \frac{\|\nabla f(\hat{w}_t)\|^2}{\sqrt{v_{t-1}}}
&\leq& \mathbb{E}^{|\mathcal{F}_t}[f(w_t)] - \mathbb{E}^{|\mathcal{F}_t}[f(w_{t+1})] + \frac{(2+\epsilon)\rho^2 \eta L^2}{\epsilon}\mathbb{E}^{|\mathcal{F}_t}(\frac{\|s_t\|^2}{u_t}+\frac{\|\nabla f(w_t)\|^2}{\hat{u}_t}) + \frac{\eta}{2} \mathbb{E}^{|\mathcal{F}_t} \frac{\|g_t\|^2}{v_t}\nonumber \\
&&+\eta \mathbb{E}^{|\mathcal{F}_t}  \langle \nabla f(\hat{w}_t), \nabla f(w_t, \xi_t)(\frac{1}{\sqrt{v_{t-1}}}-\frac{1}{\sqrt{v_t}}) \rangle + \mathbb{E}^{|\mathcal{F}_t}\langle \nabla f(w_t), \rho(\frac{\nabla f(x_{t+1},\xi_{t+1})}{\sqrt{u_{t+1}}} - \frac{\nabla f(x_t,\xi_t)}{\sqrt{u_t}}) \rangle \nonumber \\
&&+ \frac{L}{2}\mathbb{E}^{|\mathcal{F}_t}\|w_{t+1}-w_t\|^2 
\end{eqnarray}
For the terms on the RHS of (\ref{lemma2_3}), first we have
\begin{eqnarray}\label{lemma2_4}
&& \mathbb{E}^{|\mathcal{F}_t}\langle \nabla f(\hat{w}_t), \nabla f(w_t, \xi_t)(\frac{1}{\sqrt{v_{t-1}}}-\frac{1}{\sqrt{v_t}}) \rangle \nonumber \\
&\stackrel{(a)}{\leq}& \mathbb{E}^{|\mathcal{F}_t} \frac{\|\nabla f(\hat{w}_t)\|\|\nabla f(w_t, \xi_t)\|^3}{\sqrt{v_{t-1}} \sqrt{v_t}(\sqrt{v_{t-1}}+\sqrt{v_t})} \stackrel{(b)}{\leq} \mathbb{E}^{|\mathcal{F}_t} \frac{\|\nabla f(\hat{w}_t)\|\|\nabla f(w_t, \xi_t)\|^2}{\sqrt{v_{t-1}} (\sqrt{v_{t-1}}+\sqrt{v_t})} \nonumber \\
&=& \frac{\|\nabla f(\hat{w}_t)\|}{v_{t-1}^{1/4}} \mathbb{E}^{|\mathcal{F}_t}\frac{\|\nabla f(w_t,\xi_t)\|^2}{v_{t-1}^{1/4}(\sqrt{v_{t-1}}+\sqrt{v_t})} \nonumber \\
&\leq& \frac{\|\nabla f(\hat{w}_t)\|^2}{2\sqrt{v_{t-1}}} + \frac{1}{2} (\mathbb{E}^{|\mathcal{F}_t}\frac{\|\nabla f(w_t,\xi_t)\|^2}{v_{t-1}^{1/4}(\sqrt{v_{t-1}}+\sqrt{v_t})})^2 \nonumber \\
&\stackrel{(c)}{\leq}& \frac{\|\nabla f(\hat{w}_t)\|^2}{2\sqrt{v_{t-1}}} + \frac{1}{2\sqrt{v_{t-1}}}(\mathbb{E}^{|\mathcal{F}_t} \|\nabla f(w_t, \xi_t)\|^2)(\mathbb{E}^{|\mathcal{F}_t}\frac{\|\nabla f(w_t, \xi_t)\|^2}{(\sqrt{v_{t-1}}+\sqrt{v_t})^2}) \nonumber \\
&\stackrel{(d)}{\leq}& \frac{\|\nabla f(\hat{w}_t)\|^2}{2\sqrt{v_{t-1}}} + \frac{1}{\sqrt{v_{t-1}}}(D_0 +4\rho^2 L^2 + D_1 \|\nabla f(\hat{w}_t)\|^2)(\mathbb{E}^{|\mathcal{F}_t}\frac{\|\nabla f(w_t, \xi_t)\|^2}{(\sqrt{v_{t-1}}+\sqrt{v_t})^2}) \nonumber \\
&\stackrel{(e)}{\leq}& \frac{\|\nabla f(\hat{w}_t)\|^2}{2\sqrt{v_{t-1}}} + (D_0 +4\rho^2 L^2) \mathbb{E}^{|\mathcal{F}_t} \frac{\|\nabla f(w_t, \xi_t)\|^2}{\sqrt{v_{t-1}}(\sqrt{v_{t-1}}+\sqrt{v_t})^2} + D_1 \|\nabla f(\hat{w}_t)\|^2 \mathbb{E}^{|\mathcal{F}_t}(\frac{1}{\sqrt{v_{t-1}}}-\frac{1}{\sqrt{v_t}}) 
\end{eqnarray}
where (a) holds because of $\langle x,y \rangle \leq \|x\|\|y\|$; (b) holds because $\|\nabla f(w_t,\xi_t)\| \leq \sqrt{v_t}$; (c) comes from Cauchy's Inequality; (d) comes from Lemma \ref{var}; (e) holds because 
\begin{equation}
    \frac{\|\nabla f(w_t, \xi_t)\|^2}{\sqrt{v_{t-1}}(\sqrt{v_{t-1}}+\sqrt{v_t})^2} \leq \frac{\|\nabla f(w_t, \xi_t)\|^2}{\sqrt{v_{t-1}} \sqrt{v_t}(\sqrt{v_{t-1}}+\sqrt{v_t})} = \frac{1}{\sqrt{v_{t-1}}}-\frac{1}{\sqrt{v_t}},
\end{equation}
Taking the expectation on the above inequality over $\mathcal{F}_t$ and summing up over $t \in \{1,2,...,T-1\}$ yields that
\begin{equation}\label{lemma2_5}
\sum_{t=1}^{T-1} \mathbb{E} \langle \nabla f(\hat{w}_t), \nabla f(w_t, \xi_t)(\frac{1}{\sqrt{v_{t-1}}}-\frac{1}{\sqrt{v_t}}) \rangle \stackrel{(f)}{\leq}  \frac{1}{2}\sum_{t=1}^{T-1} \mathbb{E}\frac{\|\nabla f(\hat{w}_t)\|^2}{\sqrt{v_{t-1}}} + \frac{D_0+4\rho^2 L^2}{\epsilon} + D_1 \sum_{t=1}^{T-1} \mathbb{E}\|\nabla f(\hat{w}_t)\|^2(\frac{1}{\sqrt{v_{t-1}}}-\frac{1}{\sqrt{v_t}})
\end{equation}
where (f) comes from Lemma \ref{ine}.
Then, for the term $\mathbb{E}^{|\mathcal{F}_t}\langle \nabla f(w_t), \rho(\frac{\nabla f(x_{t+1},\xi_{t+1})}{\sqrt{u_{t+1}}} - \frac{\nabla f(x_t,\xi_t)}{\sqrt{u_t}}) \rangle$, summing it up over $t \in \{1,2,...,T-1\}$ yields that
\begin{eqnarray}\label{lemma2_6}
&&\sum_{t=1}^{T-1} \mathbb{E}^{|\mathcal{F}_t}\langle \nabla f(w_t), \rho(\frac{\nabla f(x_{t+1}, \xi_{t+1})}{\sqrt{u_{t+1}}} - \frac{\nabla f(x_t, \xi_t)}{\sqrt{u_t}}) \rangle \nonumber \\
&=& \rho \sum_{t=1}^{T-1} \mathbb{E}^{|\mathcal{F}_t}\langle \nabla f(w_{t+1}), \frac{\nabla f(x_{t+1}, \xi_{t+1})}{\sqrt{u_{t+1}}} \rangle - \mathbb{E}^{|\mathcal{F}_t}\langle \nabla f(w_t), \frac{\nabla f(x_t, \xi_t)}{\sqrt{u_t}} \rangle + \mathbb{E}^{|\mathcal{F}_t}\langle \nabla f(w_t)- \nabla f(w_{t+1}), \frac{\nabla f(x_{t+1}, \xi_{t+1})}{\sqrt{u_{t+1}}} \rangle \nonumber \\
&=& \mathbb{E}^{|\mathcal{F}_T}\langle \nabla f(w_T), \rho \frac{\nabla f(x_T, \xi_T)}{\sqrt{u_T}} \rangle - \mathbb{E}^{|\mathcal{F}_1}\langle \nabla f(w_1), \rho \frac{\nabla f(x_1, \xi_1)}{\sqrt{u_1}} \rangle + \rho \sum_{t=1}^{T-1} \mathbb{E}^{|\mathcal{F}_t} \langle \nabla f(w_t)-\nabla f(w_{t+1}), \frac{\nabla f(x_{t+1}, \xi_{t+1})}{\sqrt{u_{t+1}}} \rangle \nonumber \\
\end{eqnarray}
For the first term on the RHS of (\ref{lemma2_6})
\begin{eqnarray}\label{lemma2_8}
&&\mathbb{E}^{|\mathcal{F}_T}\langle \nabla f(w_T), \rho \frac{\nabla f(x_T, \xi_T)}{\sqrt{u_T}} \rangle \nonumber \\
&=& \mathbb{E}^{|\mathcal{F}_T} \langle \nabla f(x_T + \rho \frac{\nabla f(x_T, \xi_T)}{\sqrt{u_T}}) - \nabla f(x_T), \rho \frac{\nabla f(x_T, \xi_T)}{\sqrt{u_T}} \rangle + \rho \mathbb{E}^{|\mathcal{F}_T} \langle \nabla f(x_T), \frac{\nabla f(x_T, \xi_T)}{\sqrt{u_T}} \rangle \nonumber \\
&\stackrel{(g)}{\leq}& \rho^2 L + \rho \mathbb{E}^{|\mathcal{F}_T} \langle \nabla f(x_T), \frac{\nabla f(x_T, \xi_T)}{\sqrt{u_{T-1}}} \rangle + \rho \mathbb{E}^{|\mathcal{F}_T} \langle \nabla f(x_T), \nabla f(x_T, \xi_T)(\frac{1}{\sqrt{u_T}} - \frac{1}{\sqrt{u_{T-1}}}) \rangle \nonumber \\
&\stackrel{(h)}{\leq}& \rho^2 L + \rho \frac{\|\nabla f(x_T)\|^2}{\sqrt{u_{T-1}}} + \rho \mathbb{E}^{|\mathcal{F}_T} \frac{\|\nabla f(x_T)\|\|\nabla f(x_T, \xi_T)\|^3}{\sqrt{u_{T-1}} \sqrt{u_T} (\sqrt{u_{T-1}} + \sqrt{u_T})} \nonumber \\
&\stackrel{(i)}{\leq}& \rho^2 L + \rho \frac{\|\nabla f(x_T)\|^2}{\sqrt{u_{T-1}}} + \rho \mathbb{E}^{|\mathcal{F}_T}\frac{\|\nabla f(x_T)\|\|\nabla f(x_T, \xi_T)\|^2}{\sqrt{u_{T-1}} (\sqrt{u_{T-1}} + \sqrt{u_T})}
\end{eqnarray}
where (g) holds because $\langle a,b \rangle \leq \|a\|\|b\|$ and Assumption 1; (h) and (i) hold in the same way as (\ref{lemma2_4}).
For the last term on the RHS of (\ref{lemma2_8})
\begin{eqnarray}\label{lemma2_9}
&&\mathbb{E}^{|\mathcal{F}_T} \frac{\|\nabla f(x_T)\|\|\nabla f(x_T, \xi_T)\|^2}{\sqrt{u_{T-1}} (\sqrt{u_{T-1}} + \sqrt{u_T})} \nonumber \\
&\leq& \frac{\sqrt{D_1}}{2}\frac{\|\nabla f(x_T)\|^2}{\sqrt{u_{T-1}}} + \frac{1}{2\sqrt{D_1} \sqrt{u_{T-1}}} (\mathbb{E}^{|\mathcal{F}_T}\frac{\|\nabla f(x_T,\xi_T)\|^2}{\sqrt{u_{T-1}}+\sqrt{u_T}})^2 \nonumber \\
&\leq& \frac{\sqrt{D_1}}{2}\frac{\|\nabla f(x_T)\|^2}{\sqrt{u_{T-1}}} + \frac{1}{2\sqrt{D_1}\sqrt{u_{T-1}}} (\mathbb{E}^{|\mathcal{F}_T} \|\nabla f(x_T,\xi_T)\|^2) (\mathbb{E}^{|\mathcal{F}_T} \frac{\|\nabla f(x_T,\xi_T)\|^2}{(\sqrt{u_{T-1}}+\sqrt{u_T})^2}) \nonumber \\
&\leq& \frac{\sqrt{D_1}}{2}\frac{\|\nabla f(x_T)\|^2}{\sqrt{u_{T-1}}} + \frac{1}{2\sqrt{D_1}\sqrt{u_{T-1}}} (D_0 + D_1 \|\nabla f(x_T)\|^2) (\mathbb{E}^{|\mathcal{F}_T} \frac{\|\nabla f(x_T,\xi_T)\|^2}{(\sqrt{u_{T-1}}+\sqrt{u_T})^2}) \nonumber \\
&\stackrel{(j)}{\leq}& \sqrt{D_1}\frac{\|\nabla f(x_T)\|^2}{\sqrt{u_{T-1}}} + \frac{D_0}{2 \epsilon \sqrt{D_1}} 
\end{eqnarray}
where (j) holds because $\frac{\|\nabla f(x_T,\xi_T)\|^2}{(\sqrt{u_{T-1}}+\sqrt{u_T})^2} \leq 1$ and $\sqrt{u_{T-1}} \geq \epsilon$.
Combining (\ref{lemma2_8}) and (\ref{lemma2_9}) yields
\begin{equation}\label{lemma2_10}
\mathbb{E}^{|\mathcal{F}_T}\langle \nabla f(w_T), \rho \frac{\nabla f(x_T, \xi_T)}{\sqrt{u_T}} \rangle \leq \rho^2 L + \frac{\rho D_0}{2\epsilon \sqrt{D_1}} + (1+\sqrt{D_1})\rho \frac{\|\nabla f(x_T)\|^2}{\sqrt{u_{T-1}}}
\end{equation}
For the second term on the RHS of (\ref{lemma2_6})
\begin{eqnarray}\label{lemma2_11}
&&-\mathbb{E}^{|\mathcal{F}_1} \langle \nabla f(w_1), \rho \frac{\nabla f(x_1, \xi_1)}{\sqrt{u_1}} \rangle \nonumber \\
&=& -\mathbb{E}^{|\mathcal{F}_1} \langle \nabla f(x_1 + \rho \frac{\nabla f(x_1, \xi_1)}{\sqrt{u_1}}) - f(x_1), \rho \frac{\nabla f(x_1, \xi_1)}{\sqrt{u_1}} \rangle - \mathbb{E}^{|\mathcal{F}_1} \langle \nabla f(x_1), \rho \frac{\nabla f(x_1, \xi_1)}{\sqrt{u_1}} \rangle \nonumber \\
&\leq& \rho^2 L + \mathbb{E}^{|\mathcal{F}_1} \|\nabla f(x_1)\|\|\rho \frac{\nabla f(x_1,\xi_1)}{\sqrt{u_1}}\| \leq \rho^2 L + \rho \|\nabla f(x_1)\|
\end{eqnarray}
For the last term on the RHS of (\ref{lemma2_6})
\begin{eqnarray}\label{lemma2_12}
&&\sum_{t=1}^{T-1} \mathbb{E}^{|\mathcal{F}_t} \langle \nabla f(w_t)-f(w_{t+1}), \frac{\nabla f(x_{t+1}, \xi_{t+1})}{\sqrt{u_{t+1}}} \rangle \nonumber \\
&\stackrel{(k)}{\leq}& \frac{L^2}{2}\sum_{t=1}^{T-1}\mathbb{E}^{|\mathcal{F}_t}\|w_{t+1}-w_t\|^2 + \frac{1}{2}\sum_{t=1}^{T-1}\mathbb{E}^{|\mathcal{F}_t}\frac{\|\nabla f(x_{t+1},\xi_{t+1})\|^2}{u_{t+1}} \nonumber \\
&\stackrel{(l)}{\leq}& \eta^2 L^2 (\mathbb{E}^{|\mathcal{F}_{T-1}} \ln v_{T-1} - \ln v_0) + 4\rho^2 L^2 (\mathbb{E}^{|\mathcal{F}_T} \ln u_T - \ln u_0) + \frac{1}{2}(\mathbb{E}^{|\mathcal{F}_T} \ln u_T - \ln u_0) \nonumber \\
\end{eqnarray}
where (k) comes from Assumption 1; the (l) comes from Lemma \ref{ine}. 
Substituting (\ref{lemma2_10}), (\ref{lemma2_11}) and (\ref{lemma2_12}) into (\ref{lemma2_6}) and taking the expectation over $\mathcal{F}_t$ yield that \begin{eqnarray}\label{lemma2_13}
&&\sum_{t=1}^{T-1} \mathbb{E}\langle \nabla f(w_t), \rho(\frac{\nabla f(x_{t+1}, \xi_{t+1})}{\sqrt{u_{t+1}}} - \frac{\nabla f(x_t, \xi_t)}{\sqrt{u_t}}) \rangle \nonumber \\
&\leq& 2\rho^2 L + \frac{\rho D_0}{2\epsilon \sqrt{D_1}} + \rho \|\nabla f(x_1)\| - (\rho \eta^2 L^2 + 4 \rho^3 L^2 + \frac{\rho}{2}) \ln u_0 \nonumber \\
&&+ (1+\sqrt{D_1})\rho \mathbb{E} \frac{\|\nabla f(x_T)\|^2}{\sqrt{u_{T-1}}} + \rho \eta^2 L^2 \mathbb{E} \ln v_T + (4\rho^3 L^2 + \frac{\rho}{2}) \mathbb{E} \ln u_T.
\end{eqnarray}
Finally, we have
\begin{eqnarray}\label{lemma2_14}
\sum_{t=1}^{T-1}\mathbb{E}\|w_{t+1}-w_t\|^2 \leq 2\eta^2(\mathbb{E} \ln v_T - \ln v_0) + 8\rho^2 (\mathbb{E} \ln u_T - \ln u_0). 
\end{eqnarray}
Summing up (\ref{lemma2_3}) over $t \in \{1,2,...,T-1\}$, substituting (\ref{lemma2_5}), (\ref{lemma2_13}) and (\ref{lemma2_14}) into it yields that
\begin{eqnarray}
&&\frac{3\eta}{4} \sum_{t=1}^{T-1} \mathbb{E} \frac{\|\nabla f(\hat{w}_t)\|^2}{\sqrt{v_{t-1}}} \leq f(w_1)+2\rho^2L + \frac{\rho D_0}{2\epsilon \sqrt{D_1}} + \rho \|\nabla f(x_1)\| +\frac{(D_0+4\rho^2 L^2)\eta}{\epsilon} + \frac{\eta}{2}\sum_{t=1}^{T-1}\mathbb{E} \frac{\|\nabla f(\hat{w}_t)\|^2}{\sqrt{v_{t-1}}} \nonumber \\
&&+ (1+\sqrt{D_1})\rho \mathbb{E} \frac{\|\nabla f(x_T)\|^2}{\sqrt{u_{T-1}}}+ D_1 \eta\sum_{t=1}^{T-1} \mathbb{E}\|\nabla f(\hat{w}_t)\|^2 (\frac{1}{\sqrt{v_{t-1}}}-\frac{1}{\sqrt{v_t}}) +(\frac{\eta}{2}+\eta^2 L +\rho \eta^2 L^2) \mathbb{E} \ln v_T \nonumber \\
&&+\frac{3\rho^2 \eta L^2}{\epsilon} \mathbb{E} \ln \hat{u}_T+ (\frac{3\rho^2 \eta L^2}{\epsilon}+4\rho^2 L+4\rho^3 L^2 + \frac{\rho}{2}) \mathbb{E} \ln u_T - (\frac{12\rho^2 \eta L^2}{\epsilon}+\eta + \rho + (1+\rho L)(2\eta^2 L+8\rho^2 L)) \ln \epsilon. \nonumber 
\end{eqnarray}
Rearranging the above inequality yields the result. Here we simplify the formula by adopting the assumption that $\epsilon$ is a very small value to avoid the denominator to be zero.
\end{IEEEproof}

\begin{lemma}\label{relemma3}
(Restatement of Lemma 3) If $f(x)$ in Algorithm 1 satisfies Assumptions 1, we have that
\begin{equation}
    \|\nabla f(w_t,\xi_t)\|^2 \leq (\frac{\rho L}{\epsilon} + 1)\|\nabla f(x_t,\xi_t)\|^2, \quad v_t \leq (\frac{\rho L}{\epsilon}+1) u_t \nonumber
\end{equation}
\end{lemma}
\begin{IEEEproof}
\begin{eqnarray}
\|\nabla f(w_t,\xi_t)\|^2 
&=&\|\nabla f(w_t,\xi_t) - \nabla f(x_t,\xi_t)\|^2 + 2\langle \nabla f(w_t,\xi_t) - \nabla f(x_t,\xi_t), \nabla f(x_t,\xi_t) \rangle + \|\nabla f(x_t,\xi_t)\|^2 \nonumber \\
&\leq& L^2 \|w_t - x_t\|^2 + 2L \|w_t - x_t\|\|\nabla f(x_t, \xi_t)\|+\|\nabla f(x_t,\xi_t)\|^2 \nonumber \\
&=& \rho^2 L^2 \frac{\|\nabla f(x_t,\xi_t)\|^2}{u_t} + 2\rho L \frac{\|\nabla f(x_t,\xi_t)\|}{\sqrt{u_t}}\|\nabla f(x_t, \xi_t)\|+\|\nabla f(x_t,\xi_t)\|^2 \nonumber \\
&=& (\frac{\rho L}{\sqrt{u_t}}+1)^2 \|\nabla f(x_t,\xi_t)\|^2 \leq (\frac{\rho L}{\epsilon} + 1)^2 \|\nabla f(x_t,\xi_t)\|^2 \nonumber
\end{eqnarray}
where the last inequality holds because $u_t \geq u_0 = \epsilon^2$. Further, we can obtain $v_t \leq (\frac{\rho L}{\epsilon}+1) u_t$.
\end{IEEEproof}

\begin{theorem}
(Restatement of Theorem 1) If $f(x)$ in Algorithm 1 satisfies Assumptions 1 and 2, for any perturbation radius $\rho$ and learning rate $eta > 0$, we have that 
\begin{equation}
    \sum_{t=1}^T \mathbb{E} \|\nabla f(x_t)\| \leq  \sqrt{T\bigg(2A_6(A_3+2A_5\ln A_6) + 2A_7\ln(A_7+e) + 4096D_1^2 A_4^2 (2A_4+\frac{16D_1 A_4 A_5}{A_6})^2+1\bigg)} \nonumber
\end{equation}
where
\begin{eqnarray}
\hspace{-0.5cm}&&A_3 = \sqrt{\frac{\rho L}{\epsilon}+1}[\frac{4f(x_1)}{\eta}\!+\!\frac{8(D_0\!+\!4\rho^2 L^2)}{\epsilon}\!+\!16 D_1 A_1\!+\!\frac{8A_2}{\eta} - (\frac{80 \rho^2 L^2}{\epsilon} + 4\eta L) \ln \epsilon\!+\!(4\eta L (3\!+\!\rho L)\!+\!8) \ln (1+\frac{\rho L}{\epsilon})], \nonumber \\
\hspace{-0.5cm}&&A_4 = \sqrt{\frac{\rho L}{\epsilon}+1}\frac{16(8(1\!+\!2D_2)D_1\!+\!3)\rho^2 L^2}{\epsilon},\ A_5 = \sqrt{\frac{\rho L}{\epsilon}+1}[\frac{40\rho^2 L^2}{\epsilon}+\frac{4\rho}{\eta}(1\!+\!8\rho L(1\!+\!\rho L))+4\eta L (3+2\rho L)+8], \nonumber \\
\hspace{-0.5cm}&&A_6 = 2\sqrt{2D_0 T + \epsilon^2} + 4D_1 A_3 + 8D_1 A_5 \ln (4D_1 A_5+e), \quad A_7 = 2A_4 A_6+16D_1A_4A_5+8D_1A_4(A_3+2A_5\ln A_6) \nonumber
\end{eqnarray}
\end{theorem}
\begin{IEEEproof}
According to the $L$-smoothness of $f(x)$, we have
\begin{eqnarray}\label{theo_1}
\mathbb{E}^{\mathcal{F}_t}[f(x_{t+1})] &\leq& f(x_t) + \mathbb{E}^{\mathcal{F}_t}\langle \nabla f(x_t), x_{t+1}-x_t \rangle +\frac{L}{2}\mathbb{E}^{\mathcal{F}_t}\|x_{t+1}-x_t\|^2 \nonumber \\
&=& f(x_t) - \eta \mathbb{E}^{\mathcal{F}_t}\langle \nabla f(x_t), \frac{g_t}{\sqrt{v_t}} \rangle + \frac{\eta^2 L}{2}\mathbb{E}^{\mathcal{F}_t}\|\frac{g_t}{\sqrt{v_t}}\|^2 \nonumber \\
&=& f(x_t) + \underbrace{\eta \mathbb{E}^{\mathcal{F}_t} \langle \nabla f(x_t), \frac{-g_t}{\sqrt{v_{t-1}}} \rangle}_{T_1} + \underbrace{\eta \mathbb{E}^{\mathcal{F}_t}\langle \nabla f(x_t), g_t (\frac{1}{\sqrt{v_{t-1}}}-\frac{1}{\sqrt{v_t}}) \rangle}_{T_2} + \underbrace{\frac{\eta^2 L}{2} \mathbb{E}^{\mathcal{F}_t}\|\frac{g_t}{\sqrt{v_t}}\|^2}_{T_3} 
\end{eqnarray}
For $T_1$,
\begin{eqnarray}
T_1 &=& \eta \mathbb{E}^{\mathcal{F}_t} \langle \nabla f(x_t), \frac{-\nabla f(\hat{w}_t,\xi_t)}{\sqrt{v_{t-1}}} \rangle + \eta \mathbb{E}^{\mathcal{F}_t} \langle \nabla f(x_t), \frac{\nabla f(\hat{w}_t,\xi_t) - \nabla f(w_t,\xi_t)}{\sqrt{v_{t-1}}} \rangle  \nonumber \\
&\leq& \eta \mathbb{E}^{\mathcal{F}_t} \langle \nabla f(x_t), \frac{-\nabla f(\hat{w}_t)}{\sqrt{v_{t-1}}} \rangle + \eta \mathbb{E}^{\mathcal{F}_t} (\frac{\|\nabla f(x_t)\|^2}{8\sqrt{v_{t-1}}} + \frac{2L^2 \|\hat{w}_t-w_t\|^2}{\sqrt{v_{t-1}}}) \nonumber \\
&=&  \eta \mathbb{E}^{\mathcal{F}_t} \frac{1}{\sqrt{v_{t-1}}}\bigg(\langle \nabla f(x_t), \nabla f(x_t)-\nabla f(x_t + \rho \frac{\nabla f(x_t)}{\sqrt{\hat{u}_t}}) \rangle- \langle \nabla f(x_t), \nabla f(x_t)\rangle \bigg) \nonumber \\
&&+ \eta \mathbb{E}^{\mathcal{F}_t} (\frac{\|\nabla f(x_t)\|^2}{8\sqrt{v_{t-1}}} + \frac{2L^2 \|\hat{w}_t-w_t\|^2}{\sqrt{v_{t-1}}}) \nonumber \\
&\leq& \frac{\eta}{8} \mathbb{E}^{\mathcal{F}_t} \frac{\|\nabla f(x_t)\|^2}{\sqrt{v_{t-1}}} + \frac{2\eta}{\sqrt{v_{t-1}}}\mathbb{E}^{\mathcal{F}_t}\|\nabla f(x_t)-\nabla f(x_t + \rho \frac{\nabla f(x_t)}{\sqrt{\hat{u}_t}})\|^2 - \eta \mathbb{E}^{\mathcal{F}_t}\frac{\|\nabla f(x_t)\|^2}{\sqrt{v_{t-1}}} \nonumber \\
&&+ \eta \mathbb{E}^{\mathcal{F}_t} \bigg(\frac{\|\nabla f(x_t)\|^2}{8\sqrt{v_{t-1}}} + \frac{4\rho^2 L^2 (\frac{\|\nabla f(x_t,\xi_t)\|^2}{u_t}+\frac{\|\nabla f(x_t)\|^2}{\hat{u}_t})}{\sqrt{v_{t-1}}}\bigg) \nonumber \\
&\stackrel{(a)}{\leq}& -\frac{3 \eta }{4} \mathbb{E}^{\mathcal{F}_t} \frac{\|\nabla f(x_t)\|^2}{\sqrt{v_{t-1}}} + \frac{\eta}{\sqrt{v_{t-1}}} \mathbb{E}^{\mathcal{F}_t}(\frac{4\rho^2 L^2 \|\nabla f(x_t, \xi_t)\|^2}{u_{t}}+\frac{6\rho^2 L^2 \|\nabla f(x_t)\|^2}{\hat{u}_{t}}) \nonumber \\
&\leq& -\frac{3 \eta }{4} \mathbb{E}^{\mathcal{F}_t} \frac{\|\nabla f(x_t)\|^2}{\sqrt{v_{t-1}}} + \frac{\rho^2 \eta L^2}{\epsilon} \mathbb{E}^{\mathcal{F}_t}(4\frac{\|\nabla f(x_t, \xi_t)\|^2}{u_{t}}+6\frac{\|\nabla f(x_t)\|^2}{\hat{u}_{t}}) \nonumber
\end{eqnarray}
where (a) comes from Assumption 1. Taking the expectation on the above inequality over $\mathcal{F}_t$ and summing up over $t \in \{1,2,...,T\}$, then combining the result with Lemma \ref{ine} yields that
\begin{equation}\label{theo_2}
\sum_{t=1}^T \eta \mathbb{E}\langle \nabla f(x_t), \frac{-g_t}{\sqrt{v_{t-1}}} \rangle
\leq -\frac{3 \eta }{4} \sum_{t=1}^T \mathbb{E} \frac{\|\nabla f(x_t)\|^2}{\sqrt{v_{t-1}}} + \frac{\rho^2 \eta L^2}{\epsilon}(4\mathbb{E} \ln u_T + 6\mathbb{E} \ln \hat{u}_T- 20 \ln \epsilon) 
\end{equation}
For $T_2$,
\begin{eqnarray}
T_2 &=& \eta \mathbb{E}^{\mathcal{F}_t} \langle \nabla f(x_t), \frac{\nabla f(w_t, \xi_t) \|\nabla f(w_t, \xi_t)\|^2}{\sqrt{v_{t-1}}\sqrt{v_t}(\sqrt{v_{t-1}}+\sqrt{v_t})} \rangle \leq \eta \mathbb{E}^{\mathcal{F}_t} \frac{\|\nabla f(x_t)\| \|\nabla f(w_t, \xi_t)\|^3}{\sqrt{v_{t-1}}\sqrt{v_t}(\sqrt{v_{t-1}}+\sqrt{v_t})} \nonumber \\
&\leq& \eta \frac{\|\nabla f(x_t)\|}{v_{t-1}^{1/4}}\mathbb{E}^{\mathcal{F}_t} \frac{ \|\nabla f(w_t, \xi_t)\|^2}{v_{t-1}^{1/4}(\sqrt{v_{t-1}}+\sqrt{v_t})} \leq \frac{\eta}{4} \frac{\|\nabla f(x_t)\|^2}{\sqrt{v_{t-1}}} +  \eta\bigg(\mathbb{E}^{\mathcal{F}_t} \frac{\|\nabla f(w_t, \xi_t)\|^2}{v_{t-1}^{1/4}(\sqrt{v_{t-1}}+\sqrt{v_t})}\bigg)^2 \nonumber \\
&\leq& \frac{\eta}{4}  \mathbb{E}^{\mathcal{F}_t}\frac{\|\nabla f(x_t)\|^2}{\sqrt{v_{t-1}}} + \eta  (\mathbb{E}^{\mathcal{F}_t}\|\nabla f(w_t, \xi_t)\|^2)\bigg(\mathbb{E}^{\mathcal{F}_t} \frac{\|\nabla f(w_t, \xi_t)\|^2}{\sqrt{v_{t-1}}(\sqrt{v_{t-1}}+\sqrt{v_t})^2}\bigg) \nonumber \\
&\leq& \frac{\eta}{4} \mathbb{E}^{\mathcal{F}_t}\frac{\|\nabla f(x_t)\|^2}{\sqrt{v_{t-1}}} + (2D_0+8\rho^2 L^2)\eta \mathbb{E}^{\mathcal{F}_t} \frac{\|\nabla f(w_t, \xi_t)\|^2}{\sqrt{v_{t-1}}(\sqrt{v_{t-1}}+\sqrt{v_t})^2} + 2D_1 \eta \|\nabla f(\hat{w}_t)\|^2 \mathbb{E}^{\mathcal{F}_t}(\frac{1}{\sqrt{v_{t-1}}}-\frac{1}{\sqrt{v_t}}) \nonumber
\end{eqnarray}
The derivation for $T_2$ follows the same way as (\ref{lemma2_5}). Taking the expectation on the above inequality over $\mathcal{F}_t$ and summing up over $t \in \{1,2,...,T\}$ yield that
\begin{eqnarray}\label{theo_3}
&&\sum_{t=1}^T \eta \mathbb{E}\langle \nabla f(x_t), g_t (\frac{1}{\sqrt{v_{t-1}}}-\frac{1}{\sqrt{v_t}}) \rangle \nonumber \\
&\leq& \frac{\eta}{4} \sum_{t=1}^T \mathbb{E} \frac{\|\nabla f(x_t)\|^2}{\sqrt{v_{t-1}}}\!+\!\frac{(2D_0\!+\!8\rho^2 L^2) \eta}{\epsilon} + 2D_1 \eta \sum_{t=1}^T\mathbb{E} \|\nabla f(\hat{w}_t)\|^2 (\frac{1}{\sqrt{v_{t-1}}}-\frac{1}{\sqrt{v_t}})   
\end{eqnarray}
From Lemma \ref{relemma1}, we can obtain that
\begin{eqnarray}\label{theo_4}
D_1 \eta \sum_{t=1}^T \|\nabla f(\hat{w}_t)\|^2 \mathbb{E}(\frac{1}{\sqrt{v_{t-1}}}-\frac{1}{\sqrt{v_t}}) 
&\leq& D_1 \eta A_1 + \frac{D_1}{2 D_2} \eta \sum_{t=1}^T \mathbb{E} \frac{\|\nabla f(\hat{w}_t)\|^2}{\sqrt{v_{t-1}}} + \frac{8(1+2 D_2) D_1 \rho^2 \eta L^2}{\epsilon} \mathbb{E} \ln \hat{u}_T \nonumber \\
&&- D_1 \eta \mathbb{E}\frac{\|\nabla f(\hat{w}_T)\|^2}{\sqrt{v_T}}
\end{eqnarray}
By Lemma \ref{relemma2}, we can further obtain that
\begin{eqnarray}\label{theo_5}
&&- D_1 \eta \mathbb{E}\frac{\|\nabla f(\hat{w}_T)\|^2}{\sqrt{v_T}} + \frac{D_1}{2 D_2} \eta \sum_{t=1}^T \mathbb{E} \frac{\|\nabla f(\hat{w}_t)\|^2}{\sqrt{v_{t-1}}} \nonumber \\
&\leq& - \frac{D_1}{2 D_2} \eta \mathbb{E}\frac{\|\nabla f(\hat{w}_T)\|^2}{\sqrt{v_T}} + \frac{D_1}{2 D_2} \eta (\sum_{t=1}^{T-1} \mathbb{E} \frac{\|\nabla f(\hat{w}_t)\|^2}{\sqrt{v_{t-1}}} + \frac{\|\nabla f(\hat{w}_T)\|^2}{\sqrt{v_{T-1}}}) \nonumber \\
&\leq& \frac{2 D_1^2}{D_2} \eta \sum_{t=1}^{T-1} \mathbb{E} \|\nabla f(\hat{w}_t)\|^2(\frac{1}{\sqrt{v_{t-1}}}-\frac{1}{\sqrt{v_t}}) + \frac{D_1}{2 D_2} \eta \mathbb{E} \|\nabla f(\hat{w}_T)\|^2(\frac{1}{\sqrt{v_{T-1}}}-\frac{1}{\sqrt{v_T}}) + \frac{2(1\!+\!\sqrt{D_1})D_1\rho}{D_2} \mathbb{E} \frac{\|\nabla f(x_T)\|^2}{\sqrt{u_{T-1}}}\nonumber \\
&&+ \frac{A_2}{2} + \frac{\eta^2 L (1+\rho L)+\eta}{2}\mathbb{E} \ln v_T + (\frac{3\rho^2 \eta L^2}{2\epsilon}+2\rho^2 L+2\rho^3 L^2 + \frac{\rho}{4})\mathbb{E} \ln u_T +\frac{3\rho^2 \eta L^2}{2\epsilon}\mathbb{E} \ln \hat{u}_T\nonumber \\
&\leq& \frac{D_1}{2} \eta \sum_{t=1}^{T} \mathbb{E} \|\nabla f(w_t)\|^2(\frac{1}{\sqrt{v_{t-1}}}-\frac{1}{\sqrt{v_t}}) + \frac{\eta}{16}\sum_{t=1}^T \mathbb{E} \frac{\|\nabla f(x_t)\|^2}{\sqrt{v_{t-1}}} + \frac{A_2}{2} + \frac{\eta^2 L (1+\rho L)+\eta}{2}\mathbb{E} \ln v_T \nonumber \\
&&+ (\frac{3\rho^2 \eta L^2}{2\epsilon}+2\rho^2 L+2\rho^3 L^2 + \frac{\rho}{4})\mathbb{E} \ln u_T +\frac{3\rho^2 \eta L^2}{2\epsilon}\mathbb{E} \ln \hat{u}_T.
\end{eqnarray}
Note that in the second inequality, the term $-\mathbb{E}\frac{\|\nabla f(\hat{w}_T)\|^2}{\sqrt{v_T}}$ contribute to form the term $\mathbb{E}\|\nabla f(\hat{w}_T)\|^2(\frac{1}{\sqrt{v_{T-1}}}-\frac{1}{\sqrt{v_T}})$, then plusing the term $\sum_{t=1}^{T-1} \mathbb{E} \|\nabla f(\hat{w}_t)\|^2(\frac{1}{\sqrt{v_{t-1}}}-\frac{1}{\sqrt{v_t}})$ in Lemma \ref{relemma2} to constitute the complete $\sum_{t=1}^T \mathbb{E} \|\nabla f(\hat{w}_t)\|^2(\frac{1}{\sqrt{v_{t-1}}}-\frac{1}{\sqrt{v_t}})$. \\[6pt]
The above inequalities utilize the definition of $D_2$ and Lemma \ref{relemma3}. Substituting (\ref{theo_5}) into (\ref{theo_4}) yields that
\begin{eqnarray}\label{theo_6}
&&D_1 \eta \sum_{t=1}^T \|\nabla f(w_t)\|^2 \mathbb{E}(\frac{1}{\sqrt{v_{t-1}}}-\frac{1}{\sqrt{v_t}}) 
\leq (\eta^2 L (1+\rho L)+\eta) \mathbb{E}\ln v_T + (\frac{3\rho^2 \eta L^2}{\epsilon}+4\rho^2 L+4\rho^3 L^2 + \frac{\rho}{2})\mathbb{E} \ln u_T \nonumber \\ &&+ \frac{(16(1+2D_2)D_1+3)\rho^2 \eta L^2}{\epsilon}\mathbb{E} \ln \hat{u}_T+ 2 D_1 \eta A_1 + A_2+\frac{\eta}{8}\sum_{t=1}^T \mathbb{E} \frac{\|\nabla f(x_t)\|^2}{\sqrt{v_{t-1}}}.  
\end{eqnarray}
Substituting (\ref{theo_6}) into (\ref{theo_3}) yields that 
\begin{eqnarray}\label{theo_7}
&&\sum_{t=1}^T \eta \mathbb{E}\langle \nabla f(x_t), g_t (\frac{1}{\sqrt{v_{t-1}}}-\frac{1}{\sqrt{v_t}}) \rangle \nonumber \\
&\leq& \frac{\eta}{2} \sum_{t=1}^T \mathbb{E} \frac{\|\nabla f(x_t)\|^2}{\sqrt{v_{t-1}}} + \frac{(2D_0+8\rho^2 L^2) \eta}{\epsilon} +4 D_1 \eta A_1 + 2A_2 + 2(\eta^2 L (1+\rho L)+\eta)\mathbb{E}\ln v_T \nonumber \\
&&+ (\frac{6\rho^2 \eta L^2}{\epsilon}+8\rho^2 L+8\rho^3 L^2 + \rho)\mathbb{E} \ln u_T + \frac{2(16(1+2D_2)D_1+3)\rho^2 \eta L^2}{\epsilon}\mathbb{E} \ln \hat{u}_T
\end{eqnarray}
For $T_3$
\begin{equation}\label{theo_8}
    \frac{\eta^2 L}{2} \sum_{t=1}^{T} \mathbb{E}\|\frac{g_t}{\sqrt{v_t}}\|^2 \leq \frac{\eta^2 L}{2} (\mathbb{E} \ln  v_T - \ln v_0) = \frac{\eta^2 L}{2} (\mathbb{E} \ln  v_T - 2\ln \epsilon)
\end{equation}
Combining (\ref{theo_1}), (\ref{theo_2}), (\ref{theo_7}) and (\ref{theo_8}) yields that
\begin{eqnarray}
&&\frac{\eta}{4} \sum_{t=1}^T \mathbb{E} \frac{\|\nabla f(x_t)\|^2}{\sqrt{v_{t-1}}} \nonumber \\
&\leq& f(x_1)+\frac{(2D_0+8\rho^2 L^2) \eta}{\epsilon} +4 D_1 \eta A_1 + 2A_2 - (\frac{20 \rho^2 \eta L^2}{\epsilon} + \eta^2 L) \ln \epsilon + (\eta^2 L (3+2\rho L)+2\eta) \mathbb{E} \ln v_T \nonumber \\
&&+ (\frac{10\rho^2 \eta L^2}{\epsilon}+8\rho^2 L+8\rho^3 L^2 + \rho)\mathbb{E} \ln u_T + \frac{4(8(1+2D_2)D_1+3)\rho^2 \eta L^2}{\epsilon}\mathbb{E} \ln \hat{u}_T\nonumber \\
&\leq& f(x_1)+\frac{(2D_0+8\rho^2 L^2) \eta}{\epsilon} +4 D_1 \eta A_1 + 2A_2 - (\frac{20 \rho^2 \eta L^2}{\epsilon} + \eta^2 L) \ln \epsilon + (\eta^2 L (3+2\rho L)+2\eta) \ln (1+\frac{\rho L}{\epsilon}) \nonumber \\
&&+ (\frac{10\rho^2 \eta L^2}{\epsilon}+8\rho^2 L+8\rho^3 L^2 + \rho+\eta^2 L (3+2\rho L)+2\eta)\mathbb{E} \ln u_T + \frac{4(8(1+2D_2)D_1+3)\rho^2 \eta L^2}{\epsilon}\mathbb{E} \ln \hat{u}_T\nonumber
\end{eqnarray}
where the last inequality comes from Lemma \ref{relemma3}. Rearranging the result and considering that $\frac{\|\nabla f(x_t)\|^2}{\sqrt{v_{t-1}}} \geq \sqrt{\frac{\epsilon}{\rho L + \epsilon}} \frac{\|\nabla f(x_t)\|^2}{\sqrt{u_{t-1}}}$ (which comes from Lemma \ref{relemma3}) yields that
\begin{equation}
\sum_{t=1}^T \mathbb{E} \frac{\|\nabla f(x_t)\|^2}{\sqrt{u_{t-1}}} \leq A_3 + A_4 \mathbb{E} \ln \hat{u}_T + A_5 \mathbb{E} \ln u_T \leq A_3 + 2A_4 \ln \mathbb{E} \sqrt{\hat{u}_T} + 2A_5 \ln \mathbb{E} \sqrt{u_T}.\nonumber
\end{equation}
Then, we adopt the same derivation as "Stage II" in \cite{wang2023convergence} to obtain that
\begin{equation}
    \mathbb{E} \sqrt{u_T} \leq \sqrt{2D_0 T + \epsilon^2} + 2D_1 A_3 + 4D_1 A_4 \ln \mathbb{E} \sqrt{\hat{u}_T} + 4D_1 A_5 \ln \mathbb{E} \sqrt{u_T}. \nonumber
\end{equation}
From Lemma \ref{bound}, we obtain that
\begin{equation}
    \mathbb{E} \sqrt{u_T} \leq 2\sqrt{2D_0 T + \epsilon^2} + 4D_1 A_3 + 8D_1 A_4 \ln \mathbb{E} \sqrt{\hat{u}_T} + 8D_1 A_5 \ln (4D_1 A_5+e). \nonumber
\end{equation}
Since
\begin{eqnarray}
    A_3 + 2A_4 \ln \mathbb{E} \sqrt{\hat{u}_T} + 2A_5 \ln \mathbb{E}\sqrt{u_T} \geq  \sum_{t=1}^T \mathbb{E} \frac{\|\nabla f(x_t)\|^2}{\sqrt{u_{t-1}}} \geq \mathbb{E} \frac{\sum_{t=1}^T \|\nabla f(x_t)\|^2}{\sqrt{u_T}} \geq \frac{\bigg(\mathbb{E}\sqrt{\sum_{t=1}^T \|\nabla f(x_t)\|^2}\bigg)^2}{\mathbb{E} \sqrt{u_T}}. \nonumber 
\end{eqnarray}
Considering that $\hat{u}_T = \sum_{t=1}^T \|\nabla f(x_t)\|^2$, we obtain the inequality
\begin{eqnarray}
    (\mathbb{E}\sqrt{\hat{u}_T})^2 &\leq& (A_6+8D_1 A_4 \ln \mathbb{E}\sqrt{\hat{u}_T})(A_3+2A_4\ln \mathbb{E}\sqrt{\hat{u}_T}+2A_5\ln(A_6+8D_1 A_4 \ln \mathbb{E}\sqrt{\hat{u}_T})) \nonumber \\
    &\leq& (A_6+8D_1 A_4 \ln \mathbb{E}\sqrt{\hat{u}_T})(A_3+2A_4\ln \mathbb{E}\sqrt{\hat{u}_T}+2A_5\ln A_6+ \frac{16D_1 A_4 A_5}{A_6} \ln \mathbb{E}\sqrt{\hat{u}_T}) \nonumber \\
    &=& A_6(A_3+2A_5\ln A_6) + (2A_4 A_6+16D_1A_4A_5+8D_1A_4(A_3+2A_5\ln A_6)) \ln \mathbb{E}  \sqrt{\hat{u}_T} \nonumber \\
    &&+ 8D_1 A_4 (2A_4+\frac{16D_1 A_4 A_5}{A_6})(\ln \mathbb{E}\sqrt{\hat{u}_T})^2. \nonumber
\end{eqnarray}
Finally, solving the above inequality by Lemma \ref{bound2}, we obtain that
\begin{eqnarray}
    &&\sum_{t=1}^T \mathbb{E} \|\nabla f(x_t)\| \leq \sqrt{T} \mathbb{E} \sqrt{\sum_{t=1}^T \|\nabla f(x_t)\|^2}  \nonumber \\
    &\leq& \sqrt{T\bigg(2A_6(A_3+2A_5\ln A_6) + 2A_7\ln(A_7+e) + 4096D_1^2 A_4^2 (2A_4+\frac{16D_1 A_4 A_5}{A_6})^2+1\bigg)}. \nonumber
\end{eqnarray}

\end{IEEEproof}

The proof of Theorem 2 is almost the same as the above proof. The difference is the scalars are replaced with vectors, as a result, for vectors $a$ and $b$, we turn to deal with $\|a \odot b\|^2 = \sum_{l=1}^d a_l^2 b_l^2$ and $\|\frac{1}{b} \odot a\|^2 = \sum_{l=1}^d \frac{a_l^2}{b_l^2}$. We do not repeat the proof process here.

\newpage

\section{Proof of Theorem 3}
Before the proof, we define 
\begin{equation}\check{r}_t = \beta_1 \check{r}_{t-1} + (1-\beta_1)\nabla f(x_t), \quad \check{u}_t = \beta_2 \check{u}_{t-1} + (1-\beta_2)\nabla f(x_t) \odot \nabla f(x_t), \quad \check{w}_t = x_t + \rho \frac{1}{\sqrt{\check{u}_t+\mathbf{\epsilon^2}}} \odot \check{r}_t.\nonumber
\end{equation}
\begin{equation}p_t = \frac{w_t-\frac{\beta_1}{\sqrt{\beta_2}}w_{t-1}}{1-\frac{\beta_1}{\sqrt{\beta_2}}}, \quad \check{p}_t = \frac{\check{w}_t-\frac{\beta_1}{\sqrt{\beta_2}}\check{w}_{t-1}}{1-\frac{\beta_1}{\sqrt{\beta_2}}}, \quad q_t = \frac{x_t-\frac{\beta_1}{\sqrt{\beta_2}}x_{t-1}}{1-\frac{\beta_1}{\sqrt{\beta_2}}},\nonumber
\end{equation}
\begin{equation}\Tilde{u}_{t} = \beta_2 u_{t-1} + (1-\beta_2)D_0 \mathds{1}_d, \quad \Tilde{v}_{t} = \beta_2 v_{t-1} + (1-\beta_2)D_0 \mathds{1}_d,\nonumber
\end{equation}
From Lemma 6 in \cite{wang2023closing}, we have that $\frac{|\check{r}_{t,l}|}{\sqrt{\check{u}_{t,l}}}, \frac{|r_{t,l}|}{\sqrt{u_{t,l}}}, \frac{|m_{t,l}|}{\sqrt{v_{t,l}}}$ are all upper bounded by $\frac{1-\beta_1}{\sqrt{1-\beta_2} \sqrt{1-\frac{\beta_1^2}{\beta_2}}}$. \\[5pt]
The key idea behind the proof of Theorem 3 is the same as that of Theorem 1. The focus of the proof is the term $(\frac{1}{\sqrt{\beta_2 \Tilde{v}_{t,l}}}-\frac{1}{\sqrt{\Tilde{v}_{t+1,l}}})\nabla f(\check{w}_t)_l^2$. We would first bound $\mathbb{E}(\frac{1}{\sqrt{\beta_2 \Tilde{v}_{t,l}}}-\frac{1}{\sqrt{\Tilde{v}_{t+1,l}}})\nabla f(\check{w}_t)_l^2$ with $\mathbb{E} \frac{\nabla f(\check{w}_t)_l^2}{\sqrt{\Tilde{v}_{t,l}}}$ (see the derivation in the proof of theorem), which acts as Lemma \ref{relemma1} in the proof of Theorem 1, then in reverse bound $\mathbb{E} \frac{\nabla f(\check{w}_t)_l^2}{\sqrt{\Tilde{v}_{t,l}}}$ with $\mathbb{E}(\frac{1}{\sqrt{\beta_2 \Tilde{v}_t}}-\frac{1}{\sqrt{\Tilde{v}_{t+1}}})\nabla f(\check{w}_t)_l^2$ (see Lemma \ref{lemma8} below), which acts as Lemma \ref{relemma2}.

\begin{lemma}\label{lemma7}
    If $f(x)$ in Algorithm 3 satisfies Assumptions 3 and 4, we have that
    \begin{equation}
    \mathbb{E}^{|\mathcal{F}_t} \nabla f(w_t,\xi_t)_l^2 \leq C_0 + 2D_1\nabla f(\check{w}_t)_l^2, \nonumber
    \end{equation}
    where $C_0 = 2D_0 + \frac{8(1-\beta_1)^2\rho^2 L^2}{(1-\beta_2)(1-\frac{\beta_1^2}{\beta_2})}$.
\end{lemma}
\begin{IEEEproof}
\begin{eqnarray}
    \mathbb{E}^{|\mathcal{F}_t} \nabla f(w_t,\xi_t)_l^2 &=& \mathbb{E}^{|\mathcal{F}_t} (\nabla f(w_t,\xi_t)_l - \nabla f(\check{w}_t,\xi_t)_l + \nabla f(\check{w}_t,\xi_t)_l)^2 \nonumber \\
    &\stackrel{(a)}{\leq}& 2L^2 \mathbb{E}^{|\mathcal{F}_t}(w_{t,l}-\check{w}_{t,l})^2 + 2\mathbb{E}^{|\mathcal{F}_t}\nabla f(\check{w}_t,\xi_t)_l^2 \nonumber \\
    &\stackrel{(b)}{\leq}& 4\rho^2 L^2 (\frac{r_{t,l}^2}{u_{t,l}} + \frac{\check{r}_{t,l}^2}{\check{u}_{t,l}}) + 2D_0 + 2D_1\nabla f(\check{w}_t)_l^2 \nonumber \\
    &\leq& (2D_0 + \frac{8(1-\beta_1)^2\rho^2 L^2}{(1-\beta_2)(1-\frac{\beta_1^2}{\beta_2})}) + 2D_1\nabla f(\check{w}_t)_l^2, \nonumber
\end{eqnarray}
where (a) comes from Assumption 3 and (b) comes from Assumption 4.
\end{IEEEproof}

\begin{lemma}\label{lemma10}
    If $f(x)$ in Algorithm 3 satisfies Assumptions 3, we have that
\begin{equation}
    v_{t,l} \leq C_1 u_{t,l},\quad \Tilde{v}_{t,l} \leq C_1 \Tilde{u}_{t,l},
\end{equation}
where the constant $C_1 = \max\{1,2(1-\beta_2)[1+\frac{(1-\beta_1)^2 \rho^2 L^2}{(1-\beta_1^a)(1-\beta_2^b)\epsilon^2}]\}$.
\end{lemma}
\begin{IEEEproof}
\begin{eqnarray}
g_{t,l}^2 &=& (\nabla f(x_t + \rho \frac{r_t}{\sqrt{u_t+\epsilon^2}},\xi_t)_l - \nabla f(x_t,\xi_t)_l + \nabla f(x_t,\xi_t)_l)^2 \nonumber \\
&\leq& \frac{2\rho^2 L^2}{\epsilon^2} r_{t,l}^2 + 2 s_{t,l}^2 \nonumber \\
&=& \frac{2(1-\beta_1)^2\rho^2 L^2}{\epsilon^2} \sum_{\tau=1}^t (\beta_1^{t-\tau} s_{\tau,l})^2 + 2 s_{t,l}^2.
\end{eqnarray}
Thus, we have that
\begin{eqnarray}\label{lemma10_1}
v_{t,l} &=& (1-\beta_2)\sum_{k=1}^t\beta_2^{t-k} g_{k,l}^2 + \beta_2^t \epsilon^2 \nonumber \\
&\leq& \frac{2(1-\beta_1)^2(1-\beta_2)\rho^2 L^2}{\epsilon^2}\sum_{k=1}^t\beta_2^{t-k}\sum_{\tau=1}^k(\beta_1^{k-\tau}s_{\tau,l})^2+2(1-\beta_2)\sum_{k=1}^t \beta_2^{t-k} s_{k,l}^2+\beta_2^t \epsilon^2.
\end{eqnarray}
Since $\beta_1 < \sqrt{\beta_2}$, there exists constants $0<a,b<2$ satisfy that $\beta_1^{2-a} \leq \beta_2^{1+b}$. Then, we have that
\begin{eqnarray}\label{lemma10_2}
    &&\sum_{k=1}^t \beta_2^{t-k} \sum_{\tau=1}^k (\beta_1^{k-\tau} s_{\tau,l})^2 \nonumber \\
    &\leq&  \sum_{k=1}^t \beta_2^{t-k}(\sum_{\tau=1}^k \beta_1^{a(k-\tau)})(\sum_{\tau=1}^k \beta_1^{(2-a)(k-\tau)} s_{\tau,l}^2) \leq \frac{1}{1-\beta_1^a} \sum_{k=1}^t \beta_2^{t-k} \sum_{\tau=1}^k \beta_1^{(2-a)(k-\tau)} s_{\tau,l}^2 \nonumber \\
    &=&\frac{1}{1-\beta_1^a}\sum_{k=1}^t (\sum_{j=0}^{t-k} \beta_1^{(2-a)j} \beta_2^{t-k-j}) s_{k,l}^2 \leq \frac{1}{1-\beta_1^a}\sum_{k=1}^t \beta_2^{t-k}(\sum_{j=0}^{t-k} \beta_2^{bj}) s_{k,l}^2 \nonumber \\
    &\leq& \frac{1}{(1-\beta_1^a)(1-\beta_2^b)}\sum_{k=1}^t \beta_2^{t-k} s_{k,l}^2.
\end{eqnarray}
Substituting (\ref{lemma10_2}) into (\ref{lemma10_1}) yields that
\begin{eqnarray}
    v_{t,l} &\leq& 2(1-\beta_2)[1+\frac{(1-\beta_1)^2 \rho^2 L^2}{(1-\beta_1^a)(1-\beta_2^b)\epsilon^2}]\sum_{k=1}^t \beta_2^{t-k} s_{k,l}^2 + \beta_2^t\epsilon^2 \nonumber \\
    &\leq& C_1 u_{t,l}. \nonumber
\end{eqnarray}
Finally, considering the definition of $\Tilde{v}_{t,l}$, we have that
\begin{equation}
    \Tilde{v}_{t,l} \leq C_1 \Tilde{u}_{t,l}. \nonumber
\end{equation}
\end{IEEEproof}

\begin{lemma}\label{lemma8}
    If $f(x)$ in Algorithm 3 satisfies Assumptions 3 and 4, we have that
\begin{eqnarray}
    &&\frac{1}{8} \sum_{t=1}^{T-1} \sum_{l=1}^d \mathbb{E} \frac{\nabla f(\check{w}_t)_l^2}{\sqrt{\Tilde{v}_{t,l}}} \leq \frac{8\sqrt{\beta_2}D_1}{\beta_2-\beta_1^2}\sum_{t=1}^{T-1} \sum_{l=1}^d \mathbb{E} (\frac{1}{\sqrt{\beta_2 \Tilde{v}_{t,l}}}\!-\!\frac{1}{\sqrt{\Tilde{v}_{t+1,l}}})\nabla f(\check{w}_t)_l^2\!+\!\sum_{l=1}^d \mathbb{E} \frac{2\rho}{(1-\beta_1)\eta} (1\!+\!\frac{2 D_1}{\beta_2-\beta_1^2})\frac{\nabla f(x_T)_l^2}{\sqrt{\Tilde{u}_{T,l}}} \nonumber \\
    &&+C_2 + C_3 \sum_{l=1}^d \mathbb{E}\ln u_{T,l} + \frac{(1-\beta_1)^2 L^2}{(1-\frac{\beta_1}{\sqrt{\beta_2}})^2(1-\beta_2)}(\frac{4 \rho^2}{\sqrt{(1-\beta_2)D_0}}+\rho+\frac{\beta_1\rho}{(1-\beta_1)\beta_2}) \sum_{l=1}^d \mathbb{E} \ln \check{u}_{T,l} \nonumber
\end{eqnarray}
where the constants $C_2$ and $C_3$ are denoted as
\begin{align}
    &C_2 = \frac{(1-\beta_1) d}{(1-\beta_2)(1-\frac{\beta_1^2}{\beta_2})\eta}(2\rho^2 L + (1+\frac{\beta_1^2}{2\beta_2})\sqrt{(1-\beta_2)D_0}\rho) + \frac{2\rho d (1-\beta_1) \sqrt{D_0}}{(1-\frac{\beta_1^2}{\beta_2})\sqrt{1-\beta_2}\eta} + \frac{(1-\beta_1)\rho^2 d L}{(1-\beta_2)\eta} + \frac{\rho}{\sqrt{1-\beta_2}\eta}\|f(x_1)\|_1 \nonumber \\
    &+ \frac{(1-\beta_1)^2 d}{(1-\frac{\beta_1}{\sqrt{\beta_2}})^2(1-\beta_2)}(2L^2(\frac{4\rho^2}{\sqrt{(1-\beta_2)D_0}}+\rho+\frac{\beta_1\rho}{(1-\beta_1)\beta_2})+\frac{35\eta L}{(1-\beta_1)(1-\frac{\beta_1}{\sqrt{\beta_2}})})(-2\ln \epsilon - T \ln \beta_2) + \frac{1-\frac{\beta_1}{\sqrt{\beta_2}}}{(1-\beta_1)\eta} f(p_1) \nonumber \\ 
    &+\frac{d(\ln C_1\!-\!2\ln \epsilon\!-\!T \ln \beta_2)}{1-\beta_2}\bigg(\frac{\rho}{2}+2\sqrt{1-\beta_2}(\frac{C_0}{\sqrt{D_0}}\!+\!\sqrt{D_0}\!+\!\frac{2\beta_1^2 C_0}{(\beta_2-\beta_1^2)\sqrt{D_0}}) \nonumber \\
    &+\frac{(1-\beta_1)^2 }{(1-\frac{\beta_1}{\sqrt{\beta_2}})^2}(\frac{2\beta_1}{1-\beta_1}(\frac{2\beta_1\sqrt{(1-\beta_2)D_0}}{(1-\beta_1)\beta_2}\!+\!\frac{\rho}{4})\!+\!\frac{17\eta L}{(1-\beta_1)(1-\frac{\beta_1}{\sqrt{\beta_2}})})\bigg), \nonumber 
\end{align}
\begin{align}
    &C_3 = \frac{1}{1-\beta_2}\bigg(\frac{\rho}{2}+2\sqrt{1-\beta_2}(\frac{C_0}{\sqrt{D_0}}\!+\!\sqrt{D_0}\!+\!\frac{2\beta_1^2 C_0}{(\beta_2-\beta_1^2)\sqrt{D_0}})\bigg) + \frac{(1-\beta_1)^2}{(1-\frac{\beta_1}{\sqrt{\beta_2}})^2(1-\beta_2)}\bigg(\frac{2\beta_1 }{1-\beta_1}(\frac{2\beta_1\sqrt{(1-\beta_2)D_0}}{(1-\beta_1)\beta_2}+\frac{\rho}{4}) \nonumber \\
    &+\frac{17\eta L}{(1-\beta_1)(1-\frac{\beta_1}{\sqrt{\beta_2}})} + 2L^2(\frac{4\rho^2}{\sqrt{(1-\beta_2)D_0}}+\rho+\frac{\beta_1\rho}{(1-\beta_1)\beta_2})+\frac{35\eta L}{(1-\beta_1)(1-\frac{\beta_1}{\sqrt{\beta_2}})}\bigg). \nonumber
\end{align}
\end{lemma}
\begin{IEEEproof}
From the definition, we have that
\begin{eqnarray}\label{lemma8_0}
     &&p_{t+1,l}-p_{t,l} \nonumber \\
     &=& q_{t+1,l}-q_{t,l} + \frac{\rho}{1-\frac{\beta_1}{\sqrt{\beta_2}}}[(\frac{r_{t+1,l}}{\sqrt{u_{t+1,l}+\epsilon^2}} - \frac{\beta_1}{\sqrt{\beta_2}}\frac{r_{t,l}}{\sqrt{u_{t,l}+\epsilon^2}}) -  (\frac{r_{t,l}}{\sqrt{u_{t,l}+\epsilon^2}} - \frac{\beta_1}{\sqrt{\beta_2}}\frac{r_{t-1,l}}{\sqrt{u_{t-1,l}+\epsilon^2}})] \nonumber \\
     &=& - \frac{\eta}{1-\frac{\beta_1}{\sqrt{\beta_2}}}(\frac{(1-\beta_1)\nabla f(w_t,\xi_t)_l}{\sqrt{v_{t,l}}}+\beta_1 m_{t-1,l}(\frac{1}{\sqrt{v_{t,l}}}-\frac{1}{\sqrt{\beta_2 v_{t-1,l}}})) \nonumber \\
     &&+ \frac{\rho}{1-\frac{\beta_1}{\sqrt{\beta_2}}}[(\frac{r_{t+1,l}}{\sqrt{u_{t+1,l}+\epsilon^2}} - \frac{\beta_1}{\sqrt{\beta_2}}\frac{r_{t,l}}{\sqrt{u_{t,l}+\epsilon^2}}) -  (\frac{r_{t,l}}{\sqrt{u_{t,l}+\epsilon^2}} - \frac{\beta_1}{\sqrt{\beta_2}}\frac{r_{t-1,l}}{\sqrt{u_{t-1,l}+\epsilon^2}})] \nonumber
\end{eqnarray}
According to the $L$-smoothness, we have that
\begin{equation}
    f(p_{t+1}) \leq f(p_t) + \langle \nabla f(w_t), p_{t+1}-p_t \rangle + \langle \nabla f(p_t)-\nabla f(w_t), p_{t+1}-p_t \rangle + \frac{L}{2}\|p_{t+1}-p_t\|^2. \nonumber
\end{equation}
Summing up the above inequality over $\{1,...,T-1\}$ and taking the expectation yields that
\begin{eqnarray}\label{lemma8_1}
    \mathbb{E}[f(p_T)] \leq f(p_1) + \sum_{t=1}^{T-1} \mathbb{E} \langle \nabla f(w_t),p_{t+1}-p_t \rangle + \sum_{t=1}^{T-1} \mathbb{E} \langle \nabla f(p_t)-\nabla f(w_t),p_{t+1}-p_t \rangle + \frac{L}{2}\sum_{t=1}^{T-1}\mathbb{E}\|p_{t+1}-p_t\|^2
\end{eqnarray}
On the RHS of the above inequality, there are five terms that need to be bound (including three terms in $\langle \nabla f(w_t),p_{t+1}-p_t \rangle$ according to (\ref{lemma8_0})). We would analyze them in sequence in the following proof. For the term $\mathbb{E}^{|\mathcal{F}_t} \langle \nabla f(w_t),p_{t+1}-p_t \rangle$, we first have that
\begin{eqnarray}\label{lemma8_2}
    &&-\sum_{l=1}^d \mathbb{E}^{|\mathcal{F}_t} \frac{\nabla f(w_t)_l \nabla f(w_t,\xi_t)_l}{\sqrt{v_{t,l}}} \nonumber \\
    &=& \sum_{l=1}^d \mathbb{E}^{|\mathcal{F}_t} - \frac{\nabla f(\check{w}_t)_l \nabla f(w_t,\xi_t)_l}{\sqrt{v_{t,l}}} + \frac{(\nabla f(\check{w}_t)_l-\nabla f(w_t)_l)\nabla f(w_t,\xi_t)_l}{\sqrt{v_{t,l}}}  \nonumber \\
    &=& \sum_{l=1}^d \mathbb{E}^{|\mathcal{F}_t}- \frac{\nabla f(\check{w}_t)_l \nabla f(\check{w}_t,\xi_t)_l}{\sqrt{\Tilde{v}_{t,l}}} + \frac{\nabla f(\check{w}_t)_l(\nabla f(\check{w}_t,\xi_t)-\nabla f(w_t,\xi_t))}{\sqrt{\Tilde{v}_{t,l}}} + \nabla f(\check{w}_t)_l \nabla f(w_t,\xi_t)_l(\frac{1}{\sqrt{\Tilde{v}_{t,l}}}-\frac{1}{\sqrt{v_{t,l}}}) \nonumber \\
    && + \frac{(\nabla f(\check{w}_t)_l-\nabla f(w_t)_l)\nabla f(w_t,\xi_t)_l}{\sqrt{v_{t,l}}}  \nonumber \\
    &\stackrel{(a)}{\leq}&\sum_{l=1}^d \mathbb{E}^{|\mathcal{F}_t} -\frac{\nabla f(\check{w}_t)_l^2}{\sqrt{\Tilde{v}_{t,l}}} + \frac{\nabla f(\check{w}_t)_l^2}{8\sqrt{\Tilde{v}_{t,l}}} + \frac{2L^2(\check{w}_{t,l}-w_{t,l})^2}{\sqrt{\Tilde{v}_{t,l}}} + \nabla f(\check{w}_t)_l \nabla f(w_t,\xi_t)_l(\frac{1}{\sqrt{\Tilde{v}_{t,l}}}-\frac{1}{\sqrt{v_{t,l}}})\nonumber \\
    &&+ \frac{L^2(\check{w}_{t,l} - w_{t,l})^2}{2\rho} + \frac{\rho \nabla f(w_t,\xi_t)_l^2}{2 v_{t,l}} \nonumber \\
    &\leq& \sum_{l=1}^d \mathbb{E}^{|\mathcal{F}_t} -\frac{7\nabla f(\check{w}_t)_l^2}{8\sqrt{\Tilde{v}_{t,l}}} + \nabla f(\check{w}_t)_l \nabla f(w_t,\xi_t)_l(\frac{1}{\sqrt{\Tilde{v}_{t,l}}}-\frac{1}{\sqrt{v_{t,l}}}) + L^2(\frac{4 \rho^2}{\sqrt{(1-\beta_2)D_0}}+\rho)(\frac{r_{t,l}^2}{u_{t,l}} + \frac{\check{r}_{t,l}^2}{\check{u}_{t,l}}) + \frac{\rho g_{t,l}^2}{2v_{t,l}}, \nonumber \\
\end{eqnarray}
where (a) comes from Assumption 3. Then, we have
\begin{eqnarray}\label{lemma8_3}
\mathbb{E}^{|\mathcal{F}_t}\nabla f(\check{w}_t)_l \nabla f(w_t,\xi_t)_l(\frac{1}{\sqrt{\Tilde{v}_{t,l}}}-\frac{1}{\sqrt{v_{t,l}}}) \leq \mathbb{E}^{|\mathcal{F}_t}\frac{|\nabla f(\check{w}_t)_l||\nabla f(w_t,\xi_t)_l|(1-\beta_2)(D_0+g_{t,l}^2)}{\sqrt{\Tilde{v}_{t,l}} \sqrt{v_{t,l}}(\sqrt{\Tilde{v}_{t,l}}+\sqrt{v_{t,l}})} 
\end{eqnarray}
For the above inequality, we have that
\begin{eqnarray}\label{lemma8_3.1}
    \mathbb{E}^{|\mathcal{F}_t}\frac{|\nabla f(\check{w}_t)_l||g_{t,l}|(1-\beta_2)D_0}{\sqrt{\Tilde{v}_{t,l}} \sqrt{v_{t,l}}(\sqrt{\Tilde{v}_{t,l}}+\sqrt{v_{t,l}})} &\leq& \frac{|\nabla f(\check{w}_t)_l||g_{t,l}|(1-\beta_2)^{1/4} D_0^{1/4}}{\sqrt{v_{t,l}} \Tilde{v}_{t,l}^{1/4}} \nonumber \\
    &\leq& \mathbb{E}^{|\mathcal{F}_t}\frac{\nabla f(\check{w}_t)_l^2}{8\sqrt{\Tilde{v}_{t,l}}} + \mathbb{E}^{|\mathcal{F}_t}2\sqrt{(1-\beta_2)D_0} \frac{g_{t,l}^2}{v_{t,l}}
\end{eqnarray}
and
\begin{eqnarray}\label{lemma8_3.2}
&&\mathbb{E}^{|\mathcal{F}_t}\frac{(1-\beta_2)|\nabla f(\check{w}_t)_l||g_{t,l}|^3}{\sqrt{\Tilde{v}_{t,l}} \sqrt{v_{t,l}}(\sqrt{\Tilde{v}_{t,l}}+\sqrt{v_{t,l}})} \leq \mathbb{E}^{|\mathcal{F}_t}\frac{\sqrt{1-\beta_2}|\nabla f(\check{w}_t)_l|g_{t,l}^2}{\sqrt{\Tilde{v}_{t,l}} (\sqrt{\Tilde{v}_{t,l}}+\sqrt{v_{t,l}})} \nonumber \\
&\stackrel{(b)}{\leq}& \sqrt{1-\beta_2}\frac{|\nabla f(\check{w}_t)_l|}{\sqrt{\Tilde{v}_{t,l}}}(\sqrt{C_0} + \sqrt{2D_1}|\nabla f(\check{w}_t)_l|)\sqrt{\mathbb{E}^{|\mathcal{F}_t} \frac{g_{t,l}^2}{(\sqrt{\Tilde{v}_{t,l}}+\sqrt{v_{t,l}})^2}} \nonumber \\
&\leq& \mathbb{E}^{|\mathcal{F}_t} \frac{\nabla f(\check{w}_t)_l^2}{4\sqrt{\Tilde{v}_{t,l}}} + 2\sqrt{\frac{(1-\beta_2)C_0^2}{D_0}} \frac{g_{t,l}^2}{v_{t,l}} + \frac{8D_1}{\sqrt{\beta_2}}  (\frac{1}{\sqrt{\beta_2 \Tilde{v}_{t,l}}} - \frac{1}{\sqrt{\Tilde{v}_{t+1,l}}})\nabla f(\check{w}_t)_l^2.
\end{eqnarray}
The derivation here follows \cite{wang2023closing}. (b) comes from Cauchy's Inequality and Lemma \ref{lemma7}. Substituting (\ref{lemma8_3}), (\ref{lemma8_3.1}) and (\ref{lemma8_3.2}) into (\ref{lemma8_2}) yields that
\begin{eqnarray}\label{lemma8_4}
    -\sum_{l=1}^d \mathbb{E}^{|\mathcal{F}_t} \frac{\nabla f(w_t)_l \nabla f(w_t,\xi_t)_l}{\sqrt{v_{t,l}}} &\leq& \sum_{l=1}^d \mathbb{E}^{|\mathcal{F}_t} -\frac{\nabla f(\check{w}_t)_l^2}{2\sqrt{\Tilde{v}_{t,l}}} + (\frac{\rho}{2}+2\sqrt{1-\beta_2}(\frac{C_0}{\sqrt{D_0}} + \sqrt{D_0}))\frac{g_{t,l}^2}{v_{t,l}} \nonumber \\
    &&+ L^2(\frac{4 \rho^2}{\sqrt{(1-\beta_2)D_0}}+\rho)(\frac{r_{t,l}^2}{u_{t,l}} + \frac{\check{r}_{t,l}^2}{\check{u}_{t,l}})+\frac{8D_1}{\sqrt{\beta_2}}  (\frac{1}{\sqrt{\beta_2 \Tilde{v}_{t,l}}} - \frac{1}{\sqrt{\Tilde{v}_{t+1,l}}})\nabla f(\check{w}_t)_l^2 \nonumber \\
\end{eqnarray}
Secondly, we have that 
\begin{eqnarray}\label{lemma8_5}
&&\sum_{l=1}^d \mathbb{E}^{|\mathcal{F}_t} \nabla f(w_t)_l (\frac{1}{\sqrt{\beta_2 v_{t-1,l}}}-\frac{1}{\sqrt{v_{t,l}}})m_{t-1,l} \nonumber \\
&\leq& \sum_{l=1}^d \mathbb{E}^{|\mathcal{F}_t} |\nabla f(\check{w}_t)_l (\frac{1}{\sqrt{\beta_2 v_{t-1,l}}}-\frac{1}{\sqrt{\Tilde{v}_{t,l}}})m_{t-1,l}| + |\nabla f(\check{w}_t)_l (\frac{1}{\sqrt{\Tilde{v}_{t,l}}}-\frac{1}{\sqrt{v_{t,l}}})m_{t-1,l}| \nonumber \\
&&+  |(\nabla f(w_t)_l - \nabla f(\check{w}_t)_l) m_{t-1,l} (\frac{1}{\sqrt{\beta_2 v_{t-1,l}}}-\frac{1}{\sqrt{v_{t,l}}})|
\end{eqnarray}
For the above inequality, we have that
\begin{eqnarray}\label{lemma8_6.1}
    \sum_{l=1}^d \mathbb{E}^{|\mathcal{F}_t} |\nabla f(\check{w}_t)_l (\frac{1}{\sqrt{\beta_2 v_{t-1,l}}}-\frac{1}{\sqrt{\Tilde{v}_{t,l}}})m_{t-1,l}| \leq \sum_{l=1}^d \mathbb{E}^{|\mathcal{F}_t}\frac{(1-\beta_1)\nabla f(\check{w}_t)_l^2}{8\beta_1\sqrt{\Tilde{v}_{t,l}}} + \frac{2\beta_1 \sqrt{(1-\beta_2)D_0}}{(1-\beta_1)\beta_2} \frac{m_{t-1,l}^2}{v_{t-1,l}},
\end{eqnarray}
\begin{eqnarray}\label{lemma8_6.2}
    &&\sum_{l=1}^d \mathbb{E}^{|\mathcal{F}_t} |\nabla f(\check{w}_t)_l (\frac{1}{\sqrt{\Tilde{v}_{t,l}}}-\frac{1}{\sqrt{v_{t,l}}})m_{t-1,l}| \nonumber \\
    &\leq& \mathbb{E}^{|\mathcal{F}_t} |\nabla f(\check{w}_t)_l| \frac{(1-\beta_2)(D_0 + g_{t,l}^2)}{\sqrt{\Tilde{v}_{t,l}} \sqrt{\beta_2 v_{t-1,l}} (\sqrt{\Tilde{v}_{t,l}}+\sqrt{v_{t,l}})} |m_{t-1,l}| \nonumber \\
    &\leq& \mathbb{E}^{|\mathcal{F}_t} \frac{1-\beta_1}{\sqrt{\beta_2 - \beta_1^2}} \frac{|\nabla f(\check{w}_t)_l| \sqrt{1-\beta_2}g_{t,l}^2}{\sqrt{\Tilde{v}_{t,l}} (\sqrt{\Tilde{v}_{t,l}}+\sqrt{v_{t,l}})}+\frac{|\nabla f(\check{w}_t)_l|(1-\beta_2)^{1/4} D_0^{1/4} |m_{t-1,l}|}{\Tilde{v}_{t,l}^{1/4} \sqrt{\beta_2 v_{t-1,l}}} \nonumber \\
    &\leq&\mathbb{E}^{|\mathcal{F}_t} \frac{(1-\beta_1)\nabla f(\check{w}_t)_l^2}{8\beta_1\sqrt{\Tilde{v}_{t,l}}} + \frac{4(1-\beta_1)\beta_1\sqrt{1-\beta_2}C_0}{(\beta_2-\beta_1^2)\sqrt{D_0}} \frac{g_{t,l}^2}{v_{t,l}} + \frac{16(1-\beta_1)\beta_1 D_1}{(\beta_2-\beta_1^2)\sqrt{\beta_2}} (\frac{1}{\sqrt{\beta_2 \Tilde{v}_{t,l}}} - \frac{1}{\sqrt{\Tilde{v}_{t+1,l}}})\nabla f(\check{w}_t)_l^2 \nonumber \\
    &&+ \frac{(1-\beta_1)\nabla f(\check{w}_t)_l^2}{8\beta_1\sqrt{\Tilde{v}_{t,l}}} + \frac{2\beta_1 \sqrt{(1-\beta_2)D_0} m_{t-1,l}^2}{(1-\beta_1)\beta_2 v_{t-1,l}}
\end{eqnarray}
and
\begin{eqnarray}\label{lemma8_6.3}
    &&\sum_{l=1}^d \mathbb{E}^{|\mathcal{F}_t} |\nabla f(w_t)_l - \nabla f(\check{w}_t)_l||m_{t-1,l}| \frac{(1-\beta_2)g_{t,l}^2}{\sqrt{\beta_2 v_{t-1,l}} \sqrt{v_{t,l}} (\sqrt{\beta_2 v_{t-1,l}}+\sqrt{v_{t,l}})} \nonumber \\
    &\leq& \sum_{l=1}^d \mathbb{E}^{|\mathcal{F}_t}  \frac{|\nabla f(w_t)_l - \nabla f(\check{w}_t)_l||m_{t-1,l}|}{\sqrt{\beta_2 v_{t-1,l}} } \nonumber \\
    &\stackrel{(c)}{\leq}& \sum_{l=1}^d \mathbb{E}^{|\mathcal{F}_t} \frac{\rho L^2}{\beta_2} (\frac{r_{t,l}^2}{u_{t,l}}+\frac{\check{r}_{t,l}^2}{\check{u}_{t,l}}) + \frac{\rho m_{t-1,l}^2}{2v_{t-1,l}}
\end{eqnarray}
The derivation of (\ref{lemma8_6.2}) uses the same technique as (\ref{lemma8_3.1}) and (\ref{lemma8_3.2}). (c) comes from Assumption 3. Substituting (\ref{lemma8_6.1}), (\ref{lemma8_6.2}) and (\ref{lemma8_6.3}) into (\ref{lemma8_5}) yields that
\begin{eqnarray}\label{lemma8_7}
    &&\sum_{l=1}^d \mathbb{E}^{|\mathcal{F}_t} \nabla f(w_t)_l (\frac{1}{\sqrt{\beta_2 v_{t-1,l}}}-\frac{1}{\sqrt{v_{t,l}}})m_{t-1,l} \leq \frac{3(1-\beta_1)\nabla f(\check{w}_t)_l^2}{8\beta_1\sqrt{\Tilde{v}_{t,l}}} + 2(\frac{2\beta_1\sqrt{(1-\beta_2)D_0}}{(1-\beta_1)\beta_2} +\frac{\rho}{4})\frac{m_{t-1,l}^2}{v_{t-1,l}} \nonumber \\
    &&+ \frac{4(1-\beta_1)\beta_1\sqrt{1-\beta_2}C_0}{(\beta_2-\beta_1^2)\sqrt{D_0}} \frac{g_{t,l}^2}{v_{t,l}} + \frac{16(1-\beta_1)\beta_1 D_1}{(\beta_2-\beta_1^2)\sqrt{\beta_2}} (\frac{1}{\sqrt{\beta_2 \Tilde{v}_{t,l}}} - \frac{1}{\sqrt{\Tilde{v}_{t+1,l}}})\nabla f(\check{w}_t)_l^2 + \frac{\rho L^2}{\beta_2} (\frac{r_{t,l}^2}{u_{t,l}}+\frac{\check{r}_{t,l}^2}{\check{u}_{t,l}})
\end{eqnarray}
Thirdly, we have that
\begin{eqnarray}\label{lemma8_8}
    &&\rho \sum_{t=1}^{T-1} \sum_{l=1}^d \mathbb{E}^{|\mathcal{F}_t} \nabla f(w_t)_l[(\frac{r_{t+1,l}}{\sqrt{u_{t+1,l}+\epsilon^2}} - \frac{\beta_1}{\sqrt{\beta_2}}\frac{r_{t,l}}{\sqrt{u_{t,l}+\epsilon^2}}) -  (\frac{r_{t,l}}{\sqrt{u_{t,l}+\epsilon^2}} - \frac{\beta_1}{\sqrt{\beta_2}}\frac{r_{t-1,l}}{\sqrt{u_{t-1,l}+\epsilon^2}})] \nonumber \\
    &=& \sum_{l=1}^d\mathbb{E}^{|\mathcal{F}_T} \rho \nabla f(w_T)_l(\frac{ r_{T,l}}{\sqrt{u_{T,l}+\epsilon^2}} - \frac{\beta_1}{\sqrt{\beta_2}}\frac{r_{T-1,l}}{\sqrt{u_{T-1,l}+\epsilon^2}}) - \sum_{l=1}^d\mathbb{E}^{|\mathcal{F}_1}\rho \nabla f(w_1)_l \frac{r_{1,l}}{\sqrt{u_{1,l}+\epsilon^2}} \nonumber \\
    &&+ \rho\sum_{t=1}^{T-1}\sum_{l=1}^d\mathbb{E}^{|\mathcal{F}_t}(\nabla f(w_t)_l-\nabla f(w_{t+1})_l)(\frac{ r_{t+1,l}}{\sqrt{u_{t+1,l}+\epsilon^2}} - \frac{\beta_1}{\sqrt{\beta_2}}\frac{r_{t,l}}{\sqrt{u_{t,l}+\epsilon^2}})
\end{eqnarray}
For the above inequality, we respectively have that
\begin{eqnarray}\label{lemma8_8.1}
    &&\sum_{l=1}^d\mathbb{E}^{|\mathcal{F}_T} \rho \nabla f(w_T)_l(\frac{r_{T,l}}{\sqrt{u_{T,l}+\epsilon^2}} - \frac{\beta_1}{\sqrt{\beta_2}}\frac{r_{T-1,l}}{\sqrt{u_{T-1,l}+\epsilon^2}}) \nonumber \\
    &\leq& \sum_{l=1}^d\mathbb{E}^{|\mathcal{F}_T}\rho |\nabla f(w_T)_l\!-\!\nabla f(x_T)_l|(|\frac{r_{T,l}}{\sqrt{u_{T,l}+\epsilon^2}}|\!+\!|\frac{\beta_1}{\sqrt{\beta_2}}\frac{r_{T-1,l}}{\sqrt{u_{T-1,l}+\epsilon^2}}|) + \rho \nabla f(x_T)_l(\frac{r_{T,l}}{\sqrt{u_{T,l}+\epsilon^2}}\!-\!\frac{\beta_1}{\sqrt{\beta_2}}\frac{r_{T-1,l}}{\sqrt{u_{T-1,l}+\epsilon^2}}) \nonumber \\
    &\stackrel{(d)}{\leq}& \sum_{l=1}^d\mathbb{E}^{|\mathcal{F}_T} 2\rho^2 L \frac{(1-\beta_1)^2}{(1-\beta_2)(1-\frac{\beta_1^2}{\beta_2})} + \frac{(1-\beta_1)\rho \nabla f(x_T)_l^2}{\sqrt{\Tilde{u}_{T,l}}} + \rho \nabla f(x_T)_l r_{T,l}(\frac{1}{\sqrt{u_{T,l}+\epsilon^2}} - \frac{1}{\sqrt{\Tilde{u}_{T,l} + \epsilon^2}}) \nonumber \\
    &&+ \beta_1 \rho \nabla f(x_T)_l r_{T-1,l}(\frac{1}{\sqrt{\Tilde{u}_{T,l}+\epsilon^2}}-\frac{1}{\sqrt{\beta_2(u_{T-1,l}+\epsilon^2)}}) 
\end{eqnarray}
Here, (d) comes from Assumption 3, the upper bound of $\frac{r_{t,l}}{\sqrt{u_{t,l}}}$ and $r_{T,l}-\beta_1 r_{T-1,l} = (1-\beta_1) \nabla f(x_T,\xi_T)_l$. Consider that
\begin{eqnarray}\label{lemma8_8.2}
&&\mathbb{E}^{|\mathcal{F}_T}\rho \nabla f(x_T)_l r_{T,l}(\frac{1}{\sqrt{u_{T,l}+\epsilon^2}} - \frac{1}{\sqrt{\Tilde{u}_{T,l} + \epsilon^2}}) \nonumber \\
&\leq& |\rho \nabla f(x_T)_l r_{T,l} \frac{(1-\beta_2)(D_0+s_{T,l}^2)}{\sqrt{u_{T,l}} \sqrt{\Tilde{u}_{T,l}}(\sqrt{u_{T,l}}+\sqrt{\Tilde{u}_{T,l}})}| \nonumber \\
&\leq& \frac{\rho \nabla f(x_T)_l^2}{4\sqrt{\Tilde{u}_{T,l}}} + \sqrt{(1-\beta_2)D_0}\rho \frac{r_{T,l}^2}{u_{T,l}} + \frac{\rho \nabla f(x_T)_l^2}{8 \sqrt{\Tilde{u}_{T,l}}} + \frac{2\rho (1-\beta_1)^2 \sqrt{(1-\beta_2)D_0}}{1-\frac{\beta_1^2}{\beta_2}}\frac{s_T^2}{u_T} \nonumber \\
&&+\frac{\rho \nabla f(x_T)_l^2}{8 \sqrt{\Tilde{u}_{T,l}}} + \frac{4\rho (1-\beta_1)^2 D_1}{(1-\frac{\beta_1^2}{\beta_2})\beta_2}\frac{\nabla f(x_T)_l^2}{\sqrt{\Tilde{u}_{T,l}}} 
\end{eqnarray}
\begin{eqnarray}\label{lemma8_8.3}
    \mathbb{E}^{|\mathcal{F}_T} \beta_1 \rho \nabla f(x_T)_l r_{T-1,l}(\frac{1}{\sqrt{\Tilde{u}_{T,l}+\epsilon^2}}-\frac{1}{\sqrt{\beta_2(u_{T-1,l}+\epsilon^2)}}) &\leq& \beta_1 \rho \nabla f(x_T)_l r_{T-1,l}\frac{(1-\beta_2)^{1/4} D_0^{1/4}}{\Tilde{u}_T^{1/4} \sqrt{\beta_2 u_{T-1,l}}} \nonumber \\
    &\leq& \frac{\rho \nabla f(x_T)_l^2}{2\sqrt{\Tilde{u}_{T,l}}} + \frac{\beta_1^2 \rho \sqrt{(1-\beta_2)D_0}r_{T-1,l}^2}{2\beta_2 u_{T-1,l}}
\end{eqnarray}
Substituting (\ref{lemma8_8.2}) and (\ref{lemma8_8.3}) into (\ref{lemma8_8.1}) yields that 
\begin{eqnarray}\label{lemma8_8.4}
    &&\sum_{l=1}^d\mathbb{E}^{|\mathcal{F}_T} \rho \nabla f(w_T)_l(\frac{r_{T,l}}{\sqrt{u_{T,l}+\epsilon^2}} - \frac{\beta_1}{\sqrt{\beta_2}}\frac{r_{T-1,l}}{\sqrt{u_{T-1,l}+\epsilon^2}}) \nonumber \\
    &\leq& \sum_{l=1}^d\mathbb{E}^{|\mathcal{F}_T} 2\rho (1\!+\!\frac{2(1-\beta_1)^2 D_1}{\beta_2-\beta_1^2})\frac{\nabla f(x_T)_l^2}{\sqrt{\Tilde{u}_{T,l}}}\!+\!\frac{(1-\beta_1)^2}{(1-\beta_2)(1-\frac{\beta_1^2}{\beta_2})}(2\rho^2 L\!+\!(1\!+\!\frac{\beta_1^2}{2\beta_2})\sqrt{(1-\beta_2)D_0}\rho) + \frac{2\rho (1-\beta_1)^2 \sqrt{D_0}}{(1-\frac{\beta_1^2}{\beta_2})\sqrt{1-\beta_2}} \nonumber \\
\end{eqnarray}
Then, we have
\begin{eqnarray}\label{lemma8_8.5}
    - \sum_{l=1}^d\mathbb{E}^{|\mathcal{F}_1}\rho \nabla f(w_1)_l \frac{r_{1,l}}{\sqrt{u_{1,l}+\epsilon^2}} &=&\sum_{l=1}^d\mathbb{E}^{|\mathcal{F}_1} -(\nabla f(x_1+ \frac{\rho r_1}{\sqrt{u_1+\epsilon^2}})_l-f(x_1)_l) \frac{\rho r_{1,l}}{\sqrt{u_{1,l}+\epsilon^2}} - f(x_1)_l \frac{\rho r_{1,l}}{\sqrt{u_{1,l}+\epsilon^2}} \nonumber \\
    &\leq& \sum_{l=1}^d\mathbb{E}^{|\mathcal{F}_1}\rho^2 L \frac{r_{1,l}^2}{u_{1,l}} + \rho |f(x_1)_l| \frac{|r_{1,l}|}{\sqrt{u_{1,l}}} \nonumber \\
    &=& \frac{(1-\beta_1)^2\rho^2 d L}{1-\beta_2} + \frac{(1-\beta_1)\rho}{\sqrt{1-\beta_2}}\|f(x_1)\|_1 
\end{eqnarray}

\begin{eqnarray}\label{lemma8_8.6}
    &&\rho \sum_{t=1}^{T-1}\sum_{l=1}^d\mathbb{E}^{|\mathcal{F}_t}(\nabla f(w_t)_l-\nabla f(w_{t+1})_l)(\frac{ r_{t+1,l}}{\sqrt{u_{t+1,l}+\epsilon^2}} - \frac{\beta_1}{\sqrt{\beta_2}}\frac{r_{t,l}}{\sqrt{u_{t,l}+\epsilon^2}}) \nonumber \\
    &\leq&\sum_{t=1}^{T-1}\sum_{l=1}^d\mathbb{E}^{|\mathcal{F}_t} \frac{L(w_{t,l}-w_{t+1,l})^2}{2} + \rho^2 L(\frac{r_{t+1,l}^2}{u_{t+1,l}} + \frac{\beta_1^2}{\beta_2} \frac{r_{t,l}^2}{u_{t,l}}) \leq \sum_{t=1}^T \sum_{l=1}^d \mathbb{E}^{|\mathcal{F}_t} L(\frac{3\eta^2 m_{t,l}^2}{2 v_{t,l}} + \frac{5 \rho^2 r_{t,l}^2}{u_{t,l}}).
\end{eqnarray}
Substituting (\ref{lemma8_8.4}), (\ref{lemma8_8.5}) and (\ref{lemma8_8.6}) into (\ref{lemma8_8}) yields that
\begin{eqnarray}\label{lemma8_9}
    &&\rho \sum_{t=1}^{T-1} \sum_{l=1}^d \mathbb{E}^{|\mathcal{F}_t} \nabla f(w_t)_l[(\frac{r_{t+1,l}}{\sqrt{u_{t+1,l}+\epsilon^2}} - \frac{\beta_1}{\sqrt{\beta_2}}\frac{r_{t,l}}{\sqrt{u_{t,l}+\epsilon^2}}) -  (\frac{r_{t,l}}{\sqrt{u_{t,l}+\epsilon^2}} - \frac{\beta_1}{\sqrt{\beta_2}}\frac{r_{t-1,l}}{\sqrt{u_{t-1,l}+\epsilon^2}})] \nonumber \\
    &\leq& \frac{(1-\beta_1)^2 d}{(1-\beta_2)(1-\frac{\beta_1^2}{\beta_2})}(2\rho^2 L + (1+\frac{\beta_1^2}{2\beta_2})\sqrt{(1-\beta_2)D_0}\rho) + \frac{2\rho d (1-\beta_1)^2 \sqrt{D_0}}{(1-\frac{\beta_1^2}{\beta_2})\sqrt{1-\beta_2}} + \frac{(1-\beta_1)^2\rho^2 d L}{1-\beta_2} \nonumber \\
    &&+ \frac{(1-\beta_1)\rho}{\sqrt{1-\beta_2}}\|f(x_1)\|_1 + \sum_{l=1}^d \mathbb{E}^{|\mathcal{F}_T} 2\rho (1+\frac{2(1-\beta_1)^2 D_1}{\beta_2-\beta_1^2})\frac{\nabla f(x_T)_l^2}{\sqrt{\Tilde{u}_{T,l}}} + \sum_{t=1}^T \sum_{l=1}^d \mathbb{E}^{|\mathcal{F}_t} L(\frac{3\eta^2 m_{t,l}^2}{2 v_{t,l}} + \frac{5 \rho^2 r_{t,l}^2}{u_{t,l}}).
\end{eqnarray}

For the other two terms, we have 
\begin{eqnarray}\label{lemma8_10}
    \sum_{t=1}^{T-1} \mathbb{E} \langle \nabla f(p_t)-\nabla f(w_t),p_{t+1}-p_t\rangle 
    &\leq& L\sum_{t=1}^{T-1} \mathbb{E}  \|p_t-w_t\|\|p_{t+1}-p_t\|\nonumber \\
    &\leq& \frac{\frac{\beta_1}{\sqrt{\beta_2}}}{1-\frac{\beta_1}{\sqrt{\beta_2}}}L\sum_{t=1}^{T-1} \mathbb{E}\|w_t-w_{t-1}\|\bigg(\frac{\|w_{t+1}-w_t\|}{1-\frac{\beta_1}{\sqrt{\beta_2}}} + \frac{\beta_1}{\sqrt{\beta_2}}\frac{\|w_t-w_{t-1}\|}{1-\frac{\beta_1}{\sqrt{\beta_2}}}\bigg)\nonumber \\
    &\leq& 2L\bigg(\frac{\frac{\beta_1}{\sqrt{\beta_2}}}{1-\frac{\beta_1}{\sqrt{\beta_2}}}\bigg)^2\sum_{t=1}^{T-1}\mathbb{E}\|w_t-w_{t-1}\|^2 + \frac{L}{4(1-\frac{\beta_1}{\sqrt{\beta_2}})^2}\sum_{t=1}^{T-1}\mathbb{E}\|w_{t+1}-w_t\|^2 \nonumber \\
    &\leq& \frac{9L}{(1-\frac{\beta_1}{\sqrt{\beta_2}})^2}\sum_{t=1}^{T}\sum_{l=1}^d\mathbb{E}(\eta^2\frac{m_{t,l}^2}{v_{t,l}} + 2\rho^2\frac{r_{t,l}^2}{u_{t,l}}),
\end{eqnarray}
and
\begin{eqnarray}\label{lemma8_11}
\frac{L}{2}\sum_{t=1}^{T-1}\mathbb{E}\|p_{t+1}-p_t\|^2 
&\leq& \frac{L}{2}\sum_{t=1}^{T-1} \frac{2}{(1-\frac{\beta_1}{\sqrt{\beta_2}})^2} \mathbb{E} \|w_{t+1}-w_t\|^2 + 2\bigg(\frac{\frac{\beta_1}{\sqrt{\beta_2}}}{1-\frac{\beta_1}{\sqrt{\beta_2}}}\bigg)^2 \mathbb{E} \|w_t-w_{t-1}\|^2 \nonumber \\
&\leq& \frac{2L}{(1-\frac{\beta_1}{\sqrt{\beta_2}})^2}\sum_{t=1}^T\mathbb{E}\|w_t-w_{t-1}\|^2 \nonumber \\
&\leq& \frac{6L}{(1-\frac{\beta_1}{\sqrt{\beta_2}})^2}\sum_{t=1}^{T}\sum_{l=1}^d\mathbb{E}(\eta^2\frac{m_{t,l}^2}{v_{t,l}} + 2\rho^2\frac{r_{t,l}^2}{u_{t,l}}),
\end{eqnarray}
In the above two inequalities, we adopt the assumption $\beta_1 < \sqrt{\beta_2}$ when accumulating the terms. Finally, taking the expectation on (\ref{lemma8_4}) and (\ref{lemma8_7}) over $\mathcal{F}_t$ and summing up over $\{1,2,...,T-1\}$. Combing the results with (\ref{lemma8_1}), (\ref{lemma8_9}), (\ref{lemma8_10}) and (\ref{lemma8_11}) yields that
\begin{eqnarray}
    &&\frac{1}{8} \sum_{t=1}^{T-1} \sum_{l=1}^d \mathbb{E} \frac{\nabla f(\check{w}_t)_l^2}{\sqrt{\Tilde{v}_{t,l}}} \leq \frac{1-\frac{\beta_1}{\sqrt{\beta_2}}}{(1-\beta_1)\eta} f(p_1) + \frac{8\sqrt{\beta_2} D_1}{\beta_2-\beta_1^2}\sum_{t=1}^{T-1} \sum_{l=1}^d \mathbb{E} (\frac{1}{\sqrt{\beta_2 \Tilde{v}_{t,l}}}-\frac{1}{\sqrt{\Tilde{v}_{t+1,l}}})\nabla f(\check{w}_t)_l^2\nonumber \\
    &&+ \frac{(1-\beta_1) d}{(1-\beta_2)(1-\frac{\beta_1^2}{\beta_2})\eta}(2\rho^2 L + (1+\frac{\beta_1^2}{2\beta_2})\sqrt{(1-\beta_2)D_0}\rho) + \frac{2\rho d (1-\beta_1) \sqrt{D_0}}{(1-\frac{\beta_1^2}{\beta_2})\sqrt{1-\beta_2}\eta} + \frac{(1-\beta_1)\rho^2 d L}{(1-\beta_2)\eta} + \frac{\rho}{\sqrt{1-\beta_2}\eta}\|f(x_1)\|_1 \nonumber \\
    &&+ (\frac{\rho}{2}+2\sqrt{1-\beta_2}(\frac{C_0}{\sqrt{D_0}}+\sqrt{D_0})+\frac{4\beta_1^2 \sqrt{1-\beta_2}C_0}{(\beta_2-\beta_1^2)\sqrt{D_0}}) \sum_{t=1}^T \sum_{l=1}^d \mathbb{E} \frac{g_{t,l}^2}{v_{t,l}}+L^2(\frac{4 \rho^2}{\sqrt{(1-\beta_2)D_0}}+\rho+\frac{\beta_1\rho}{(1-\beta_1)\beta_2}) \sum_{t=1}^T \sum_{l=1}^d \mathbb{E}\frac{\check{r}_{t,l}^2}{\check{u}_{t,l}}\nonumber \\
    &&+(\frac{2\beta_1 }{1-\beta_1}(\frac{2\beta_1\sqrt{(1-\beta_2)D_0}}{(1-\beta_1)\beta_2}\!+\!\frac{\rho}{4})\!+\!\frac{17\eta L}{(1-\beta_1)(1-\frac{\beta_1}{\sqrt{\beta_2}})}) \sum_{t=1}^T \sum_{l=1}^d \mathbb{E} \frac{m_{t,l}^2}{v_{t,l}}\!+\!\sum_{l=1}^d \mathbb{E} \frac{2\rho}{(1-\beta_1)\eta} (1+\frac{2(1-\beta_1)^2 D_1}{\beta_2-\beta_1^2})\frac{\nabla f(x_T)_l^2}{\sqrt{\Tilde{u}_{T,l}}}\nonumber \\
    &&+ (L^2(\frac{4 \rho^2}{\sqrt{(1-\beta_2)D_0}}+\rho+\frac{\beta_1\rho}{(1-\beta_1)\beta_2})+\frac{35\rho}{(1-\beta_1)(1-\frac{\beta_1}{\sqrt{\beta_2}})\eta}) \sum_{t=1}^T \sum_{l=1}^d \mathbb{E}\frac{r_{t,l}^2}{u_{t,l}} \nonumber 
\end{eqnarray}
Substituting Lemma \ref{ine2} and \ref{lemma10} into the above inequality yields the result.

\end{IEEEproof}

\begin{theorem}
    (Restatement of Theorem 3) If $f(x)$ in Algorithm 3 satisfies Assumptions 3 and 4, and $0\leq \beta_1 < \sqrt{\beta_2} < 1$, $\beta_2 \geq \frac{\sqrt{D_3^2+4D_3}-D_3}{2}$. Then, for any $\beta_2$, perturbation radius $\rho$ and learning rate $\eta$ satisfy that $1-\beta_2 = O(T^{-1})$, $\eta=O(T^{-\frac{1}{2}})$, $\rho=O(T^{-\frac{1}{2}})$, we have the convergence rate
\begin{equation}
    \frac{1}{T} \sum_{t=1}^T \mathbb{E} \|\nabla f(x_t)\|_1 \leq O\bigg(\frac{\ln T}{T^{1/4}}\bigg), \nonumber
\end{equation}
where the constant $D_3$ satisfies that \\[10pt]
$D_3 = \max\{4\sqrt{\beta_2}, \frac{256\sqrt{\beta_2}D_1}{\beta_2-\beta_1^2}, \frac{2048\sqrt{C_1} D_1 \rho}{(1-\beta_1)(1-\frac{\beta_1^2}{\beta_2})\sqrt{\beta_2}\eta}(1+\frac{2D_1}{\beta_2-\beta_1^2})\}.$
\end{theorem}

\begin{IEEEproof}
From the definition, we have that
\begin{eqnarray}
     q_{t+1,i}-q_{t,i} = -\eta \frac{1-\beta_1}{1-\frac{\beta_1}{\sqrt{\beta_2}}}\frac{g_{t,i}}{\sqrt{\Tilde{v}_{t,i}}} - \eta \frac{1}{1-\frac{\beta_1}{\sqrt{\beta_2}}}(\frac{1}{\sqrt{v_{t,i}}}-\frac{1}{\sqrt{\Tilde{v}_{t,i}}})m_{t,i} + \eta \frac{\beta_1}{1-\frac{\beta_1}{\sqrt{\beta_2}}}(\frac{1}{\sqrt{\beta_2 v_{t-1,i}}}-\frac{1}{\sqrt{\Tilde{v}_t}})m_{t-1,i},  
\end{eqnarray}
Considering the $L$-smoothness of $f(x)$, we further obtain that
\begin{eqnarray}
    &&\mathbb{E}^{|\mathcal{F}_t}[f(q_{t+1})] \nonumber \\
    &\leq& f(q_t) + \mathbb{E}^{|\mathcal{F}_t}\langle \nabla f(q_t), q_{t+1}-q_t \rangle + \frac{L}{2}\mathbb{E}^{|\mathcal{F}_t}\|q_{t+1}-q_t\|^2 \nonumber \\
    &=& f(q_t) -\eta \frac{1-\beta_1}{1-\frac{\beta_1}{\sqrt{\beta_2}}}\mathbb{E}^{|\mathcal{F}_t} \langle \nabla f(x_t), \frac{1}{\sqrt{\Tilde{v}_t}} \odot g_t \rangle - \eta \frac{1}{1-\frac{\beta_1}{\sqrt{\beta_2}}} \mathbb{E}^{|\mathcal{F}_t}\langle \nabla f(x_t), (\frac{1}{\sqrt{v_t}}-\frac{1}{\sqrt{\Tilde{v}_t}}) \odot m_t \rangle \nonumber \\
    &&+ \eta \frac{\beta_1}{1-\frac{\beta_1}{\sqrt{\beta_2}}}\mathbb{E}^{|\mathcal{F}_t}\langle \nabla f(x_t), (\frac{1}{\sqrt{\beta_2 v_{t-1}}}-\frac{1}{\sqrt{\Tilde{v}_t}}) \odot m_{t-1}\rangle + \mathbb{E}^{|\mathcal{F}_t}\langle \nabla f(q_t)-\nabla f(x_t),q_{t+1}-q_t\rangle + \frac{L}{2}\mathbb{E}^{|\mathcal{F}_t}\|q_{t+1}-q_t\|^2, \nonumber
\end{eqnarray}
Taking the expectation over $\mathcal{F}_t$ and summing up the above inequality over $\{1,...,T\}$ yields that 
\begin{eqnarray}\label{theo2_1}
    &&\mathbb{E}[f(q_{T+1})] - f(q_1) \nonumber \\
    &\leq&- \frac{\eta(1-\beta_1)}{1-\frac{\beta_1} {\sqrt{\beta_2}}}\sum_{t=1}^T \sum_{l=1}^d \mathbb{E}\frac{\nabla f(x_t)_l g_{t,l}}{\sqrt{\Tilde{v}_{t,l}}} - \frac{\eta}{1-\frac{\beta_1}{\sqrt{\beta_2}}} \sum_{t=1}^T \sum_{l=1}^d \mathbb{E}\nabla f(x_t)_l m_{t,l}(\frac{1}{\sqrt{v_{t,l}}}-\frac{1}{\sqrt{\Tilde{v}_{t,l}}}) \nonumber \\    
    &&+ \frac{\eta \beta_1}{1-\frac{\beta_1}{\sqrt{\beta_2}}} \sum_{t=1}^T \sum_{l=1}^d \mathbb{E}\nabla f(x_t)_l m_{t-1,l}(\frac{1}{\sqrt{\beta_2 v_{t-1,l}}}-\frac{1}{\sqrt{\Tilde{v}_{t,l}}}) + \frac{L}{2}\sum_{t=1}^T \mathbb{E}\|q_{t+1}-q_t\|^2 \nonumber \\
    &&+ \sum_{t=1}^T \mathbb{E}\langle \nabla f(q_t)-\nabla f(x_t),q_{t+1}-q_t\rangle, 
\end{eqnarray}
Firstly, similar to (\ref{theo_2}), we obtain that
\begin{eqnarray}
\label{theo2_2}
    - \sum_{t=1}^T \sum_{l=1}^d \mathbb{E}\frac{\nabla f(x_t)_l g_{t,l}}{\sqrt{\Tilde{v}_{t,l}}} 
    \leq -\frac{3}{4} \sum_{t=1}^T\sum_{l=1}^d \mathbb{E}\frac{\nabla f(x_t)_l^2}{\sqrt{\Tilde{v}_{t,l}}} + \frac{\rho^2 L^2}{ \sqrt{(1-\beta_2)D_0}} \sum_{t=1}^T \sum_{l=1}^d \mathbb{E}(\frac{4r_{t,l}^2}{u_{t,l}}+\frac{6\check{r}_{t,l}^2}{\check{u}_{t,l}}) \nonumber \\
\end{eqnarray}
Secondly, following the derivation in \cite{wang2023closing}, we have
\begin{equation}\label{theo2_3}
\sum_{t=1}^T \sum_{l=1}^d \mathbb{E} \nabla f(x_t)_l m_{t,l}(\frac{1}{\sqrt{\Tilde{v}_{t,l}}}-\frac{1}{\sqrt{v_{t,l}}}) 
\leq \sum_{t=1}^T \sum_{l=1}^d \mathbb{E} |\nabla f(x_t)_l| |m_{t,l}| \frac{(1-\beta_2)(g_{t,l}^2+D_0)}{\sqrt{v_{t,l}} \sqrt{\Tilde{v}_{t,l}}(\sqrt{v_{t,l}} +\sqrt{\Tilde{v}_{t,l}})}
\end{equation}
For the above inequality, by Lemma \ref{lemma7}, we have
\begin{eqnarray}\label{theo2_4}
    &&\sum_{t=1}^T \sum_{l=1}^d \mathbb{E} |\nabla f(x_t)_l| |m_{t,l}| \frac{(1-\beta_2)g_{t,l}^2}{\sqrt{v_{t,l}} \sqrt{\Tilde{v}_{t,l}}(\sqrt{v_{t,l}} +\sqrt{\Tilde{v}_{t,l}})} \leq \frac{1-\beta_1}{4}\sum_{t=1}^T \sum_{l=1}^d \mathbb{E} \frac{\nabla f(x_t)_l^2}{\sqrt{\Tilde{v}_{t,l}}} + \frac{2(1-\beta_1)\sqrt{1-\beta_2}C_0}{(1-\frac{\beta_1^2}{\beta_2})\sqrt{D_0}} \sum_{t=1}^T \sum_{l=1}^d\mathbb{E} \frac{g_{t,l}^2}{v_{t,l}}\nonumber \\
    && + \frac{8(1-\beta_1) D_1}{(1-\frac{\beta_1^2}{\beta_2})\sqrt{\beta_2}}\sum_{t=1}^T \sum_{l=1}^d\mathbb{E} (\frac{1}{\sqrt{\beta_2 \Tilde{v}_{t,l}}} - \frac{1}{\sqrt{\Tilde{v}_{t+1,l}}})\nabla f(\check{w}_t)_l^2 \nonumber \\
\end{eqnarray}
Further, by $\|x\|^2 - \|y\|^2 \leq 2\|x-y\|\|x\| + \|x\|^2 + \|y\|^2 - 2\langle x,y \rangle = 2\|x-y\|\|x\| + \|x-y\|^2$, we have
\begin{eqnarray}
    \sum_{t=2}^T \sum_{l=1}^d \mathbb{E} \frac{\nabla f(\check{w}_t)_l^2}{\sqrt{\beta_2 \Tilde{v}_{t,l}}}
    &\leq& \sum_{t=2}^T \sum_{l=1}^d \mathbb{E} \frac{\nabla f(\check{w}_{t-1})_l^2}{\sqrt{\beta_2 \Tilde{v}_{t,l}}}  + \frac{1}{D_3}\sum_{t=2}^T \sum_{l=1}^d \mathbb{E} \frac{\nabla f(\check{w}_t)_l^2}{\sqrt{\Tilde{v}_{t,l}}} + \frac{(D_3+1)L^2}{\beta_2\sqrt{(1-\beta_2) D_0}}\sum_{t=2}^T \mathbb{E}\|\check{w}_t-\check{w}_{t-1}\|^2 \nonumber \\
    &\leq& \sum_{t=2}^T \sum_{l=1}^d \mathbb{E} \frac{\nabla f(\check{w}_{t-1})_l^2}{\sqrt{\beta_2 \Tilde{v}_{t,l}}} + \frac{1}{D_3}\sum_{t=2}^T \sum_{l=1}^d \mathbb{E} \frac{\nabla f(\check{w}_t)_l^2}{\sqrt{\Tilde{v}_{t,l}}}  \nonumber \\
    &&+ \frac{3(D_3+1)L^2}{\beta_2\sqrt{(1-\beta_2) D_0}}\sum_{t=1}^T \sum_{l=1}^d \mathbb{E}(\eta^2 \frac{m_{t,l}^2}{v_t} + 2\rho^2 \frac{\check{r}_{t,l}^2}{\check{u}_{t,l}})
\end{eqnarray}
Thus, we have
\begin{eqnarray}\label{theo2_5}
    &&\sum_{t=1}^T \sum_{l=1}^d\mathbb{E} (\frac{1}{\sqrt{\beta_2 \Tilde{v}_{t,l}}} - \frac{1}{\sqrt{\Tilde{v}_{t+1,l}}})\nabla f(\check{w}_t)_l^2 \nonumber \\
    &\leq&\sum_{l=1}^d \mathbb{E}(\frac{1}{\sqrt{\beta_2\Tilde{v}_1}}-\frac{1}{\sqrt{\Tilde{v}_2}})\nabla f(\check{w}_1)_l^2 + \sum_{t=2}^T \sum_{l=1}^d\mathbb{E} (\frac{\nabla f(\check{w}_{t-1})_l^2}{\sqrt{\beta_2 \Tilde{v}_{t,l}}} - \frac{\nabla f(\check{w}_t)_l^2}{\sqrt{\Tilde{v}_{t+1,l}}}) 
    + \frac{1}{D_3}\sum_{t=2}^T \sum_{l=1}^d \mathbb{E} \frac{\nabla f(\check{w}_t)_l^2}{\sqrt{\Tilde{v}_{t,l}}} \nonumber \\
    &&+ \frac{3(D_3+1)L^2}{\beta_2\sqrt{(1-\beta_2) D_0}}\sum_{t=1}^T \sum_{l=1}^d \mathbb{E}(\eta^2 \frac{m_{t,l}^2}{v_t} + 2\rho^2 \frac{\check{r}_{t,l}^2}{\check{u}_{t,l}}) \nonumber\\
    &\stackrel{(a)}{\leq}& \frac{||\nabla f(\check{w}_1)||^2}{\sqrt{(1-\beta_2)\beta_2 D_0}} + (\frac{1}{\beta_2}-\frac{1}{\sqrt{\beta_2}} + \frac{1}{D_3}) \sum_{t=1}^T \sum_{l=1}^d \mathbb{E}\frac{\nabla f(\check{w}_t)_l^2}{\sqrt{\Tilde{v}_{t,l}}} -\sum_{l=1}^d \mathbb{E}\frac{\nabla f(\check{w}_T)_l^2}{\sqrt{\Tilde{v}_{T+1,l}}} \nonumber \\
    &&+\frac{3(D_3+1)L^2}{\beta_2\sqrt{(1-\beta_2) D_0}}\sum_{t=1}^T \sum_{l=1}^d \mathbb{E}(\eta^2 \frac{m_{t,l}^2}{v_t} + 2\rho^2 \frac{\check{r}_{t,l}^2}{\check{u}_{t,l}}),
\end{eqnarray}
where (a) comes from the fact that $\Tilde{v}_{t+1,l} \geq \beta_2 \Tilde{v}_{t,l}$. Since $\beta_2 \geq \frac{\sqrt{D_3^2+4D_3}-D_3}{2}$, we have $\frac{1}{\beta_2}-\frac{1}{\sqrt{\beta_2}} + \frac{1}{D_3} \leq \frac{2}{D_3}$. Thus, we have
\begin{eqnarray}\label{theo2_54}
    &&\sum_{t=1}^T \sum_{l=1}^d\mathbb{E} (\frac{1}{\sqrt{\beta_2 \Tilde{v}_{t,l}}} - \frac{1}{\sqrt{\Tilde{v}_{t+1,l}}})\nabla f(\check{w}_t)_l^2 \nonumber \\
    &\leq& \frac{||\nabla f(\check{w}_1)||^2}{\sqrt{(1-\beta_2)\beta_2 D_0}} + \frac{2}{D_3}\sum_{t=1}^{T-1} \sum_{l=1}^d \mathbb{E}\frac{\nabla f(\check{w}_t)_l^2}{\sqrt{\Tilde{v}_{t,l}}} + \sum_{l=1}^d (\frac{2}{D_3 \sqrt{\Tilde{v}_{T,l}}} - \frac{1}{\sqrt{\Tilde{v}_{T+1,l}}})\nabla f(\check{w}_T)_l^2 \nonumber \\
    &&+\frac{3(D_3+1)L^2}{\beta_2\sqrt{(1-\beta_2) D_0}}\sum_{t=1}^T \sum_{l=1}^d \mathbb{E}(\eta^2 \frac{m_{t,l}^2}{v_t} + 2\rho^2 \frac{\check{r}_{t,l}^2}{\check{u}_{t,l}})
\end{eqnarray}
Substituting Lemma \ref{lemma8} into (\ref{theo2_54}), considering $D_3 \geq \frac{256\sqrt{\beta_2}D_1}{\beta_2-\beta_1^2}$ and $\frac{2}{D_3 \sqrt{\Tilde{v}_{T,l}}}\!-\!\frac{1}{\sqrt{\Tilde{v}_{T+1,l}}} \leq \frac{1}{2}(\frac{1}{\sqrt{\beta_2 \Tilde{v}_{T,l}}}-\frac{1}{\sqrt{\Tilde{v}_{T+1,l}}})$ which comes from $D_3 \geq 4\sqrt{\beta_2}$, we have
\begin{eqnarray}\label{theo2_55}
    &&\sum_{t=1}^T \sum_{l=1}^d\mathbb{E} (\frac{1}{\sqrt{\beta_2 \Tilde{v}_{t,l}}} - \frac{1}{\sqrt{\Tilde{v}_{t+1,l}}})\nabla f(\check{w}_t)_l^2 \nonumber \\
    &\leq& \frac{1}{2}\sum_{t=1}^T \sum_{l=1}^d\mathbb{E} (\frac{1}{\sqrt{\beta_2 \Tilde{v}_{t,l}}}\!-\!\frac{1}{\sqrt{\Tilde{v}_{t+1,l}}})\nabla f(\check{w}_t)_l^2 +\frac{||\nabla f(\check{w}_1)||^2}{\sqrt{(1-\beta_2)\beta_2 D_0}} + \frac{16C_2}{D_3} + \frac{16C_3}{D_3} \sum_{l=1}^d \mathbb{E}\ln u_{T,l} \nonumber \\
    &&+ \frac{16(1-\beta_1)^2}{(1-\frac{\beta_1}{\sqrt{\beta_2}})^2(1-\beta_2)D_3}L^2(\frac{4 \rho^2}{\sqrt{(1-\beta_2)D_0}}+\rho+\frac{\beta_1\rho}{(1-\beta_1)\beta_2}) \sum_{l=1}^d \mathbb{E} \ln \check{u}_{T,l} \nonumber \\
    &&+ \frac{3(1-\beta_1)^2(D_3+1)L^2}{(1-\frac{\beta_1}{\sqrt{\beta_2}})^2\beta_2\sqrt{(1-\beta_2)^3 D_0}}\sum_{l=1}^d \mathbb{E}(\eta^2 (\ln v_{t,l}-2\ln \epsilon-T\ln \beta_2) + 2\rho^2 (\ln \check{u}_{t,l}-2\ln \epsilon-T\ln \beta_2)) \nonumber \\
    &&+\frac{32\rho}{D_3(1-\beta_1)\eta}(1+\frac{2D_1}{\beta_2-\beta_1^2})\sum_{l=1}^d \mathbb{E}\frac{\nabla f(x_T)_l^2}{\sqrt{\Tilde{u}_{T,l}}}
\end{eqnarray}
Rearranging (\ref{theo2_55}) and considering $D_3 \geq \frac{2048\sqrt{C_1} D_1 \rho}{(1-\beta_1)(1-\frac{\beta_1^2}{\beta_2})\sqrt{\beta_2}\eta}(1+\frac{2D_1}{\beta_2-\beta_1^2})$, then substituting the result into (\ref{theo2_4}) yields that
\begin{eqnarray}\label{theo2_6}
    &&\sum_{t=1}^T \sum_{l=1}^d \mathbb{E} |\nabla f(x_t)_l| |m_{t,l}| \frac{(1-\beta_2)g_{t,l}^2}{\sqrt{v_{t,l}} \sqrt{\Tilde{v}_{t,l}}(\sqrt{v_{t,l}} +\sqrt{\Tilde{v}_{t,l}})} \nonumber \\
    &\leq& \frac{(1-\beta_1)}{4}\sum_{t=1}^T \sum_{l=1}^d \mathbb{E} \frac{\nabla f(x_t)_l^2}{\sqrt{\Tilde{v}_{t,l}}} + \frac{1-\beta_1}{8\sqrt{C_1}}\sum_{t=1}^T \sum_{l=1}^d \mathbb{E} \frac{\nabla f(x_t)_l^2}{\sqrt{\Tilde{u}_t}} + \frac{16(1-\beta_1)D_1 ||\nabla f(\check{w}_1)||^2}{(1-\frac{\beta_1^2}{\beta_2})\beta_2 \sqrt{(1-\beta_2) D_0}} + \frac{128D_1 C_2}{(1-\frac{\beta_1^2}{\beta_2})\sqrt{\beta_2}D_3} \nonumber \\
    &&+ \frac{2(1-\beta_2)d C_0 \ln C_1 - d(2(1-\beta_2)C_0 +48(D_3+1)(\eta^2+2\rho^2)L^2)(2\ln \epsilon+T \ln \beta_2)}{(1-\frac{\beta_1}{\sqrt{\beta_2}})^3\sqrt{(1-\beta_2)^3 \beta_2^3 D_0}}\nonumber \\
    &&+ \frac{2(1-\beta_1)}{1-\frac{\beta_1^2}{\beta_2}}\bigg(\frac{C_0}{\sqrt{(1-\beta_2)D_0}}+\frac{8D_1}{\sqrt{\beta_2}}(\frac{16C_3}{D_3}+\frac{3(1-\beta_1)^2(D_3+1)\eta^2 L^2}{(1-\frac{\beta_1}{\sqrt{\beta_2}})^2\beta_2\sqrt{(1-\beta_2)^3 D_0}})\bigg)\sum_{l=1}^d \mathbb{E}\ln u_{T,l} \nonumber \\
    &&+ \frac{16(1-\beta_1)^3D_1}{(1-\frac{\beta_1^2}{\beta_2})(1-\frac{\beta_1}{\sqrt{\beta_2}})^2(1-\beta_2)\sqrt{\beta_2}}\bigg(\frac{16L^2}{D_3}(\frac{4\rho^2}{\sqrt{(1-\beta_2)D_0}}+\rho+\frac{\beta_1\rho}{(1-\beta_1)\beta_2}) + \frac{6(D_3+1)\rho^2 L^2}{\beta_2\sqrt{(1-\beta_2)D_0}}\bigg)\sum_{l=1}^d \mathbb{E} \ln \check{u}_{T,l} \nonumber \\
\end{eqnarray}
Then, we have
\begin{eqnarray}\label{theo2_7}
&&\sum_{t=1}^T \sum_{l=1}^d \mathbb{E} |\nabla f(x_t)_l| |m_{t,l}| \frac{(1-\beta_2)D_0}{\sqrt{v_{t,l}} \sqrt{\Tilde{v}_{t,l}}(\sqrt{v_{t,l}}+\sqrt{\Tilde{v}_{t,l}})} \nonumber \\
&\leq& \frac{1-\beta_1}{16}\sum_{t=1}^T \sum_{l=1}^d \mathbb{E} \frac{\nabla f(x_t)_l^2}{\sqrt{\Tilde{v}_{t,l}}} + \frac{4\sqrt{(1-\beta_2)D_0}}{1-\beta_1} \sum_{t=1}^T \sum_{l=1}^d \mathbb{E}\frac{m_{t,l}^2}{v_{t,l}}
\end{eqnarray}
Thirdly, we respectively have
\begin{eqnarray}\label{theo2_9}
    \beta_1\sum_{t=1}^T \sum_{l=1}^d \mathbb{E}\nabla f(x_t)_l m_{t-1,l}(\frac{1}{\sqrt{\beta_2 v_{t-1,l}}}-\frac{1}{\sqrt{\Tilde{v}_{t,l}}}) \leq \frac{1-\beta_1}{16} \sum_{t=1}^T \sum_{l=1}^d \mathbb{E}\frac{\nabla f(x_t)_l^2}{\sqrt{\Tilde{v}_{t,l}}} + \frac{4\beta_1^2 \sqrt{(1-\beta_2)D_0}}{(1-\beta_1)\beta_2} \sum_{t=1}^T \sum_{l=1}^d \mathbb{E} \frac{m_{t-1,l}^2}{v_{t-1,l}}
\end{eqnarray}
and
\begin{eqnarray}\label{theo2_10}
    &&\sum_{t=1}^T \mathbb{E} \langle \nabla f(q_t)-\nabla f(x_t),q_{t+1}-q_t\rangle +\frac{L}{2}\mathbb{E}\|q_{t+1}-q_t\|^2\nonumber \\
    &\leq&\eta^2 L\bigg(\frac{5}{2}\bigg(\frac{\frac{\beta_1}{\sqrt{\beta_2}}}{1-\frac{\beta_1}{\sqrt{\beta_2}}}\bigg)^2 \sum_{t=1}^T \sum_{l=1}^d\mathbb{E}\frac{m_{t-1,l}^2}{v_{t-1,l}} + \frac{3}{2}\bigg(\frac{1}{1-\frac{\beta_1}{\sqrt{\beta_2}}}\bigg)^2\sum_{t=1}^T \sum_{l=1}^d\mathbb{E}\frac{m_{t,l}^2}{v_{t,l}}\bigg) \nonumber \\
    &\leq& \frac{4\eta^2 L}{(1-\frac{\beta_1}{\sqrt{\beta_2}})^2} \sum_{t=1}^T \sum_{l=1}^d\mathbb{E}\frac{m_{t,l}^2}{v_{t,l}}.
\end{eqnarray}
Next, substituting (\ref{theo2_2}), (\ref{theo2_6}) and (\ref{theo2_7}), (\ref{theo2_9}) and (\ref{theo2_10}) into (\ref{theo2_1}), then combing the result with $\frac{\nabla f(x_t)_l^2}{\sqrt{\Tilde{v}_{t,l}}} \geq \frac{1}{\sqrt{C_1}} \frac{\nabla f(x_t)_l^2}{\sqrt{\Tilde{u}_{t,l}}}$ which comes from Lemma \ref{lemma10} yields that
\begin{equation}
    \sum_{t=1}^T \sum_{l=1}^d \mathbb{E} \frac{\nabla f(x_t)_l^2}{\sqrt{\Tilde{u}_{t,l}}} \leq  C_4 +C_5 \sum_{l=1}^d \mathbb{E} \ln \check{u}_{T,l} + C_6 \sum_{l=1}^d \mathbb{E} \ln u_{T,l}
\end{equation}
where the constants $C_4$, $C_5$ and $C_6$ are as follows
\begin{align}
    &C_4=4\sqrt{C_1}\bigg[\bigg(\frac{2C_0}{(1-\frac{\beta_1}{\sqrt{\beta_2}})^3\sqrt{(1-\beta_2) \beta_2^3 D_0}}+ \frac{8\sqrt{(1-\beta_2)D_0}}{(1-\beta_1)^2} + \frac{4\eta L}{(1-\frac{\beta_1}{\sqrt{\beta_2}})(1-\beta_1)}\bigg)d \ln C_1\nonumber \\
    &-\bigg(\frac{2(1-\beta_2)C_0 +48(D_3+1)(\eta^2+2\rho^2)L^2}{(1-\frac{\beta_1}{\sqrt{\beta_2}})^3\sqrt{(1-\beta_2)^3 \beta_2^3 D_0}}+\frac{40\rho^2 L^2}{\sqrt{(1-\beta_2)D_0}}+\frac{8\sqrt{(1-\beta_2)D_0}}{(1-\beta_1)^2} + \frac{4\eta L}{(1-\frac{\beta_1}{\sqrt{\beta_2}})(1-\beta_1)}\bigg)d(2\ln \epsilon + T \ln \beta_2) \nonumber \\
    &+ \frac{1-\frac{\beta_1}{\sqrt{\beta_2}}}{(1-\beta_1)\eta}f(q_1) + \frac{16D_1 ||\nabla f(\check{w}_1)||^2}{(1-\frac{\beta_1^2}{\beta_2})\beta_2 \sqrt{(1-\beta_2) D_0}}+\frac{128D_1 C_2}{(1-\beta_1)(1-\frac{\beta_1^2}{\beta_2})\sqrt{\beta_2}D_3}\bigg]\nonumber 
\end{align}
\begin{eqnarray}
    C_5 &=& 4\sqrt{C_1}\bigg[\frac{6\rho^2 L^2}{\sqrt{(1-\beta_2)D_0}}+\frac{16(1-\beta_1)^2D_1}{(1-\frac{\beta_1^2}{\beta_2})(1-\frac{\beta_1}{\sqrt{\beta_2}})^2(1-\beta_2)\sqrt{\beta_2}}\bigg(\frac{16L^2}{D_3}(\frac{4\rho^2}{\sqrt{(1-\beta_2)D_0}}+\rho+\frac{\beta_1\rho}{(1-\beta_1)\beta_2}) \nonumber \\
    &&+ \frac{6(D_3+1)\rho^2 L^2}{\beta_2\sqrt{(1-\beta_2)D_0}}\bigg)\bigg],\nonumber
\end{eqnarray}
\begin{eqnarray}
    C_6 &=& 4\sqrt{C_1}\bigg[\frac{4\rho^2 L^2}{\sqrt{(1-\beta_2)D_0}} + \frac{2}{1-\frac{\beta_1^2}{\beta_2}}\bigg(\frac{C_0}{\sqrt{(1-\beta_2)D_0}}+\frac{8D_1}{\sqrt{\beta_2}}(\frac{16C_3}{D_3}+\frac{3(1-\beta_1)^2(D_3+1)\eta^2 L^2}{(1-\frac{\beta_1}{\sqrt{\beta_2}})^2\beta_2\sqrt{(1-\beta_2)^3 D_0}})\bigg) \nonumber \\
    &&+\frac{8\sqrt{(1-\beta_2)D_0}}{(1-\beta_1)^2} + \frac{4\eta L}{(1-\frac{\beta_1}{\sqrt{\beta_2}})(1-\beta_1)}\bigg]. \nonumber
\end{eqnarray}
Here, we also utilize Lemma \ref{ine2} and \ref{lemma10} which indicates $v_t \leq C_1 u_t$. Then, we follow Lemma 9 in \cite{wang2023closing} to obtain that
\begin{eqnarray}
    \sum_{t=1}^{T+1} \sum_{l=1}^d \mathbb{E}\sqrt{\Tilde{u}_{t,l}} &\leq& \frac{3(1+\sqrt{\beta_2})D_1}{\sqrt{\beta_2}}(C_4 +C_5 \sum_{l=1}^d \mathbb{E} \ln \check{u}_{T,l} + C_6 \sum_{l=1}^d \mathbb{E} \ln u_{T,l})+(T+1)d\sqrt{D_0+\epsilon^2} \nonumber \\
    &\stackrel{(b)}{\leq}&  \frac{3(1+\sqrt{\beta_2})D_1}{\sqrt{\beta_2}}(C_4 +2dC_5 \ln \mathbb{E} \sum_{l=1}^d  \sqrt{\check{u}_{T,l}} - 2d C_5 \ln d) + (T+1)d\sqrt{D_0+\epsilon^2} \nonumber \\
    &&+ \frac{6(1+\sqrt{\beta_2})d D_1 C_6}{\beta_2} (\ln \sum_{t=1}^{T+1} \sum_{l=1}^d \mathbb{E}\sqrt{\Tilde{u}_{t,l}} - \ln d), \nonumber
\end{eqnarray}
where (b) holds since 
\begin{equation}
\sum_{l=1}^d \mathbb{E} \ln \check{u}_{T,l} = 2\mathbb{E} \sum_{l=1}^d \ln \sqrt{\check{u}_{T,l}}\leq 2d \mathbb{E}\ln \frac{\sum_{l=1}^d \sqrt{\check{u}_{T,l}}}{d} \leq 2d(\ln \mathbb{E} \sum_{l=1}^d \sqrt{\check{u}_{T,l}}- \ln d), \nonumber
\end{equation}
and
\begin{eqnarray}
\sum_{l=1}^d \mathbb{E}\ln u_{T,l} &\leq& \frac{2}{\sqrt{\beta_2}}\mathbb{E} \sum_{l=1}^d \ln \sqrt{\Tilde{u}_{T+1,l}} \leq \frac{2d}{\sqrt{\beta_2}}\mathbb{E} \ln \frac{\sum_{l=1}^d \sqrt{\Tilde{u}_{T+1,l}}}{d} \leq \frac{2d}{\sqrt{\beta_2}} \mathbb{E}(\ln \sum_{t=1}^{T+1} \sum_{l=1}^d \sqrt{\Tilde{u}_{t,l}} - \ln d) \nonumber \\
&\leq& \frac{2d}{\sqrt{\beta_2}} ( \ln \mathbb{E}\sum_{t=1}^{T+1} \sum_{l=1}^d \sqrt{\Tilde{u}_{t,l}} - \ln d). \nonumber
\end{eqnarray}
By adopting Lemma \ref{bound}, we have that
\begin{eqnarray}
    \sum_{t=1}^{T+1} \sum_{l=1}^d \mathbb{E}\sqrt{\Tilde{u}_{t,l}} &\leq& \frac{6(1+\sqrt{\beta_2})D_1}{\sqrt{\beta_2}}(C_4 +\frac{2dC_6}{\sqrt{\beta_2}} \ln (\frac{6(1+\sqrt{\beta_2}) d D_1 C_6}{\beta_2}+e) - 2(C_5 + \frac{C_6}{\sqrt{\beta_2}})d \ln d) + 2(T+1)d\sqrt{D_0+\epsilon^2}\nonumber \\
    &&+  \frac{12(1+\sqrt{\beta_2})d D_1 C_5}{\sqrt{\beta_2}} \ln \sum_{l=1}^d \mathbb{E}  \sqrt{\check{u}_{T,l}} \nonumber
\end{eqnarray}
Further, since $\sqrt{\check{u}_{T,l}} \leq \sqrt{(1-\beta_2)\sum_{t=1}^T \nabla f(x_t)_l^2} \leq \sqrt{1-\beta_2}\sum_{t=1}^T |\nabla f(x_t)_l|$, we have 
\begin{eqnarray}
    &&(\sqrt{1-\beta_2}\sum_{l=1}^d \sum_{t=1}^T \mathbb{E} |\nabla f(x_t)_l|)^2 \leq (1-\beta_2)(\sum_{t=1}^T \sum_{l=1}^d \mathbb{E} \frac{\nabla f(x_t)_l^2}{\sqrt{\Tilde{u}_{t,l}}})(\sum_{t=1}^{T+1} \sum_{l=1}^d \mathbb{E}\sqrt{\Tilde{u}_{t,l}}) \nonumber \\
    &\leq& (1-\beta_2)\bigg(C_4 + 2dC_5 \ln \sum_{l=1}^d \mathbb{E}  \sqrt{\check{u}_{T,l}} - 2d C_5 \ln d + \frac{2d C_6}{\sqrt{\beta_2}} \ln (C_7 + \frac{12(1+\sqrt{\beta_2})dC_5 D_1}{\sqrt{\beta_2}} \ln \sum_{l=1}^d \mathbb{E}  \sqrt{\check{u}_{T,l}}) - \frac{2d C_6}{\sqrt{\beta_2}} \ln d\bigg) \nonumber \\
    &&\times \bigg(C_7+\frac{12(1+\sqrt{\beta_2})dC_5 D_1}{\sqrt{\beta_2}} \ln \sum_{l=1}^d \mathbb{E}  \sqrt{\check{u}_{T,l}}\bigg) \nonumber \\
    &\leq& (1-\beta_2)\bigg(2dC_5 \ln \sqrt{1-\beta_2} \sum_{l=1}^d \sum_{t=1}^T \mathbb{E} |\nabla f(x_t)_l| + \frac{2d C_6}{\sqrt{\beta_2}} \ln C_7 + \frac{24(1+\sqrt{\beta_2})d^2 C_5 C_6 D_1}{\beta_2 C_7} \ln \sqrt{1-\beta_2}\sum_{l=1}^d \sum_{t=1}^T \mathbb{E} |\nabla f(x_t)_l|\nonumber \\
    && + C_4 - 2(C_5 + \frac{C_6}{\sqrt{\beta_2}})d \ln d\bigg) \times \bigg(C_7+\frac{12(1+\sqrt{\beta_2})dC_5 D_1}{\sqrt{\beta_2}} \ln \sqrt{1-\beta_2}\sum_{l=1}^d \sum_{t=1}^T \mathbb{E} |\nabla f(x_t)_l|\bigg) \nonumber \\
    &=& (1-\beta_2)(C_8 + C_9\ln \sqrt{1-\beta_2}\sum_{l=1}^d \sum_{t=1}^T \mathbb{E} |\nabla f(x_t)_l| + C_{10} (\ln \sqrt{1-\beta_2} \sum_{l=1}^d \sum_{t=1}^T \mathbb{E} |\nabla f(x_t)_l|)^2) \nonumber
\end{eqnarray}
where
\begin{equation}
    C_7 = \frac{6(1+\sqrt{\beta_2})D_1}{\sqrt{\beta_2}}(C_4 +\frac{2dC_6}{\sqrt{\beta_2}} \ln (\frac{6(1+\sqrt{\beta_2}) d D_1 C_6}{\beta_2}+e) - 2(C_5 + \frac{C_6}{\sqrt{\beta_2}})d \ln d) + 2(T+1)d\sqrt{D_0+\epsilon^2}, \nonumber
\end{equation}
\begin{equation}
    C_8 = C_7(\frac{2d C_6}{\sqrt{\beta_2}} \ln C_7 + C_4 - 2(C_5 + \frac{C_6}{\sqrt{\beta_2}})d \ln d), \nonumber
\end{equation}
\begin{equation}
    C_9 = 2dC_5 C_7 + \frac{24(1+\sqrt{\beta_2})d^2 C_5 C_6 D_1}{\beta_2} + \frac{12(1+\sqrt{\beta_2})dC_5 D_1}{\sqrt{\beta_2}}(\frac{2d C_6}{\sqrt{\beta_2}} \ln C_7 + C_4 - 2(C_5 + \frac{C_6}{\sqrt{\beta_2}})d \ln d), \nonumber
\end{equation}
\begin{equation}
    C_{10} = (2dC_5 + \frac{24(1+\sqrt{\beta_2})d^2 C_5 C_6 D_1}{\beta_2 C_7})\frac{12(1+\sqrt{\beta_2})dC_5 D_1}{\sqrt{\beta_2}}. \nonumber
\end{equation}
Solving the above inequality with Lemma \ref{bound2} yields that
\begin{eqnarray}
    \sum_{t=1}^T \mathbb{E} \|\nabla f(x_t)\|_1 = \sum_{t=1}^T \sum_{l=1}^d \mathbb{E} |\nabla f(x_t)_l| \leq \sqrt{2C_8 + 2C_9 \ln (C_9+e) + 64 (1-\beta_2) C_{10}^2 + 1}. \nonumber
\end{eqnarray}
Considering that $1-\beta_2 = O(T^{-1})$, $\eta=O(T^{-\frac{1}{2}})$, $\rho=O(T^{-\frac{1}{2}})$, we in sequence obtain that
\begin{equation}
    C_0 = O(1), C_1 = O(1), \nonumber
\end{equation}
\begin{eqnarray}
    C_2 &=& O(\frac{\rho^2}{(1-\beta_2)\eta}) + O(\frac{\rho}{\sqrt{1-\beta_2}\eta}) + O(\frac{1}{1-\beta_2}(\frac{\rho^2}{\sqrt{1-\beta_2}}+\rho+\rho^2+\eta)(1+T(1-\beta_2))) + O(\frac{1}{\eta}) \nonumber \\
    &&+ O(\frac{1}{1-\beta_2}(\rho+\sqrt{1-\beta_2})(1+T(1-\beta_2))) \nonumber \\
    &=& O(\sqrt{T}) \nonumber
\end{eqnarray}
Here we adopts that $\ln \frac{1}{\beta_2} \leq \frac{1-\beta_2}{\beta_2}$.
\begin{equation}
    C_3 = O(\frac{1}{1-\beta_2}(\rho + \sqrt{1-\beta_2}+\eta + \frac{\rho^2}{\sqrt{1-\beta_2}} + \rho^2)) = O(\sqrt{T}) \nonumber
\end{equation}
\begin{eqnarray}
    C_4 &=& O(\frac{1}{\sqrt{1-\beta_2}}\!+\!\sqrt{1-\beta_2} + \eta + (\frac{\eta^2 + \rho^2}{(1-\beta_2)^{3/2}} + \frac{\rho^2+1}{\sqrt{1-\beta_2}} + \sqrt{1-\beta_2} + \eta)(1+T(1-\beta_2)) + \frac{1}{\eta}\!+\!\frac{1}{\sqrt{1-\beta_2}} + C_2) \nonumber \\
    &=& O(\sqrt{T)} \nonumber
\end{eqnarray}
\begin{eqnarray}
    C_5 = O(\frac{\rho^2}{\sqrt{1-\beta_2}} + \frac{1}{1-\beta_2}(\frac{\rho^2}{\sqrt{1-\beta_2}} + \rho)) = O(\sqrt{T}) \nonumber
\end{eqnarray}
\begin{equation}
    C_6 = O(\frac{\rho^2}{\sqrt{1-\beta_2}} + \frac{1}{\sqrt{1-\beta_2}} + C_3 + \frac{\eta^2}{(1-\beta_2)^{3/2}} + \sqrt{1-\beta_2} + \eta) = O(\sqrt{T}) \nonumber
\end{equation}
\begin{equation}
    C_7 = O(C_4 + C_6 \ln C_6 + T) = O(T) \nonumber
\end{equation}
\begin{equation}
    C_8 = C_7(C_6 \ln C_7 + C_4) = O(T^{3/2} \ln T) \nonumber
\end{equation}
\begin{equation}
    C_9 = O(C_5 C_7 + C_5 C_6 + C_5(C_6 \ln C_7 + C_4)) = O(T^{3/2}) \nonumber
\end{equation}
\begin{equation}
    C_{10} = (C_5 + \frac{C_5 C_6}{C_7})C_5 = O(T) \nonumber
\end{equation}
and finally,
\begin{eqnarray}
    \frac{1}{T}\sum_{t=1}^T \mathbb{E} \|\nabla f(x_t)\|_1 = O(\frac{\ln T}{T^{1/4}}). \nonumber
\end{eqnarray}
\end{IEEEproof}

\end{document}